\documentclass{article}

\PassOptionsToPackage{numbers, compress}{natbib}
\usepackage[main, preprint]{neurips_2026}

\usepackage[utf8]{inputenc}
\usepackage[T1]{fontenc}
\usepackage{hyperref}
\usepackage{url}
\usepackage{booktabs}
\usepackage{amsfonts}
\usepackage{amsmath}
\usepackage{amssymb}
\usepackage{nicefrac}
\usepackage{microtype}
\usepackage[table]{xcolor}
\usepackage{calc}
\usepackage{graphicx}
\usepackage{algorithm}
\usepackage{algpseudocode}
\usepackage{multirow}
\usepackage{subcaption}
\usepackage{enumitem}
\usepackage{wrapfig}
\usepackage{float}
\usepackage{algorithm}
\usepackage{algpseudocode}
\usepackage{amsmath, amssymb}
\usepackage[capitalize,noabbrev]{cleveref}
\usepackage[most]{tcolorbox}
\usepackage{tikz}
\usetikzlibrary{fit,calc}
\usepackage{listings}
\usepackage{makecell}
\usepackage{fontawesome5}

\definecolor{CustomRed}{HTML}{E30A17}
\definecolor{DarkCyan}{HTML}{0E9594}
\definecolor{OrangeWeb}{HTML}{FFA400}
\definecolor{Avocado}{HTML}{688E26}
\definecolor{TyrianPurple}{HTML}{550527}
\definecolor{BurntUmber}{HTML}{772E25}
\definecolor{FireEngineRed}{HTML}{C33626}
\definecolor{Carmine}{HTML}{990011}
\definecolor{CaribbeanCurrent}{HTML}{187077}
\definecolor{PersianIndigo}{HTML}{E3B505}
\definecolor{LightGreen}{HTML}{D2f4d3}
\definecolor{PigmentGreen}{HTML}{08A045}
\definecolor{VanillaPink}{HTML}{E57a81}

\colorlet{PrimaryColor}{CaribbeanCurrent}
\colorlet{SecondaryColor}{OrangeWeb}
\colorlet{ModelCompletion}{PigmentGreen}  

\hypersetup{
    colorlinks=true,
    linkcolor=red,
    citecolor=blue,
    filecolor=magenta,
    urlcolor=blue,
linktocpage}

\colorlet{mypink}{red!30}
\colorlet{myblue}{orange!30}
\colorlet{mypurple}{green!25}

\definecolor{codebg}{HTML}{E8F0FE}
\newcommand{\agenticlogo}{%
  \raisebox{-0.25ex}{\includegraphics[height=1.15em]{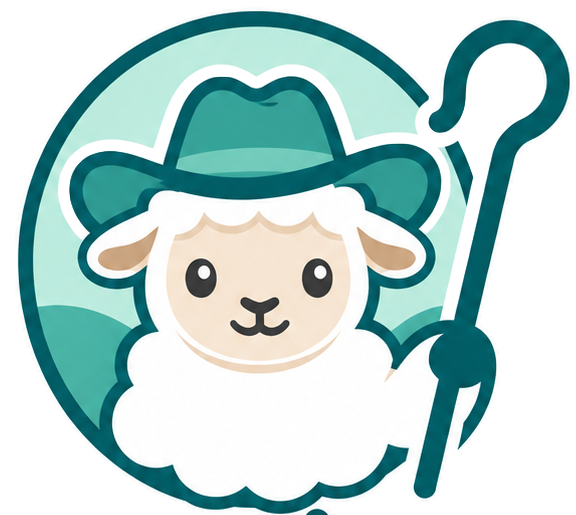}}\hspace{0.15em}%
  \textcolor{PrimaryColor}{\textbf{\textsc{Shepherd}}}%
}
\newcommand{\logonoback}{
\raisebox{-0.25ex}{\includegraphics[height=1.15em]{assets/logo/shepherd_logo_0612.png}}\hspace{0.0em}
}
\newcommand{\agenticlogonoback}{%
  \logonoback%
  \textcolor{PrimaryColor}{\textbf{\textsc{Shepherd}}}%
}
\newcommand{\agentic}{\textcolor{PrimaryColor}{\textbf{\textsc{Shepherd}}}}

\newcommand{\agentictitle}{%
  \raisebox{-0.4ex}{\includegraphics[height=1.5em]{assets/logo/shepherd_logo_0612.png}}\hspace{0.2em}%
  \textcolor{PrimaryColor}{\textbf{\textsc{Shepherd}}}%
}

\newcommand{\cbo}{CRO}

\newcommand{\alternativeone}[1]{\textcolor{blue}{#1}}

\renewcommand{\alternativeone}[1]{}

\newif\ifdraftresults
\draftresultstrue

\newsavebox{\hbfullbox}\sbox{\hbfullbox}{\tikz[baseline=-0.4ex]\fill (0,0) circle (0.42ex);}
\newsavebox{\hbhalfbox}\sbox{\hbhalfbox}{\tikz[baseline=-0.4ex]{\draw (0,0) circle (0.42ex); \fill (0,0.42ex) arc[start angle=90, end angle=270, radius=0.42ex] -- cycle;}}
\newsavebox{\hbemptybox}\sbox{\hbemptybox}{\tikz[baseline=-0.4ex]\draw (0,0) circle (0.42ex);}
\newcommand{\capfull}{\usebox{\hbfullbox}}
\newcommand{\caphalfr}{\reflectbox{\usebox{\hbhalfbox}}}
\newcommand{\caphalf}{\usebox{\hbhalfbox}}
\newcommand{\capempty}{\usebox{\hbemptybox}}

\title{\agentictitle{}: Enabling Programmable Meta-Agents via Reversible Agentic Execution Traces

}

\author{%
  Simon Yu$^{*1}$ \quad Derek Chong$^{*2}$ \quad Ananjan Nandi$^{*2}$ \\
  \textbf{Dilara Soylu}$^{2}$ \quad \textbf{Jiuding Sun}$^{2}$ \quad
  \textbf{Christopher D Manning}$^{2}$ \quad \textbf{Weiyan Shi}$^{1}$ \\
  $^{1}$Northeastern University \quad $^{2}$Stanford University \\
  \texttt{\{yu.chi, we.shi\}@northeastern.edu} \\
\texttt{\{derekch, ananjan, soylu, sunjd24, manning\}@stanford.edu} \\
  $^{*}$Equal contribution
}

\begin{document}

\maketitle

\vspace{-3 em}
\begin{center}
\large
\href{https://shepherd-agents.ai}{\textcolor{black}{\faGlobe}\ Website}\quad
\href{https://shepherd-agents.ai/blog}{\textcolor{black}{\faBook}\ Blog}\quad
\href{https://github.com/shepherd-agents/shepherd}{\textcolor{black}{\faGithub}\ Framework}
\end{center}

\begin{figure}[!h]
 \normalsize
  \centering
  \vspace{-1em}\includegraphics[width=0.95\linewidth]{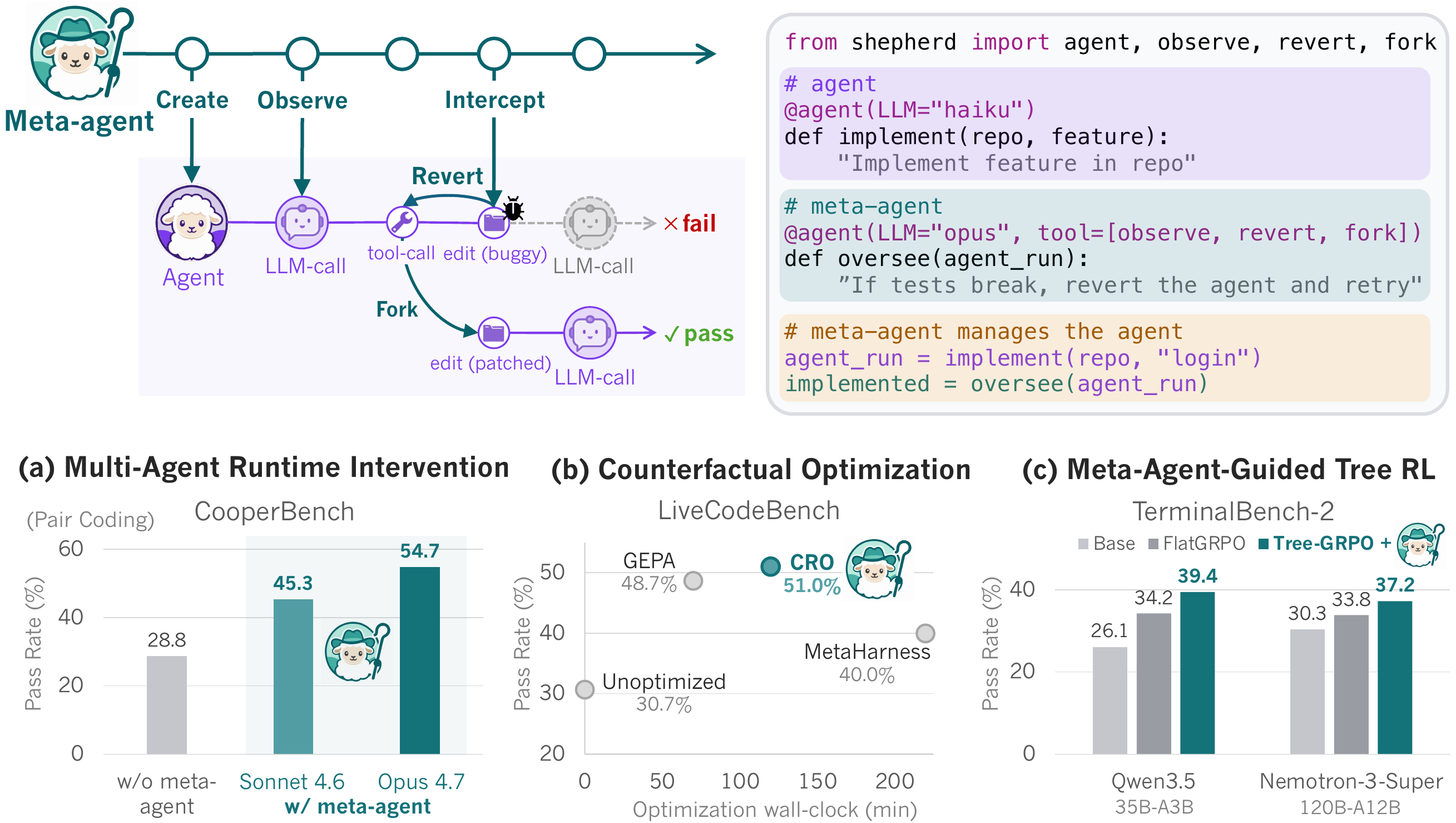}
  \caption{
  \textbf{\agentic{} meta-agents.} \emph{Top:} A meta-agent optimizes an agent's execution trace. \emph{Bottom:} Results from three meta-agents: (A) runtime intervention; (B) meta-optimization; \textbf{(C)} Tree-GRPO.
  }
  \label{fig:teaser}
\end{figure}
\begin{abstract}
As LLM agent systems take on more complex tasks, they increasingly rely on meta-agents: higher-order agents that create, operate on and manage other agents. Meta-agent operations such as coordinating agents, halting risky actions before execution, or repairing failed runs, require runtime manipulation of agentic execution.
Yet existing agentic substrates make this difficult: they expose only transcripts and environment snapshots, forcing meta-agents to build ad hoc tooling to reconstruct and operate over full execution state.
Therefore, we introduce \agentic{}, a Python substrate grounded in
functional programming principles, where an agent's execution is itself a \textit{first-class object} that a meta-agent can easily inspect and transform. Every model action, tool call, and environment change becomes a structured event in a reversible, Git-like execution trace, where any past state can be reverted 5× faster than \texttt{docker commit} and fork. Three example use cases show \agentic{}'s versatility: (1) a supervisor meta-agent prevents conflicts among parallel coding agents, lifting pair-coding pass rate from 28.8\% to 54.7\% on CooperBench; (2) a counterfactual optimization meta-agent repairs agent workflows by proposing edits and replaying runs from the point of changed behavior, outperforming MetaHarness on Terminal-Bench~2.0 by 12.8\% with 58\% lower wall-clock; (3) a training meta-agent picks fork points during rollouts to improve credit assignment in long-horizon agentic RL, doubling GRPO's uplift on Terminal-Bench~2.0. We open-source \agentic{} to enable principled and efficient operations over agentic execution for both users and meta-agents.

\end{abstract}

\section{Introduction}
\label{sec:intro}

As LLM-based agentic systems mature, we increasingly see the use of agents that act on other agents at runtime.
\citet{asawa_how_2026, lin2025stop} develop advisor
agents that learn intervention policies from execution traces to steer agents away from dead ends; meta-optimizers such as GEPA and MetaHarness optimize agentic workflows to improve performance on a given downstream task ~\citep{agrawal_gepa_2025, lee_meta-harness_2026}; and
\citet{hou2025treerlllmreinforcementlearning, ji2026treesearchllmagent}
build tree-search RL that branches rollouts to compare rewards from alternate continuations and improve per-step credit assignment. We call these systems \emph{meta-agents}: higher-order agents that operate over other agents and their execution traces, during or after execution. Meta-agents are increasingly central to extracting capability from agentic systems~\citep{zhang_mgh_2026}.

\begin{wraptable}{r}{0.6\textwidth}
\vspace{-1.0em}
\caption{\textbf{Support for meta-agent operations on a running agent.}
\capfull{} = fully supported;
\caphalf{} = supported only for the agent; \caphalfr{} = supported only for the environment;
\capempty{} = not supported.}
\label{tab:framework-capabilities}
\centering\footnotesize
\begingroup
\setlength{\tabcolsep}{2.5pt}
\renewcommand{\arraystretch}{1.08}
\resizebox{\linewidth}{!}{%
\begin{tabular}{@{}lcccc@{}}
\toprule
\textbf{Method}
& \makecell{\textbf{Intercept} \\ execution} & \makecell{\textbf{Fork}  \\ agent + env.} & \makecell{\textbf{Revert} \\ to past state} & \makecell{\textbf{Modify} \\ agent behavior} \\
\midrule
BranchFS
  & \capempty & \caphalfr & \caphalfr & \capempty \\
Docker
  & \capempty & \caphalfr & \caphalfr & \capempty \\
OpenHands
  & \caphalf & \caphalf & \caphalf & \capempty \\
AgentGit
  & \capempty & \caphalf & \caphalf & \capempty \\
\rowcolor{PrimaryColor!15}
\textbf{\agentic{}}
  & \capfull & \capfull & \capfull & \capfull \\
\bottomrule
\end{tabular}
}
\endgroup
\vspace{-0.8em}
\end{wraptable}


Yet, existing agentic substrates are not designed for meta-agents. Consider a supervisor meta-agent that forks a coding agent before a risky write, watches the branch execute, and reverts the change in the event of failure. To do this, the meta-agent must \textit{observe} the running agent, \textit{intercept} and \textit{fork} it before the write, \textit{revert} it on failure, \textit{modify} it to fix the failure, and then \textit{resume} execution. Recent work exposes fragments of these operations: OpenHands surfaces a session's event stream~\citep{wang2026openhands}, AgentGit gives the worker Git-like commit tools~\citep{li_agentgit_2025}, BranchFS isolates the filesystem~\citep{wang_fork_2026} (\Cref{tab:framework-capabilities}). However, these substrates are designed to maintain runtime state for the running agent, not to give a meta-agent the operations it needs to act on that agent. As a result, meta-agent implementations need to reinvent custom tooling to support these operations in practice.

We argue for a different approach: give an agent and its execution the first-class treatment functions get in functional programming – \textit{structured data} a meta-agent can hold, execute, copy, and rewrite. This approach then enables \textit{algebraic effect handlers}~\citep{plotkin_pretnar_2009} that intercept and observe execution without modifying it, allowing its reversion, as well as  \textit{continuations}~\citep{felleisen_1988} that support pause-inspect-decide-resume patterns for meta-agents. We therefore propose
\agenticlogo{}, a principled functional programming model for higher-order agents, instantiated as an intuitive Python substrate (Figure~\ref{fig:teaser}, Section~\ref{sec:primitives}).

\agentic{} defines agents as \textit{tasks} with typed inputs and outputs, and records the execution of these tasks in a reversible, Git-like
\emph{execution trace}: every agent action (tool call, filesystem modification, database operation) becomes a commit in this trace, every fork is a branch, and every past agent-environment state can be reverted to through checkouts.
A meta-agent can then \emph{observe} a task's execution by subscribing to its commits, \emph{intercept} by pushing events into its trace, \emph{revert} to any prior agent-environment state by checking out the corresponding commit, \emph{fork} a new copy of this state, and
\emph{modify} the task itself by rewriting its definition.

Our principled functional programming grounding gives \agentic{} meta-agents several powerful capabilities:
an observing meta-agent does not perturb the observed agent's execution, parallel agents can run as isolated processes over forked environments within a shared sandbox, and reverting to any commit restores a byte-identical copy of the corresponding agent-environment state. 
We formalize these properties through a small algebraic-effects calculus mechanized in Lean, which also provides a precise semantic contract for the execution trace.
The implementation is model-agnostic and 
lightweight: for a 5.8\,GB docker image, \agentic{} forks the agent-environment state at $5\times$ the speed of a \texttt{docker
commit}, and reuses over $95\%$ of the LLM provider's KV cache.

\agentic{} enables the easy implementation of meta-agent applications that
previously required substantial bespoke engineering. We showcase three example meta-agents
spanning a typical agent lifecycle. \emph{During execution}, a \emph{runtime
supervisor meta-agent} (\S\ref{sec:during}) watches the execution traces of parallel coding workers and intercepts before they conflict, raising CooperBench~\cite{khatua_cooperbench_2026} joint pass rate from
28.8\% to 54.7\%. \emph{After execution}, a \emph{counterfactual meta-optimizer} (\S\ref{sec:cbo}) diagnoses failures in prior agent runs and validates proposed fixes by branching runs at the first point where the fix would change behavior, outperforming state-of-the-art meta-optimizers such as  MetaHarness~\cite{lee_meta-harness_2026} by up to 27.5\% on benchmarks such as LiveCodeBench~\cite{jain2024livecodebenchholisticcontaminationfree} and Terminal-Bench~2.0~\cite{merrill2026terminalbenchbenchmarkingagentshard}, while cutting wall-clock time by up to 58\%.
\emph{During training}, a \emph{tree-search RL trainer}
(\S\ref{sec:training}) forks rollouts at meta-agent-chosen
turns and samples sibling continuations to improve credit assignment by computing advantages from outcome rewards, outperforming GRPO~\cite{shao2024deepseekmathpushinglimitsmathematical}'s performance when used to train Qwen3.5-35B-A3B~\cite{qwen3.5} on Terminal-Bench~2.0 by 5.2 points.

In summary, we contribute (i) \agenticlogo{}, a programming model for higher-order agents whose core operations are grounded in functional programming and mechanized in Lean; (ii) a Python framework instantiating this model in a performant and efficient substrate; and (iii) three meta-agents spanning the agent lifecycle built using the \agentic{} substrate.

\section{Related Work}
\label{sec:related}

\paragraph{Meta-Agents.} Meta-agent applications are emerging in recent work~\citep{zhang_mgh_2026, zhang2026recursivelanguagemodels}. Darwin-G\"{o}del Machines~\citep{zhang_dgm_2025} and Group-Evolving Agents~\citep{weng_gea_2026} maintain dynamic archives of self-modifying code.  Hyperagents~\citep{zhang_hyperagents_2026} and Meta-Harness~\citep{lee_meta-harness_2026} optimize a task agent's problem-solving strategies at the meta level. Other lines refine an agent's context and long-term retrieval through evolutionary search~\citep{novikov_alphaevolve_2025, wang_thetaevolve_2025} or memory-augmented architectures~\citep{zhang_ace_2025, li_combee_2026, zhang_memrl_2026, xia_agent0_2025, wei_evomemory_2025, shinn_reflexion_2023}. Each of these methods reinvents the runtime machinery needed to act on agents – parsing transcripts, building bespoke environment snapshots, re-executing with modified source code. \agentic{} provides a unified substrate for these operations, letting a meta-agent observe agents without perturbing them, fork their coupled agent-environment state, and replay prior execution byte-identically.


\paragraph{Agentic Meta-Optimization.} Orchestrating multiple LLM agents is a common strategy for performing complex tasks and inference-time scaling. Standard multi-agent frameworks~\citep{wu_autogen_2023, hong_metagpt_2023, noauthor_httpswwwanthropiccomengineeringmanaged-agents_nodate} route natural-language messages between workers; CooperBench~\citep{khatua_cooperbench_2026} shows the coordination failures that can result from this. Pipeline optimizers~\citep{khattab_dspy_2023, cheng_trace_2024, zhou_language_2023} and test-time scaling methods~\citep{kim_scalingtts_2026, lee_aggagent_2026} use parallel rollouts and majority voting, evaluating each candidate by full end-to-end re-execution. GEPA~\citep{agrawal_gepa_2025} introduces a reflective meta-agent that proposes workflow edits. These methods treat the underlying execution as a black box and re-run candidates from scratch. \agentic{} adds a different evaluation primitive: a meta-agent can branch a worker's context at the exact commit where an edit first alters behavior and replay only the affected suffix (\Cref{sec:microbench}), so candidates reuse computation effectively.

\paragraph{Agentic Runtime and Infrastructure.} A parallel line of research adds infrastructure support for agentic state management, placing the checkpoint primitive at different layers of the software stack. AgentGit~\citep{li_agentgit_2025} exposes version-control operations as cooperative tools the agent can invoke from within a LangGraph workflow. BranchFS~\citep{wang_fork_2026} adds a kernel-level \texttt{branch()} system call that isolates filesystem state, independent of the worker's tool-call structure. AgentSPEX~\citep{wang_agentspex_2026} embeds checkpointing into a domain-specific language for agent workflows. Each of these places the checkpoint primitive at a different point in the stack, with different trade-offs between agent autonomy, transparency, and language-level integration. \agentic{} sits at a different point in this design space: it couples environment states with the agent execution state, so the substrate observes the same events the worker emits without requiring the worker to be rewritten or replayed from scratch. This lets the same substrate support runtime supervision, post-hoc trajectory optimization, meta-optimization and stateful RL under a unified interface.

\section{The \agenticlogo{} Programming Model}
\label{sec:primitives}

The supervisor meta-agent in Figure~\ref{fig:teaser} needs the following operations on worker agents: \emph{observe} what it is doing at runtime, \emph{intercept} and \emph{fork} execution before a risky write to try an alternative continuation, \emph{revert} changes in case of failure, and then \emph{modify} the agent to fix the issue and \emph{resume} execution. These are operations difficult to support in existing substrates because agentic execution is full of side effects such as model calls, filesystem writes, and tool invocations, that resist being treated as inspectable, manipulable values.
\begin{wraptable}{r}{0.55\textwidth}
\caption{Mapping of \agentic{} primitives to runtime meta-agent operations enabled and functional programming (FP) constructs.}
\label{tab:map}
\centering\footnotesize
\begingroup
\setlength{\tabcolsep}{2.5pt}
\renewcommand{\arraystretch}{1.08}
\resizebox{\linewidth}{!}{%
\begin{tabular}{lll}
\toprule
\textbf{Primitive} & \textbf{Runtime Operation Enabled} & \textbf{FP construct} \\
\midrule
Task &  Modify agent behavior & typed function \\
Effect & Observe and intercept execution & algebraic effect \\
Scope & Fork execution state & scoped effect handler \\
Trace & Revert and replay from past state & persistent data structure \\
\bottomrule
\end{tabular}
}
\endgroup
\vspace{-0.8em}
\end{wraptable}

Functional programming has a long tradition of structuring effectful computation by drawing a boundary separating \emph{what} a computation describes from \emph{how} its effects reach the world. Computation inside this boundary becomes observable, substitutable, branchable, and replayable. \agentic{} extends this discipline to agentic execution, treating it as a first-class object~\citep{xia_interaction_2020}. Doing so requires elevating four concepts to first-class status: what the agent \emph{is} (\textbf{tasks}, \S~\ref{sec:task}), what it \emph{does} (\textbf{effects}, \S~\ref{sec:effect}), where it \emph{runs} (\textbf{scopes}, \S~\ref{sec:scope}), and what has already been \emph{done} (\textbf{execution trace}, \S~\ref{sec:trace-graph}). Each primitive is grounded in a functional-programming construct (Table~\ref{tab:map}), and key properties of \agentic{} rest on a small Lean-mechanized semantics for typed effect traces (Appendix~\ref{app:lean_proof}).

\vspace{-1em}
\subsection{Task: Agent Definition}
\label{sec:task}


For a supervisor meta-agent to modify an agent – by changing its code, composing them into workflows or even synthesizing helper agents at runtime – the agent must be a \emph{value}: something that can be held, passed as an argument, or returned from a call. Functional programming gives this property to functions. A function with a typed input and a typed output is substitutable for any other function with the same type, and a higher-order function can take functions as arguments. Agents in \agentic{} have the same shape. A \textbf{task} is a typed function over agentic execution with typed input and output, and a body that may call LLMs, tools, and other tasks. They are declared using the \texttt{@agent} decorator over typed Python functions, and can call models through \emph{providers}:

\begin{lstlisting}
@agent(LLM="haiku")
def implement(repo: GitWorkspace, feature: str) -> GitPatch:
"""Implement the feature in the repo."""

@agent(LLM="opus")
def oversee(agent_run: Task[GitPatch]) -> GitPatch:
"""Watch the agent; fork before risky edits; merge the passing patch."""
\end{lstlisting}


\paragraph{Tasks are substitutable values.} A task is fully specified by its signature and docstring – \agentic{} compiles them into an LLM prompt guided by the docstring, where the output is validated against the defined type. Users may also implement a body for the task when a mix of deterministic and LLM-driven computation is required. As shown above, meta-agents in \agentic{} are just tasks whose arguments happen to be other tasks, and they therefore hierarchically allow for meta-meta-agents over their execution. Any task with the same typed signature can also substitute for each other, allowing meta-agents to easily edit behavior at runtime. Because a task is fully specified by this typed signature and docstring, a meta-agent can also \emph{create} new sub-agents at runtime by synthesizing fresh task definitions, not only modify existing ones.


A task turns an agent into a value, but when an agent interacts with its environment, the substrate still has no handle on it. The next primitive gives every such action a typed value of its own.

\subsection{Effects: Agent Actions}
\label{sec:effect}


A supervisor meta-agent needs to see what the worker agent is doing while it executes, without perturbing it. Functional programming solves the analogous problem for effectful programs with \emph{algebraic effects}~\citep{plotkin_power_2003, plotkin_pretnar_2009}: any operation that touches the outside world (read a file, send a message) is reified as a typed event, and a \emph{handler} decides what that event means. The same program can be run under different handlers without changing its source: one handler could execute it, while another logs it. 

\agentic{} applies this discipline to agent actions. An \textbf{effect} is a typed record of a single action attempted by a worker. They can record LLM calls, tool calls, environment mutations, or even user-defined custom actions. Every effect a worker emits is appended to the immutable \textit{effect stream} of the scope it's running in (\S\ref{sec:scope}), and a meta-agent can observe the stream by subscribing to it, or intercept a running execution by pushing effects into the stream.

\paragraph{An action's intent and outcome are separate events.} Each action emits two effects: an \emph{intent} when the worker issues the action, and an \emph{outcome} when the world responds. Therefore, a meta-agent can read an intent, decide it shouldn't materialize, and act before the outcome arrives, as below:

\begin{lstlisting}
async for effect in work.trace.live():
    if isinstance(effect, ToolCallIntent) and is_destructive(effect):
        if check(effect) == "deny":
            work.discard()  # outcome never materializes
\end{lstlisting}

\paragraph{Observation is non-perturbing.} Since the worker's effect stream is immutable, it is byte-identical whether or not a meta-agent is watching. Because intent is decoupled from execution, a meta-agent can also replay a slice of the effect stream under a different handler. For example, a meta-agent can replay a tool-call under a modified task body to see how the underlying agent's behavior changes.

\paragraph{Most effects are reversible.} Every effect carries a reversibility tier that determines its behavior when \textit{materialized}, or executed against the world. \emph{Reversible} (such as filesystem writes, sandbox state) and \emph{compensable} effects (such as database writes) are captured by the substrate at emission time and can be rolled back at will by an user or meta-agent. Reversible effects roll back natively through their scope (\S~\ref{sec:scope}), and compensable effects roll back through user-supplied compensation handlers invoked by the substrate. However, \emph{irreversible} effects (such as model calls, emails) materialize on emission, and the stream can only record them for audit.

Reading, reverting and reinterpreting actions is enough for an observer. However, a meta-agent trying alternative execution paths needs to be able to fork the worker and its environment atomically.

\subsection{Scopes: Agent Environments}
\label{sec:scope}

When the supervisor meta-agent forks an agent, it must run in its own, isolated world. Any effects emitted in the branch must not leak into the parent's effect stream. The functional programming construct enabling this is the \emph{region-scoped effect handler}. A function can open a fresh handler for some sub-region of its execution, run inside it, and then either propagate that region's effects outward (commit) or abandon them (revert). Regions nest cleanly: each level owns its own effect interpretation and cannot contaminate others. A \textbf{scope} in \agentic{} implements this construct for agentic execution. It binds the worker's sandbox handles, model providers, tool surfaces, and effect-stream cursor, and it owns the effects emitted by tasks running inside it.

\paragraph{Scopes support four primitives.} \texttt{emit} writes an effect to the scope's stream. \texttt{fork} opens a copy-on-write child scope. \texttt{merge} propagates a child's effects into its parent. \texttt{discard} abandons a child, leaving the parent untouched. Filling in \texttt{supervise}'s body from \S\ref{sec:task}:

\begin{lstlisting}
@agent
def oversee(work: Task[Patch]) -> Patch:
    child = scope.fork()
    result = await work(child)
    if result.failed: child.discard()
    else: scope.merge(child)
    return result
\end{lstlisting}



\paragraph{The agent and its environment are forked atomically.} \texttt{scope.fork()} captures the worker's filesystem, processes, and bindings in one atomic copy-on-write step. A subsequent \texttt{discard} therefore rolls back every trace of what the worker agent touched. The substrate realizes this through overlay-filesystem virtualization and the native checkpoint facilities of containerized sandboxes, behind a unified device-layer interface (Appendix~\ref{app:backends}).

\paragraph{Scopes can be parallelized, nested, reverted and resumed.} A meta-agent can create concurrent forks from a parent scope and merge or discard each independently. Scopes also nest: a meta-meta-agent can fork, observe, and resume a meta-agent without contaminating the worker beneath. Discarding a child scope leaves the parent byte-identical to the moment of fork. Resuming a scope also preserves the worker agent's world: a paused worker resumed by a meta-agent sees the bindings recorded in its original scope, not whatever the meta-agent currently holds.

A scope owns the \emph{present} region of execution. However, to \emph{revert} to an arbitrary past state, execution history itself must be a navigable value. We address this next.

\subsection{Execution Trace: Agent Execution History}
\label{sec:trace-graph}

A supervisor meta-agent that wants to revert a worker agent – return to an earlier moment in its execution and either inspect it or run forward from there under different conditions – needs every past state of the worker's execution to be reachable on demand. In functional programming, \emph{persistent data structures}~\citep{okasaki_purely_1999} provide this property: every version of the structure remains accessible after modification, with new versions sharing structure with old ones for efficiency. In \agentic{}, the counterpart to this is the \textbf{execution trace}. It is a persistent Git-like commit graph: each scope's effect stream materializes as a sequence of typed commits on a branch of the graph. The four scope operations of \S~\ref{sec:scope} then compile to Git-like operations as well:
\begin{lstlisting}
scope.emit(effect)   <=>  shepherd commit -m "<effect>"
scope.fork()         <=>  shepherd checkout -b <child-branch>
scope.merge(child)   <=>  shepherd merge <child-branch>
scope.discard(child) <=>  shepherd branch -D <child-branch>
\end{lstlisting}

\paragraph{Every past state is reachable and replayable.} A meta-agent can navigate to any commit by hash and read the exact agent-environment state at that moment, with its full scope intact. Locating the specific commit where a regression first appeared, or where two siblings began to diverge, reduces to graph traversal on this trace. Replaying from a past commit also produces a byte-exact reconstruction of the scope before diverging, so the only cost paid by the meta-agent is the executed suffix.

\paragraph{Divergent branches share storage.} Just as persistent data structures share substructures across versions, the execution trace shares storage across branches: two siblings forked from the same commit share their entire prefix by content hash. A meta-agent fanning out across many forks pays only for the divergent suffixes, and any set of branches can be diffed based on their captured effects, therefore letting a meta-agent decide which to merge on the basis of their behavior.

\section{Framework Performance} 
\label{sec:microbench}
\begin{table}[h]
\caption{Fork/revert latency (ms), storage, and host resource cost at $K{=}4$ concurrent branches across three Terminal-Bench~2.0 images. \agentic{}'s 134--143\,ms fork is 2--3\% of one agent turn (Appendix~\ref{app:agent-latency}). Best per group in \textbf{bold}; full $\pm$std and protocol in Appendix~\ref{app:microbench-protocol}.}
\label{tab:framework-perf}
\centering\small
\resizebox{0.85\textwidth}{!}{%
\begin{tabular}{llrrrrr}
\toprule
\multirow{2}{*}{\textbf{Image}} & \multirow{2}{*}{\textbf{Method}} & \multirow{2}{*}{\textbf{Fork $\downarrow$}} & \multirow{2}{*}{\textbf{Revert $\downarrow$}} & \multirow{2}{*}{\textbf{Storage $\downarrow$}} & \multicolumn{2}{c}{\textbf{Branching (K=4)}} \\

& & & & & \textbf{Disk $\downarrow$} & \textbf{RAM $\downarrow$} \\
\midrule
\multirow{5}{*}{\shortstack[l]{openssl-selfsigned-cert\\(42\,MB image)}} & Full copy        & 5{,}154 ms & 2{,}067 ms & 268\,MB & 804\,MB & 112\,MB \\
                          & Docker commit    & 658 ms & 749 ms & 30\,KB & 90\,KB & 29.8\,MB \\
                          & Modal snapshot   & 3{,}764 ms & 2{,}260 ms & --- & --- & --- \\
                          & BranchFS~\citep{wang_fork_2026}         & 266 ms & 360 ms & 12\,KB & 48\,KB & 22.7\,MB \\
\rowcolor{PrimaryColor!15}                          & \textbf{\agenticlogonoback{}} & \textbf{134 ms} & \textbf{142 ms} & \textbf{10\,KB} & \textbf{30\,KB} & \textbf{20.5\,MB} \\
\midrule
\multirow{5}{*}{\shortstack[l]{caffe-cifar-10\\(200\,MB image)}} & Full copy        & 5{,}971 ms & 3{,}446 ms & 645\,MB & 1.9\,GB & 230\,MB \\
                          & Docker commit    & 692 ms & 761 ms & 30\,KB & 90\,KB & 38.3\,MB \\
                          & Modal snapshot   & 3{,}291 ms & 2{,}463 ms & --- & --- & --- \\
                          & BranchFS~\citep{wang_fork_2026}         & 272 ms & 357 ms & 12\,KB & 48\,KB & 29.4\,MB \\
\rowcolor{PrimaryColor!15}                          & \textbf{\agenticlogonoback{}} & \textbf{135 ms} & \textbf{140 ms} & \textbf{10\,KB} & \textbf{30\,KB} & \textbf{27.1\,MB} \\
\midrule
\multirow{5}{*}{\shortstack[l]{pytorch-model-recovery\\(5.8\,GB image)}} & Full copy        & 53{,}462 ms & 25{,}943 ms & 8.3\,GB & 24.9\,GB & 910\,MB \\
                          & Docker commit    & 725 ms & 828 ms & 30\,KB & 90\,KB & 30.2\,MB \\
                          & Modal snapshot   & 3{,}160 ms & 2{,}328 ms & --- & --- & --- \\
                          & BranchFS~\citep{wang_fork_2026}         & 280 ms & 358 ms & 12\,KB & 48\,KB & \textbf{22.7\,MB} \\
\rowcolor{PrimaryColor!15}                          & \textbf{\agenticlogonoback{}} & \textbf{143 ms} & \textbf{147 ms} & \textbf{10\,KB} & \textbf{30\,KB} & 25.7\,MB \\
\bottomrule
\end{tabular}%
}
\end{table}


Three cost properties are crucial for meta-agents in \agentic{} to be performant: \texttt{scope.fork()} must be image-size-independent so branching is affordable; the trace must scale with what the agent writes and not the image so it can persist across long executions; and the byte-identical replay guarantee must reach the LLM provider's prompt cache so re-execution efficiently reuses KV cache. We measure each on real Terminal-Bench~2.0 images. Full results are in Appendix~\ref{app:framework-perf-extended}.


\paragraph{Fork and revert are fast and image-size-independent.} \agentic{} fork creates a new copy-on-write layer on top of the existing filesystem instead of duplicating it, so cost is constant regardless of image size. As a result, forks take 134–143 ms regardless of image size (42 MB to 5.8 GB); on the 5.8 GB image, K forks cost K × 143 ms against K × 53.5 s for full-rootfs copies, a 192× per-branch slowdown (Table~\ref{tab:framework-perf}). The next-fastest alternative, BranchFS~\citep{wang_fork_2026}, branches the filesystem alone via FUSE; like the other methods, it supports no notion of the agent itself (Table~\ref{tab:framework-capabilities}).


\paragraph{Replay reuses the LLM provider's prompt cache.} Because \agentic{} fork preserves the parent's byte-identical LLM message prefix, the provider's prompt cache resolves it without invalidation. On Anthropic Claude Haiku 4.5 across 8 Terminal-Bench~2.0 tasks, cache-hit rate plateaus at $\sim$95\% from $K{=}2$ onwards, within 5\% of the byte-identical ceiling. Cache reuse compounds whenever a meta-agent fans out (Tree-GRPO siblings, \Cref{sec:training}) or replays completed trajectories (trajectory compression, \Cref{app:trajprune}). We provide further detail in Appendix~\ref{app:kvcache-detail}.

\section{Experiments}
\label{sec:results}

\begin{figure}[h!]
  \centering
  \includegraphics[width=\linewidth]{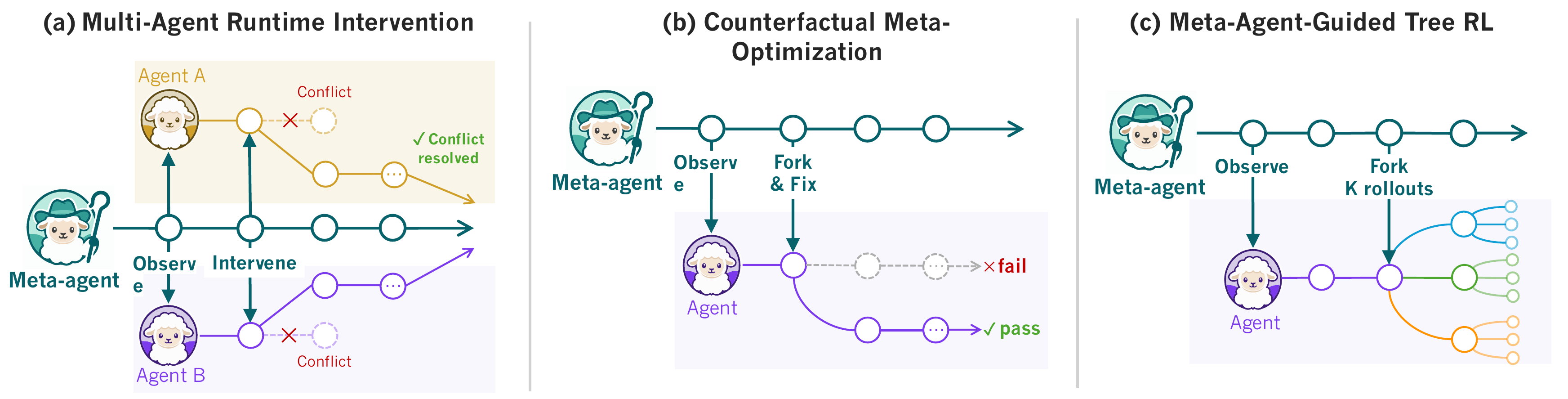}
  \caption{Headline results from the three meta-agent applications built on \agentic{}: (A)~runtime supervision on CooperBench, (B)~counterfactual meta-optimization (\cbo{}), and (C)~meta-agent-guided Tree-GRPO on Terminal-Bench~2.0.}
  \label{fig:results-teaser}
\end{figure}

The \agentic{} substrate enables a wide range of meta-agent applications. In this section, we demonstrate three such applications, \textit{spanning the agent development stack}. (1) \textit{During execution}, a \textbf{runtime supervisor agent} observes two parallel coding workers and intercepts them mid-trajectory, raising the pair pass rate on CooperBench from 28.8\% to 54.7\% 
(\S\ref{sec:during}). (2) For \textit{post-hoc workflow optimization}, a \textbf{meta-optimizer} branches execution traces to test counterfactual workflow edits, outperforming
MetaHarness on four of five datasets at up to 58\% lower wall-clock (\S\ref{sec:cbo}). (3) For \textit{agentic RL training}, a meta-agent selects fork points during RL rollouts to extract per-step credit, lifting Terminal-Bench~2.0 by $5.2$ points on Qwen3.5-35B-A3B over Flat GRPO (\S\ref{sec:training}).

Each application exercises a different \agentic{} property. \S~\ref{sec:during} requires \emph{non-perturbing observation}: a supervisor that subscribes to both workers' effect streams gains access to their traces without perturbing them. \S~\ref{sec:cbo} leans on \emph{byte-identical replay}: candidate edits are validated against a fixed baseline by re-executing only the affected suffix. \S~\ref{sec:training} exercises \emph{cheap branching}: \agentic{}'s cheap agent–environment state forking makes per-step credit assignment via sibling rollouts affordable. We report a fourth use case, which compresses completed trajectories into shorter reruns under meta-agent hindsight, in Appendix~\ref{app:trajprune}. We note that these applications are not exhaustive and discuss further possible applications enabled by \agentic{} in Appendix \ref{app:future_works}.

\subsection{Meta-Agent for Multi-Agent Coordination: Runtime Supervisor}
\label{sec:during}

\paragraph{Motivation.} CooperBench~\citep{khatua_cooperbench_2026} documents a \emph{curse of coordination}: even when allowed to communicate, parallel coding agents coordinate poorly enough that they succeed less often than a single agent working alone. \agentic{}'s effect stream and scope primitives let a meta-agent close that gap by inspecting both workers' execution in real time and intercepting before damage compounds.

\paragraph{Method.} Two Claude Haiku~4.5 worker agents run in parallel forked scopes, each assigned one complementary feature to implement. A Claude Sonnet~4.6 or Opus~4.7 meta-agent subscribes to both effect streams via \agentic{} and is provided with three coordination tools: \texttt{inject} (push guidance into a worker's session), \texttt{handoff} (fork the leading worker's scope as the follower's new root and restart), and \texttt{discard} (abort a stuck worker via \texttt{scope.discard()}).

\paragraph{Setup.} We evaluate on CooperBench~\citep{khatua_cooperbench_2026}. Baselines are \emph{solo} (one Haiku~4.5 agent handling both features sequentially) and \emph{coop} (two parallel Haiku~4.5 agents in forked scopes with peer-to-peer messaging via the relay sandbox, no supervisor). Full protocol, dataset construction, and per-condition operational notes are in Appendix~\ref{app:live-detail}.


\begin{figure}[h!]
  \centering
  \includegraphics[width=0.75\linewidth]{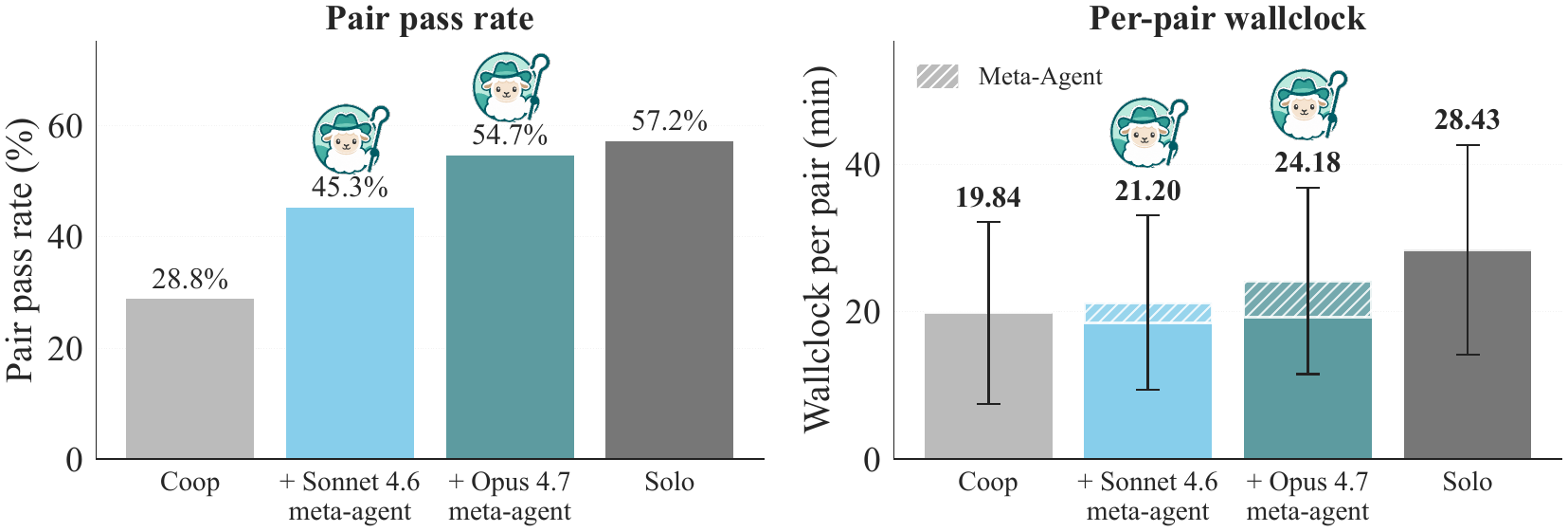}
  \caption{Runtime supervision experiments on CooperBench, with Claude Haiku 4.5 as worker. \textbf{Left (Pair pass rate):} Sonnet 4.6 and Opus 4.7 meta-agents close most of the
 gap from coop (28.8\%) to the solo ceiling (57.2\%). \textbf{Right (Per-pair wall-clock):} mean wall-clock minutes per pair. Solid bars are the worker wall-clock; the hatched overlay is the additional meta-agent overhead.\vspace{-1.0em}}
  \label{fig:live-results}
\end{figure}

\paragraph{Results.} On the full 479-pair set, the coop baseline lands at {28.8\%} pair pass rate, reproducing CooperBench's documented coordination penalty against the solo ceiling of {57.2\%}, showing a 28.4-point gap. A Sonnet meta-agent recovers {45.3\%}, and an Opus meta-agent reaches {54.7\%}, closing $91\%$ of the curse-of-coordination gap (\Cref{fig:live-results}, left). In the naive coop baseline, each worker sees only what the other chose to send, which is filtered through the sender's imperfect view of the shared state. The supervisor, by contrast, subscribes directly to both workers' effect streams, observing their actions completely without perturbing them and steering them accordingly. Appendix~\ref{app:live-detail} shows that the supervisor prefers the lightweight \texttt{inject} action while still taking \texttt{handoff} and \texttt{discard} actions when necessary.

The wall-clock cost is modest: solo takes 28.4\,min per pair on average, while the supervised conditions stay close to the parallel coop baseline (19.8\,min). Sonnet finishes in 21.2\,min (1.4\,min meta overhead) and Opus in 24.2\,min (4.3\,min meta overhead), running at roughly 75\% and 85\% of solo time respectively (\Cref{fig:live-results}, right).  This is because the meta-agent only observes worker agents once every few seconds, and does not interfere with them unless an interception is required (Appendix~\ref{app:live-detail}). We treat this as a proof of existence that \agentic{} enables effective runtime supervision; the supervision token-cost vs.\ task-execution-cost trade-off is discussed in Appendix~\ref{sec:limitations}.

\subsection{Meta-Agent for Meta-Optimization: Counterfactual Replay Optimization (\cbo{})}
\label{sec:cbo}


\textbf{Motivation.} When an agentic workflow fails, the failure usually traces
to a small set of faulty or missing agent calls~\cite{zhang2025agentracerinducingfailurellm, cemri2025multiagentllmsystemsfail}. Counterfactual analysis~\cite{wachter2017counterfactual} provides standard machinery for diagnosing such failures: would a localized change to a suspect set of calls have fixed the outcome? Answering this in a standard agentic runtime is difficult – workflow re-runs reintroduce stochastic and environmental variation unrelated to the edit, and flat transcripts give little structure to localize failures against. \cbo{} sidesteps both problems by replaying counterfactuals through forks of \agentic{}'s execution trace rather than re-executing changed workflows from scratch.

\paragraph{Method.} \cbo{} maintains a pool of agentic workflow variants together with their
\agentic{} execution traces when run on the training set. At each step, the proposer analyzes these traces to identify failure modes, picks a parent candidate and emits a set of candidate edits as counterfactual experiments to patch these failure modes. Every edit is paired with a \emph{fix set} of training
examples it should repair and a \emph{guard set} whose performance must not
regress. \agentic{} validates these edits through \emph{counterfactual replay} on
the combined fix and guard set: for each edit, it forks the parent's execution trace
at the first commit that would be affected by the edit and replays the suffix with the edited workflow. Candidates that outperform their parent on this combined set are evaluated on the dev set and added to the candidate pool; after a set number of iterations \cbo{} returns the highest-scoring member on the dev set as the selected candidate (Algorithm~\ref{alg:cbo}, Appendix~\ref{app:cbo}).

\paragraph{Setup.} We evaluate on subsets of HoVer~\cite{jiang2020hoverdatasetmanyhopfact},
MATH~\cite{hendrycks2021measuringmathematicalproblemsolving}, IFBench~\cite{pyatkin2025generalizingverifiableinstructionfollowing},
LiveCodeBench~\cite{jain2024livecodebenchholisticcontaminationfree}, and Terminal-Bench~2.0 (~\cite{merrill2026terminalbenchbenchmarkingagentshard}),
comparing \cbo{} against the baseline workflow, GEPA (optimizing workflow code)~\cite{agrawal_gepa_2025}, and
MetaHarness~\cite{lee_meta-harness_2026}. The executor is \texttt{GPT-5.4-mini} and
meta-optimizers use \texttt{GPT-5.4} (in the Codex harness for MetaHarness and GEPA). We stop each algorithm after 20 candidates on HoVer, MATH, IFBench, and LiveCodeBench, and after 10 on Terminal-Bench~2.0. Per-dataset settings, baselines, and full results are in Appendix~\ref{app:cbo}.

  \begin{figure}[h]
    \centering
    \includegraphics[width=0.85\linewidth]{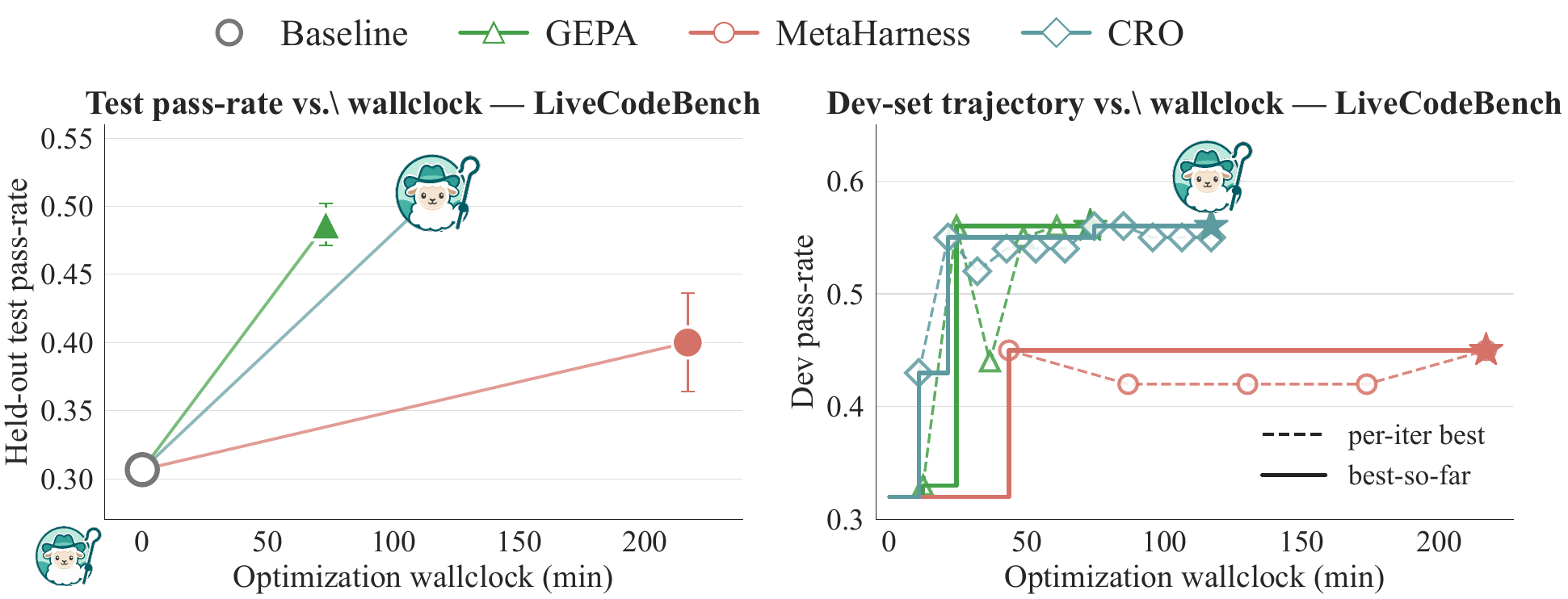}
    \caption{LiveCodeBench comparison. \textbf{Left:} held-out test pass-rate versus optimization wall-clock. \textbf{Right:} dev-set trajectory for each method across optimization wall-clock. \cbo{} subtask-cache reuse is reported separately in \Cref{fig:hotpotqa-cache-reuse}. 
    }
    \label{fig:hotpotqa-results}
  \end{figure}

\begin{table}[t]
  \centering
  \small
  \setlength{\tabcolsep}{5pt}
  \renewcommand{\arraystretch}{1.15}
  \caption{Test-set performance and meta-optimization wall-clock time across datasets
  and methods. HoVer/MATH/IFBench/LiveCodeBench report mean $\pm$ std, and Terminal-Bench~2.0 report average@5 on the test split
  (Test) over three evaluations and meta-optimization wall-clock in minutes (Wall).
  Best test mean per dataset in \textbf{bold}; second-best \underline{underlined}.}
  \label{tab:cbo}
  \newcommand{\tbcell}[1]{\makebox[\widthof{$00.0 \pm 0.0$}][c]{#1}}
  \resizebox{\textwidth}{!}{%
  \begin{tabular}{l rr rr rr rr rr}
    \toprule
    & \multicolumn{2}{c}{\textbf{HoVer}}
    & \multicolumn{2}{c}{\textbf{MATH}}
    & \multicolumn{2}{c}{\textbf{IFBench}}
    & \multicolumn{2}{c}{\textbf{LiveCodeBench}}
    & \multicolumn{2}{c}{\textbf{Terminal-Bench~2.0}} \\
    \cmidrule(lr){2-3} \cmidrule(lr){4-5} \cmidrule(lr){6-7} \cmidrule(lr){8-9} \cmidrule(lr){10-11}
    \textbf{Method}
      & \multicolumn{1}{c}{Test} & Wall
      & \multicolumn{1}{c}{Test} & Wall
      & \multicolumn{1}{c}{Test} & Wall
      & \multicolumn{1}{c}{Test} & Wall
      & \multicolumn{1}{c}{Test} & Wall \\
    \midrule
    Baseline      & $43.7 \pm 0.0$              & ---   & $60.7 \pm 1.2$              & ---   & $42.4 \pm 1.8$              & ---   & $30.7 \pm 2.1$              & ---   & \tbcell{$\underline{31.2}$} & ---   \\
    GEPA          & $43.7 \pm 0.0$              & $67$  & $74.0 \pm 3.5$              & $20$  & $50.1 \pm 1.2$              & $50$  & $\underline{48.7 \pm 1.5}$  & $73$  & \tbcell{$\underline{31.2}$} & $157$ \\
    MetaHarness   & $\underline{77.8 \pm 0.4}$  & $235$ & $\underline{79.3 \pm 1.2}$  & $101$ & $\mathbf{52.3 \pm 1.4}$     & $126$ & $40.0 \pm 3.6$              & $217$ & \tbcell{$\underline{31.2}$} & $173$ \\
    \midrule
    \rowcolor{PrimaryColor!15}
    \textbf{\cbo{}}\logonoback & $\mathbf{79.4 \pm 0.2}$     & $120$ & $\mathbf{80.0 \pm 2.0}$     & $42$  & $\underline{51.3 \pm 1.1}$  & $82$  & $\mathbf{51.0 \pm 1.7}$     & $117$ & \tbcell{$\mathbf{35.2}$}    & $73$  \\
    \bottomrule
  \end{tabular}
  }
\end{table}

\paragraph{Results.} \cbo{} obtains the best performance on four of five datasets (Table~\ref{tab:cbo}). It has higher held-out test score and lower wall-clock simultaneously compared to MetaHarness across these datasets, with savings ranging 27–58\%. Notably, on Terminal-Bench~2.0, the most execution-bound benchmark in the suite, GEPA and MetaHarness both fail to improve over the baseline on the subset under study, while \cbo{} improves performance by 4 pts while also needing the least wall-clock. On IFBench, MetaHarness edges \cbo{} by 1.0 pt on test (within a standard deviation) but still takes 37\% longer.

\begin{wrapfigure}{h}{0.4\textwidth}
\centering
\includegraphics[width=\linewidth]{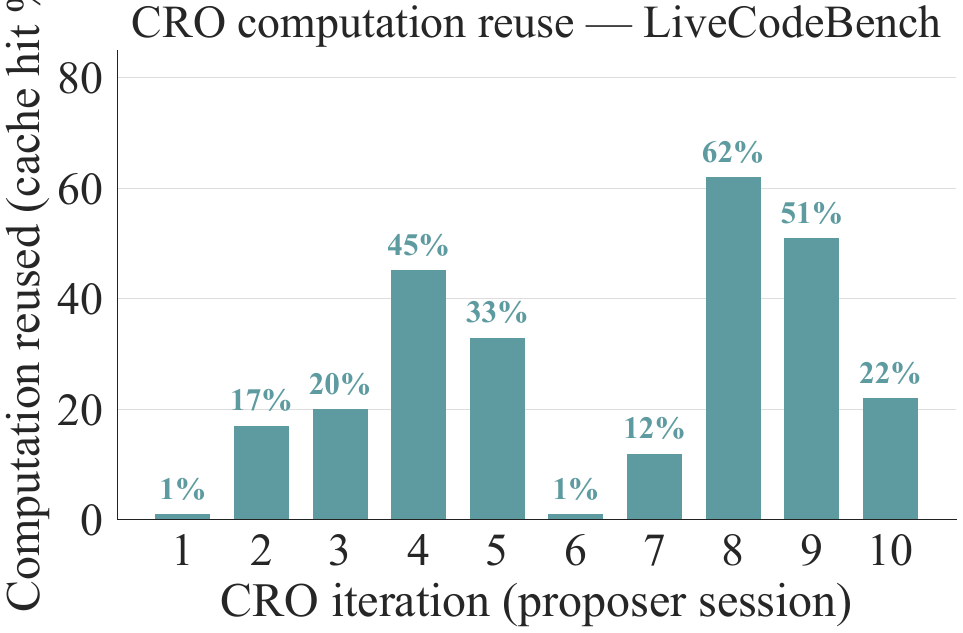}
\caption{Computation reuse on LiveCodeBench with \cbo{}.}
\label{fig:hotpotqa-cache-reuse}
\vspace{-0.8em}
\end{wrapfigure}
We attribute \cbo{}'s downstream performance gains to two sources. First, \cbo{}'s counterfactual experiments hold the unaffected part of the computation constant, letting the model isolate and patch failures, compared to the noise introduced by MetaHarness's full-pipeline reruns; we find qualitatively that the generated hypotheses are high-quality (Appendix~\ref{app:cbo}). Second, \agentic{}'s Git-like execution trace lets the LLM navigate prior executions more naturally than the flat logs used by other methods. On the other hand, \cbo{}'s wall-clock savings come from three sources: counterfactual replay re-executes only the suffix downstream of each edit's first effect via \agentic{} rather than the full pipeline, the byte-identical prefixes allow for KV-cache reuse from the LLM, and target-set gating evaluates each candidate against a small subset of the training set before committing to a full dev-set evaluation. On LiveCodeBench, computation reuse rises from $\sim$1\% on the first cold proposer session to over 60\% later in the run (Figure~\ref{fig:hotpotqa-cache-reuse}).

\subsection{Meta-Agent for Training: Meta-Agent-Guided Tree-RL}
\label{sec:training}

\paragraph{Motivation.} RLVR for long-horizon tasks suffers from sparse, episode-level rewards: an agent that takes dozens of steps receives a single binary signal at the end~\citep{chen2025contextlite, wang2025practitionersguidemultiturnagentic, tan2026hindsightcreditassignmentlonghorizon}. For fine-grained reward signals, one approach is Tree-search RL~\citep{hou2025treerlllmreinforcementlearning, ji2026treesearchllmagent}, which derives step-level advantages from outcome rewards alone via sibling rollouts. In stateless environments, this is trivial~\citep{xie2024montecarlotreesearch}. However, with agent tasks that modify real filesystems and services, exact and cheap state-forking is required to make this approach feasible, which is provided by \agentic{}.

\paragraph{Method.} GRPO-style training for long-horizon agents samples $G$ trajectories per prompt and assigns each one a single outcome reward, distributed uniformly across all actions~\citep{shao2024deepseekmathpushinglimitsmathematical}. We recover finer-grained credit by tree-searching intermediate states~\citep{sutton1998reinforcement, _wiechowski_2022}: along each root rollout, a meta-agent picks a fork turn $t^{}$ and we sample $K$ sibling branches forward from that state, yielding $G(K{+}1)$ trajectories per task at the cost of only $K$ extra branch rollouts (forking is exact and cheap, \Cref{tab:framework-perf}). Credit assignment operates at two levels: prefix actions before $t^{}$ inherit the standard inter-root GRPO advantage across the $G$ roots, while suffix actions take an intra-tree advantage computed within each $(K{+}1)$-member fork group, surfacing per-step outcome differences without a learned value function or process reward model. The full procedure is given as Algorithm~\ref{alg:mcts-rl} in Appendix~\ref{app:training-config}, with qualitative examples of the meta-agent's branching decisions in \Cref{app:meta-agent-qualitative}.

\paragraph{Setup.} We use both \texttt{Qwen3.5-35B-A3B} and \texttt{Nemotron-3-Super-120B-A12B} as the base models~\citep{qwen3.5, nvidia_nemotron_3_2025}, training via tinker~\citep{tml2025tinker}. We train on a filtered subset of the \textbf{Endless Terminals} corpus~\citep{gandhi_endless_2026}: starting from the 2{,}492-task pool, we drop tasks where the base policy passes all 8 sampled rollouts (pass@8=1.0), leaving 442 tasks for Qwen3.5 and 530 for Nemotron-3. We hold out \textbf{Terminal-Bench~2.0}~\citep{merrill2026terminalbenchbenchmarkingagentshard} as an out-of-distribution test set evaluated every 10 training steps. We compare two RL setups at matched generation compute: \textbf{Flat GRPO}, independent root rollouts only, one group baseline across $G{=}8$ roots; and \emph{Meta-Agent Guided Tree-RL} (\textbf{Tree-GRPO}): root rollouts plus $K{=}4$ sibling branches forked at a meta-agent-chosen turn, with the two-level baseline from \Cref{alg:mcts-rl}. The full configuration for training is deferred to \Cref{app:training-config}.


\paragraph{Results.}
\begin{wraptable}{r}{0.55\textwidth}
\vspace{-1.5em}
\caption{Held-out Terminal-Bench~2.0 avg@5 (89 tasks, 5 seeds). Settings in \Cref{app:training-config}.}
\label{tab:tb2-final}
\centering\footnotesize
\resizebox{\linewidth}{!}{
\begin{tabular}{lcc}
\toprule
\textbf{Method} & \textbf{Qwen3.5-35B-A3B} & \textbf{Nemotron-3-Super-120B-A12B} \\
\midrule
Base                  & 26.1\%$_{\pm 4.21}$ & 30.3\%$_{\pm 3.62}$ \\
Flat GRPO             & 34.2\%$_{\pm 4.05}$ & 33.8\%$_{\pm 3.41}$ \\
\rowcolor{PrimaryColor!15}
\textbf{Tree-GRPO}\logonoback  & \textbf{39.4\%}$_{\pm 3.87}$ & \textbf{37.2\%}$_{\pm 3.19}$ \\
\bottomrule
\end{tabular}
}
\vspace{-0.8em}
\end{wraptable}

The results are in \Cref{tab:tb2-final}. Meta-Agent Guided Tree-RL is the strongest setting on both base models. First, the uplift of Tree-GRPO over Flat GRPO is consistent across model scale: a $\sim$3B-active-parameter and a $\sim$12B-active-parameter model gain $5.2$ and $3.4$ points respectively. This improvement is the consequence of a training-time pattern: Tree-GRPO's produces higher reward variance on both models (\Cref{fig:mcts-curves}). Flat GRPO assigns each action in a trajectory the same outcome-derived advantage. Tree-GRPO's intra-tree baseline instead isolates the suffix actions taken after the fork point: $K+1$ siblings sharing a prefix differ only in what happened after the fork point, so the per-step advantage there reflects local choice quality rather than global trajectory outcome. \agentic{} makes this affordable, and we find that the intra-tree baseline produces more informative gradient steps as well. This behavior transfers to the Endless Terminals validation split (\Cref{fig:mcts-ood}, deferred to \Cref{app:training-config}).






\section{Conclusion}
\label{sec:conclusion}
We presented \agenticlogo{}, a substrate that lets a meta-agent hold, inspect, fork, and modify another agent's execution as a first-class object, like functions in functional programming. Meta-agents in \agentic{} are higher-order agents over agentic execution, as we show across three applications: runtime supervision of parallel coding agents, counterfactual replay for workflow optimization, and Tree-GRPO for finer-grained credit assignment in long-horizon RL. More broadly, \agentic{} opens a path toward meta-agent systems that are active operators over agentic execution. Future agents could use the substrate to compress execution traces into reusable workflows, run counterfactual agentic interpretability probes, safely gate irreversible actions, or learn policies for managing agentic execution. As agentic systems become longer-lived, more stateful, and more consequential, we believe this execution-level control will become a core abstraction for effective meta-agents. \agentic{} is a step toward making that abstraction readily available and programmable.

\bibliographystyle{plainnat}
\bibliography{references}
\appendix


\section{Limitations and Future Works}
\label{sec:limitations}

\subsection{Limitations}\label{app:limitations}

\paragraph{Proof-of-existence framing.} Each of our three case studies is reported as a proof of existence: the substrate primitives suffice to drive a meaningful uplift on a representative dataset, given a meta-agent we wrote for that case. We do not claim optimality of the meta-agent policies, robustness across model families and benchmarks at scale, or that the headline numbers cannot be matched without \agentic{}. Establishing a full burden of proof, in the sense of head-to-head comparisons against every plausible alternative substrate and policy, is outside the scope of this work, which primarily serves to introduce \agentic{} as a framework.

\paragraph{Supervision and proposer cost.} The live-supervision and \cbo{} results assume access to a meta-agent strong enough to act usefully (Sonnet~4.6 / Opus~4.7 in \S\ref{sec:during}; GPT-5.4 in \S\ref{sec:cbo}). For short tasks, the meta-agent's token cost can exceed the worker's, just as for existing meta-optimisers~\citep{agrawal_gepa_2025, lee_meta-harness_2026}. We report total dollar cost per arm in Appendix~\ref{app:\cbo{}-perdataset}; the regime where this trade-off is favourable depends on task length and on the cost ratio between the worker and the meta-agent.

\paragraph{Counterfactual replay assumes weak coupling between edits and side effects.} \cbo{} replays a candidate edit's suffix from the first event whose dependencies are affected by the edit. When an edit touches a component whose effects propagate widely (e.g.\ the system prompt of a tool used in every step), the suffix is the entire trajectory and the cache buys nothing. We observe this regime on the cold first proposer session of every dataset (Appendix~\ref{app:\cbo{}-perdataset}); it amortises away within two to three sessions on the benchmarks we study.

\subsection{Future Works}\label{app:future_works}

\paragraph{Agent interpretability.} Mechanistic interpretability of agentic systems today is largely observational: a transcript is annotated, a probe is fit, conclusions are drawn. \agentic{}'s coupled fork turns these into testable counterfactual interventions. Editing a single component (a tool, a system-prompt span, a sub-task definition) and replaying the suffix from the first commit it affects holds every other source of variance fixed, isolating the contribution of that component to the eventual outcome. \cbo{}'s propose-and-replay machinery is one instance of this loop driven by a task-success objective; a natural next step is to run the same loop under an interpretability objective, e.g.\ minimal edits that flip a specific decision, or the smallest prompt span whose removal preserves task success, turning \cbo{} into a tool for explanation rather than optimisation.

\paragraph{Reversible environments for continual learning.} The Tree-GRPO results in \S\ref{sec:training} work because forking the agent's filesystem and processes is essentially free. The same primitive applies to any domain whose state lives on disk or in a sandboxed process tree: computer-use, web browsing, and large-codebase edit tasks all fit. Once forks are cheap, a single agent can train over weeks of interaction rather than a single rollout episode, with the execution trace maintaining a coherent history of every revisit, every backtrack, and every retried tool call. Continual learning then becomes a question of scheduling reads against this graph rather than re-engineering the substrate.

\paragraph{Post-training agents to use the substrate.} The case studies in \S\ref{sec:results} fix the meta-agent and let it govern a base worker. The dual question is whether the worker itself can be post-trained to use \agentic{}'s primitives natively: an action space that includes \texttt{fork}, \texttt{discard}, and \texttt{replay} over its own typed effect stream, not just tool calls into the environment. This is strictly harder than tool-use post-training because the model has to learn when to back up, when to retry, and when to spawn a sibling, not just which tool to invoke. The substrate provides exactly the typed state needed to define these actions and to compute exact returns over them.

\paragraph{Reversible sandboxes as a safety property.} Every effect on the substrate carries a reversibility tier (\S\ref{sec:effect}): reversible filesystem mutations, compensable side effects, and irreversible external calls each have a different rollback contract. This makes \agentic{} a candidate substrate for safety-critical agent deployments: an external override can preempt the materialisation of any irreversible effect, downstream checks can fire a rollback up to the last materialisation point, and the execution trace gives auditors a content-addressed record of what actually happened rather than what the agent reported. The substrate does not solve the alignment problem, but it makes the standard halt-and-inspect loop something a deployment system can implement against a stable interface rather than bespoke per-agent plumbing.

\section{Mechanized Core and Proof Envelopes}
\label{app:lean_proof}

\agentic{} separates the production runtime from the semantic object that is
mechanized in Lean. The production framework executes ordinary Python tasks,
provider SDK calls, shell commands, sandbox operations, retries, scheduling, and
carrier storage. Those executions are not themselves verified. The verified
artifact is a small algebraic-effects trace machine, together with proof-backed
profiles for static fragments whose lowered traces fall inside its boundary.

\paragraph{Claim tiers.} Each inspectable run may carry a proof envelope with
a profile and an explicit strength. The profile is one of
\texttt{runtime\_only},
\texttt{reference\_core\_a},
\texttt{core0},
\texttt{core\_a},
\texttt{core0h}, or
\texttt{extension}; the strength is one of
\texttt{runtime\_only},
\texttt{reference\_validated},
\texttt{forward\_simulation}, or
\texttt{semantic\_adequacy}. Ordinary Python runs default to
\texttt{runtime\_only}. A trace becomes \texttt{reference\_core\_a} when it is
accepted by the executable kernel-v3 reference validator but does not yet claim
Lean theorem coverage. A trace becomes \texttt{core0} or \texttt{core\_a} only
when the static lowering evidence, generated-trace validation, and Lean-side
fragment assumptions all match the completed generated trace. Core-0H is
sidecar-gated: it can receive a \texttt{forward\_simulation} envelope only when
a validated content-addressed Core-0H sidecar manifest is attached, and the
classifier does not infer it from arbitrary structured handlers. Core-0/Core-A envelopes claim
\texttt{semantic\_adequacy}. Core-A proof-backed envelopes currently require
exactly one selected direct abort capture, no selected abort-path resume/return,
and no selection-closure suffix; handler-side effects, sequencing before abort,
multiple abort captures, or unused abort-only handler definitions remain
reference-validatable. Incomplete or live prefixes remain reference-validatable
unless a future prefix certificate is attached.
Publication controls such as
forwarding, terminal delay/fork, and replay remain \texttt{extension} unless a
future proof envelope names a stronger theorem.

\paragraph{Lean theorem surface.} The Lean development builds with
\texttt{lake build} in the kernel-v3 proof artifact. The current theorem surface
used by the proof envelope is:

\begin{table}[h]
\caption{Mechanized theorem surface used by the proof-envelope claim.}
\label{tab:lean-theorem-surface}
\centering\scriptsize
\begin{tabular}{p{0.21\linewidth}p{0.36\linewidth}p{0.33\linewidth}}
\toprule
\textbf{Profile} & \textbf{Representative Lean theorem} & \textbf{Meaning} \\
\midrule
Core-0 & \texttt{source\_eval\_to\_machine};
\newline\texttt{core0\_machine\_}\newline\texttt{eval\_to\_source} &
Forward source-to-machine simulation and restricted reverse simulation for the
ordinary callable-resumption fragment. \\
Core-A & \texttt{source\_eval\_to\_machine};
\newline\texttt{coreA\_machine\_}\newline\texttt{eval\_to\_source} &
Core-0 plus the direct abort-without-resume handler boundary currently admitted
by the envelope. \\
Core-0H & \texttt{core0h\_source\_}\newline\texttt{eval\_to\_machine} &
Forward simulation for deterministic two-phase handler bodies with matching
evidence; this profile is currently forward-only. \\
Trace monotonicity & \texttt{trace\_monotonic};
\texttt{core0h\_trace\_monotonic} &
Machine execution extends traces by appending records rather than rewriting
prior trace prefixes. \\
Branch replay skeleton & \texttt{single\_child\_branch\_}\newline\texttt{replay\_sound} &
One-child structural replay soundness for an exact suffix replay model; this is
not a production replay refinement proof. \\
\bottomrule
\end{tabular}
\end{table}

\paragraph{Executable reference boundary.} The Python
\texttt{agentic-kernel-v3-reference} package in the submitted code package sits between the production runtime
and the Lean development. It validates Core-0/Core-A trace lifecycles, reruns a
static \texttt{KernelProgram} to check exact generated-trace agreement, and
emits an explicit \texttt{ProofEnvelope}. Public \texttt{Run[T]} values carry a
\texttt{proof} field; current production Python executions default to
\texttt{runtime\_only}. The envelope records a content-addressed evidence
identifier over the proof authority, validator, program digest, and trace
digest. Kernel envelopes derive \texttt{proof\_backed} from
\texttt{proof\_strength}, so profile names alone do not upgrade a trace. Runtime
metadata is carrier metadata: non-runtime strengths must cite kernel-v3
reference provenance and a \texttt{proof-evidence:sha256} identifier, but public
\texttt{Run[T]} metadata exposes such imports as \texttt{claimed\_proof\_backed}
rather than runtime-verified proof authority. Lean theorem ids are centralized in
a proof-surface ABI table checked by a Lean module with \texttt{\#check} commands
and typed signature wrappers, so stale theorem names or materially changed
theorem statements fail the artifact build. Live prefixes and structured handler
bodies without validator-issued Core-0H two-phase sidecar manifests are
classified as reference-validatable rather than proof-backed. The artifact gate
is \texttt{make verify-proof-envelope-claim}. This makes the formal claim
inspectable without implying that all executable Agentic programs are
proof-backed.

\paragraph{Non-claims.} The proof envelope does not verify arbitrary Python
control flow, provider SDK behavior, model outputs, prompt-cache state, shell
commands, filesystem mutation correctness, Docker or sandbox implementations,
meta-git carrier storage, scheduling, cancellation, retries, recovery, or
multi-branch replay. The meta-agent applications in the main paper rely on the
production substrate plus empirical validation; the Lean artifact supplies the
semantic core and the boundary discipline for proof-backed fragments.

\section{Framework Performance: Extended Results}
\label{app:framework-perf-extended}

This appendix groups all extended results that support \Cref{sec:microbench}: the measurement protocol (Appendix~\ref{app:microbench-protocol}), the mutation-size sweep that defuses the large-file concern (Appendix~\ref{app:mutation-sweep}), the agent-turn latency reference behind the ``2--3\% of one turn'' claim (Appendix~\ref{app:agent-latency}), depth scaling under stacked overlay layers (Appendix~\ref{app:scaling}), the effect-stream observation overhead (Appendix~\ref{app:observe-detail}), the KV-cache reuse breakdown (Appendix~\ref{app:kvcache-detail}), and the cross-backend portability check (Appendix~\ref{app:backends}).

\subsection{Measurement Protocol}
\label{app:microbench-protocol}

\paragraph{Hardware.} The \Cref{tab:framework-perf} measurements for \agentic{}, \texttt{docker commit}, and full rootfs copy run on the same Vultr cloud instance (2~vCPU, 16\,GB RAM, SSD, Ubuntu~22.04, Docker~29.3.1, overlay2 storage driver on extfs). Modal numbers come from Modal's hosted gVisor runtime (separate hardware) and are reproduced from a fresh benchmark whose latency, therefore, additionally includes network round-trip from the host-side agent. The \Cref{tab:scaling} measurements use E2B Firecracker micro-VMs. Cross-backend numbers (Appendix~\ref{app:backends}) use E2B (Firecracker), Modal (gVisor), and Daytona (managed Linux containers).

\paragraph{Workloads.} \Cref{tab:framework-perf} uses three real Terminal-Bench~2.0 Docker images spanning two orders of magnitude: \emph{openssl-selfsigned-cert} (42\,MB), \emph{caffe-cifar-10} (200\,MB), and \emph{pytorch-model-recovery} (5.8\,GB). Full copy tars the entire container rootfs, excluding \texttt{/proc}, \texttt{/sys}, \texttt{/dev}, and \texttt{/tmp}, and is $O(n)$ in image size. \texttt{docker commit}, Modal \texttt{snapshot\_filesystem()}, and \agentic{}'s overlay delta are all $O(1)$ in image size on the overlay2 driver. \Cref{tab:scaling} additionally uses three synthetic working directories: \emph{small} (10 files, 30\,KB), \emph{medium} (100 files, 100\,MB), \emph{large} (100 files, 1\,GB). KV-cache experiments use real Terminal-Bench~2.0 tasks with a Haiku~4.5 agent.

\paragraph{Per-step mutation pattern.} The agent's writes inside the workdir are simulated with a fixed 5+3 random-file pattern: 5 baseline writes establish the parent state \emph{before} the checkpoint, then each of the $K{-}1{=}3$ sibling branches adds 1 more random write \emph{after} the fork, all written into the measured workdir. Each write defaults to 10,KB. \Cref{tab:framework-perf}'s storage columns count only the post-fork branch deltas: the \textbf{Storage} column reports a single sibling's overlay delta (1 write = 10,KB); the \textbf{Disk @ K=4} column reports the sum of the three siblings' overlay deltas ($3\times10,\textrm{KB}=30,\textrm{KB}$). The 5 baseline writes (50,KB) sit in the parent's overlay and are not counted there. \Cref{tab:mutation-sweep} in \Cref{app:mutation-sweep} reports the \emph{full} overlay upper directory (baseline + branches, $8\times10,\textrm{KB}\approx80,\textrm{KB}$ at this default size) and sweeps the per-write size from 1,KB to 100,MB to verify the substrate scales 1:1 with what the agent writes and adds no multiplier on top.

\paragraph{Pattern A: agent-on-host, sandbox-in-container.} The fork/revert latencies in \Cref{tab:framework-perf} measure agent-perceived wall-clock under Pattern A, which matches how production agents (mini-swe-agent, SWE-Agent, smolagents) use sandboxes today. A Python agent process on the host imports \texttt{litellm} + \texttt{tenacity} (matching Terminus-2's startup), warms an OpenAI \texttt{gpt-5.4-mini} client with one call, then drives the sandbox container through a tmux session. Each operation is timed from "operation issued" until the new sandbox's tmux returns to a probe bash command (\texttt{echo READY\_<uuid>} + \texttt{capture-pane} until the marker appears). Method-specific operations:
\begin{itemize}[noitemsep,topsep=0pt,leftmargin=*]
    \item \texttt{docker commit}: Fork = \texttt{docker commit} + \texttt{docker run -d} new container + tmux ready in new container. Revert = \texttt{docker rm -f} + \texttt{docker run -d} from saved image + tmux ready. Both ops are required; measuring only \texttt{docker run} understates by the rm/commit cost.
    \item Modal: Fork = \texttt{snapshot\_filesystem()} + \texttt{Sandbox.create(image=snap)} + tmux ready. Revert = \texttt{terminate()} + \texttt{Sandbox.create(image=snap)} + tmux ready.
    \item \agentic{}: Fork = overlay-layer fork (in-container, same tmux session). Revert = overlay umount/remount of saved layer (in-container, same tmux session).
    \item BranchFS~\citep{wang_fork_2026}: BranchFS daemon runs on the host with a FUSE mount; the TB2 image is bind-mounted into a Docker container at \texttt{/workspace} so the agent's tmux can read/write through the FUSE layer. Fork = \texttt{branchfs create}, which creates a copy-on-write branch and auto-switches the mount, then tmux probe. Revert = \texttt{branchfs abort}, which discards the leaf branch's delta and switches the mount back to the parent, then tmux probe. BranchFS branch ops are O(1) in base directory size, so fork/revert latency is image-independent (confirmed across 42\,MB / 200\,MB / 5.8\,GB images: \textasciitilde{}270\,ms fork, \textasciitilde{}358\,ms revert).
    \item Full copy: latency reported in \Cref{tab:framework-perf} is computed (not measured) as rootfs \texttt{tar}/\texttt{restore} (Vultr) plus an empirically observed $\sim$340\,ms container-startup + tmux-probe overhead. Running the full agent-revert protocol takes hours per cell on the 5.8\,GB image (each fork moves the full rootfs tar across two \texttt{docker cp} boundaries), so we report the computed values.
\end{itemize}

{\paragraph{Disk and RAM at $K{=}4$.} Disk @ K=4 is the host disk delta over $K{-}1$ added branches: commit-layer deltas for \texttt{docker commit}, full rootfs tars for \texttt{full copy} (computed = $(K{-}1) \times$ per-branch tar size), overlay-layer files for \agentic{}, BranchFS storage tree (\texttt{/var/lib/branchfs}) over $K$ branches. RAM @ K=4 is summed RSS via \texttt{docker stats --no-stream} over the alive containers ($K$ for \texttt{docker commit} / \texttt{full copy}, where each branch needs a separate restarted container; 1 container reading for \agentic{}, which forks the worker process per branch but reuses one host container as the overlay-layer host (the K forked worker processes are aggregated under that one RSS reading); 1 container + BranchFS daemon RSS for BranchFS). Modal's per-sandbox host RAM/disk is hidden behind Modal's runtime, so its $K{=}4$ cells are marked ---.}

\paragraph{Protocol.} Each benchmark begins with 2--3 warm-up iterations (discarded) to prime filesystem caches and JIT paths. Measurements are wall-clock \texttt{time.monotonic()} around the operation under test. For \Cref{tab:framework-perf} fork/revert latency: 10 repetitions per cell for \agentic{} / \texttt{docker commit} / Modal, 5 repetitions for full-copy storage measurements; 3 repetitions for the $K{=}4$ resource columns. Error bars are $\pm 1\sigma$. Storage is measured via \texttt{du~-sb} on the overlay upper directory (for \agentic{}) or the checkpoint artifact (for baselines). Bench code is shared under \texttt{exp/framework-perf}, please refer to the readme.


\subsection{Substrate scaling under varying mutation size}
\label{app:mutation-sweep}

A natural reviewer concern about the 10\,KB delta in \Cref{tab:framework-perf} is whether the substrate stays small when the agent writes large files. We sweep the per-step write size from 1\,KB to 100\,MB on the 5.8\,GB pytorch-model-recovery image, keeping the same 5+3 mutation pattern but changing how many bytes each write puts into the workdir. \Cref{tab:mutation-sweep} reports the resulting overlay disk delta and fork latency across three repetitions per size.

Two findings: (a) the disk delta tracks what the agent writes 1:1 (8\,KB at 1\,KB/step up to 800\,MB at 100\,MB/step), so the substrate adds zero overhead beyond the agent's actual emissions; (b) fork and revert latency stay flat at $\sim$340\,ms across all sizes, since both operations only swap overlay metadata, not file contents. We additionally verify revert correctness across the same sweep: in 12 reps spanning all four sizes, the workdir after revert exactly matches the pre-fork state (5 baseline files preserved, 3 post-fork branch files cleanly discarded; 12/12 PASS). Absolute latencies here are about 200\,ms higher than \Cref{tab:framework-perf} because this sweep ran on Mac/Docker Desktop's Linux VM, which adds \texttt{docker exec} overhead per call; the scaling shape is what the table claims.

\begin{table}[h]
\caption{Substrate scaling under varying per-step mutation size on the 5.8\,GB pytorch-model-recovery image (5+3 random-write pattern, $K{=}4$ branches, 3 reps; medians). The disk delta scales 1:1 with what the agent writes; fork and revert latency stay flat. Revert correctness (post-revert workdir matches pre-fork state) is 12/12 PASS across the four sizes.}
\label{tab:mutation-sweep}
\centering\small
\begin{tabular}{rrrrr}
\toprule
\textbf{Per-step write} & \textbf{Total written} & \textbf{Disk delta $\downarrow$} & \textbf{Fork (ms) $\downarrow$} & \textbf{Revert (ms) $\downarrow$} \\
\midrule
1\,KB    & 8\,KB    & 8.4\,KB    & 339 & 348 \\
10\,KB   & 80\,KB   & 80.4\,KB   & 386 & 346 \\
1\,MB    & 8\,MB    & 8.0\,MB    & 331 & 385 \\
100\,MB  & 800\,MB  & 800\,MB    & 343 & 366 \\
\bottomrule
\end{tabular}
\end{table}


\subsection{Agent per-turn latency reference}
\label{app:agent-latency}

To anchor the claim that \agentic{}'s fork is small relative to a typical agent turn, we instrumented the bench harness with per-turn wall-clock timing and ran two Terminal-Bench~2.0 tasks for up to 20 turns each (Anthropic Claude Haiku~4.5, E2B sandbox). Across 17 measured turns we observe:

\begin{itemize}[noitemsep,topsep=0pt,leftmargin=*]
    \item \textbf{LLM call} (Anthropic API round-trip): mean 5.36\,s, median 5.56\,s, p10/p90 = 2.69 / 7.64\,s.
    \item \textbf{Tool call} (sandbox bash execution): mean 0.20\,s, median 0.17\,s, p10/p90 = 0.08 / 0.61\,s.
    \item \textbf{Per-turn total}: mean 5.50\,s, median 5.81\,s, p10/p90 = 2.70 / 7.77\,s.
\end{itemize}

The LLM call dominates (~98\% of per-turn wall-clock); the tool call is small here because Terminal-Bench tasks involve mostly light bash (file reads, package installs). Heavier tool actions (compilation, training) push per-turn time well above the median. \agentic{}'s 134--143\,ms fork (\Cref{tab:framework-perf}) is therefore around 2--3\% of a typical Haiku 4.5 turn and well below the noise floor of one LLM call's response-time variance.


\subsection{Scaling Behaviour}
\label{app:scaling}

\Cref{tab:scaling} reports how checkpoint/revert latency behaves as the number of stacked overlay layers grows (left) and as the effect stream accumulates events (right). \agentic{}'s overlay checkpoint remains in the 157--252\,ms band (on E2B) through 50 stacked layers; \texttt{docker commit} is roughly constant as well but at a 2.8$\times$ higher baseline (451--558\,ms). The OverlayFS lower-directory chain is bounded by the kernel's page-size limit at approximately 60~layers; trajectories exceeding this depth require periodic compaction of frozen layers.

Per-event effect-stream overhead (record and observe) is constant at approximately 120\,ms on E2B (network-dominated) through 200~steps. Stream size grows linearly at ${\sim}130$\,B/event.

\begin{table}[h]
\caption{\textbf{Left:} Checkpoint/revert latency (ms) as overlay layers stack (E2B), compared to \texttt{docker commit} (Vultr). \textbf{Right:} Per-event effect-stream overhead (E2B) as trajectories grow.}
\label{tab:scaling}
\centering\small
\begin{minipage}[t]{0.52\textwidth}
\centering
\begin{tabular}{rrr|rr}
\toprule
& \multicolumn{2}{c|}{\textbf{Agentic}} & \multicolumn{2}{c}{\textbf{Docker commit}} \\
\textbf{Depth} & \textbf{Ckpt} & \textbf{Revert} & \textbf{Commit} & \textbf{Revert} \\
\midrule
1  & 173 & 255 & 451 & 300 \\
5  & 157 & 170 & 524 & 255 \\
10 & 252 & 167 & 481 & 258 \\
25 & 179 & 170 & 558 & 260 \\
50 & 237 & 167 & 503 & 297 \\
\bottomrule
\end{tabular}
\end{minipage}
\hfill
\begin{minipage}[t]{0.44\textwidth}
\centering
\begin{tabular}{rrrr}
\toprule
\textbf{Steps} & \textbf{Record} & \textbf{Observe} & \textbf{Stream} \\
\midrule
1   & 106\,ms & 107\,ms & 134\,B \\
10  & 127\,ms & 178\,ms & 1.3\,KB \\
50  & 109\,ms & 177\,ms & 6.5\,KB \\
100 & 197\,ms & 117\,ms & 13.1\,KB \\
200 & 171\,ms & 115\,ms & 26.4\,KB \\
\bottomrule
\end{tabular}
\end{minipage}
\end{table}

\subsection{Observe Overhead Detail}
\label{app:observe-detail}

\Cref{tab:observe} compares effect-stream recording throughput on a local Docker host (no network roundtrip) versus E2B Firecracker (remote API). The local overhead is 3.1\,ms per event (5\%); the E2B figure (113\,ms, 87\%) is dominated by the network roundtrip for each \texttt{exec} call and does not reflect framework serialization cost. We separately verified that subscribing a supervisor to the effect stream adds exactly zero tokens to the worker's context by comparing the worker's message list with and without a supervisor attached: the two lists are byte-identical across a 10-step trajectory.

\begin{table}[h]
\caption{\textbf{Left:} Effect-stream recording throughput, local vs.\ remote. \textbf{Right:} Context inflation test: supervisor subscription adds 0 tokens to the worker.}
\label{tab:observe}
\centering\small
\begin{minipage}[t]{0.52\textwidth}
\centering
\begin{tabular}{lrr}
\toprule
\textbf{Metric} & \textbf{Local} & \textbf{E2B} \\
\midrule
Raw throughput       & 17\,evt/s & 8\,evt/s \\
Logged throughput    & 16\,evt/s & 4\,evt/s \\
Overhead / event     & 3.1\,ms (5\%) & 113\,ms (87\%) \\
Observe latency      & 64\,ms & 104\,ms \\
\bottomrule
\end{tabular}
\end{minipage}
\hfill
\begin{minipage}[t]{0.44\textwidth}
\centering
\begin{tabular}{lr}
\toprule
\textbf{Condition} & \textbf{Worker context} \\
\midrule
Without supervisor & 21 msgs, 1449 chars \\
With supervisor    & 21 msgs, 1449 chars \\
\addlinespace
Context inflation  & \textbf{0 chars (0.0\%)} \\
\bottomrule
\end{tabular}
\end{minipage}
\end{table}


\subsection{KV-Cache Reuse Detail}
\label{app:kvcache-detail}

\Cref{tab:kv-K-sweep-haiku} reports the per-task, per-fork-depth view behind the §3 summary. Each task is run once on the Anthropic API with Claude Haiku 4.5 to generate an initial trajectory, with checkpoints saved at fork depths step~10, step~25, and step~50 (the third only fires when the trajectory reaches that depth). At each saved fork point we then run $K$ branches: each branch reverts the sandbox to the checkpoint, restores the LLM message prefix with \texttt{cache\_control: \{"type": "ephemeral"\}} on the last prefix message, and continues sampling. The provider's prompt cache (5-minute TTL) charges $0.10\times$ the input-token rate for resolved-prefix tokens and $1.25\times$ the rate for the first branch's cache write; branches $2$ through $K$ amortise the write across additional reads, which is why savings climb sharply $K{=}1 \to K{=}2$ and stabilise after.

\begin{table}[h]
\caption{Per-task KV-cache reuse on the Anthropic API (Claude Haiku 4.5) across 8 Terminal-Bench~2.0 tasks. Each cell reports \emph{savings\%~/~hit\%} for $K$ branches forked and replayed from a checkpoint at the listed step depth. The hit rate is the substrate-fidelity check (does \texttt{revert} restore the LLM message prefix byte-for-byte); the plateau at $\sim$95\% from $K{=}2$ onwards is within 5\% of the 100\% ceiling. Savings climb sharply $K{=}1 \to K{=}2$ as the cache-write penalty amortises across one extra branch, then stabilise as per-branch suffix generation grows linearly in $K$. Empty cells are fork-depth/$K$ combinations not reached within the per-task wall-clock budget.}
\label{tab:kv-K-sweep-haiku}
\centering\small
\begin{tabular}{lr ccccc}
\toprule
& & \multicolumn{5}{c}{\textbf{Branching factor $K$}} \\
\cmidrule(lr){3-7}
\textbf{Task} & \textbf{Fork} & $K{=}1$ & $K{=}2$ & $K{=}4$ & $K{=}8$ & $K{=}16$ \\
\midrule
openssl-selfsigned-cert            & step 10 & 61 / 83 & 72 / 96 & 70 / 95 & 72 / 96 & 71 / 95 \\
                                   & step 25 & 60 / 79 & 79 / 98 & 79 / 97 & 79 / 98 & 80 / 98 \\
\addlinespace
nginx-request-logging              & step 10 & 59 / 87 & 61 / 93 & 63 / 93 & 62 / 93 & 62 / 93 \\
                                   & step 25 & 66 / 88 & ---     & ---     & ---     & ---     \\
\addlinespace
build-cython-ext                   & step 10 & 60 / 87 & 67 / 95 & 67 / 95 & 67 / 95 & 67 / 94 \\
                                   & step 25 & 68 / 89 & 78 / 97 & 77 / 97 & 77 / 97 & ---     \\
\addlinespace
configure-git-webserver            & step 10 & 62 / 88 & 68 / 94 & 67 / 94 & 68 / 95 & 68 / 95 \\
                                   & step 25 & 62 / 82 & 76 / 97 & 79 / 97 & 77 / 97 & ---     \\
\addlinespace
feal-differential-cryptanalysis    & step 10 & 57 / 86 & 63 / 93 & 63 / 93 & 63 / 93 & ---     \\
\addlinespace
llm-inference-batching-scheduler   & step 10 & 55 / 83 & 63 / 92 & 63 / 92 & 64 / 93 & 64 / 93 \\
\addlinespace
make-doom-for-mips                 & step 10 & 56 / 84 & 68 / 93 & 64 / 93 & 59 / 91 & 62 / 92 \\
                                   & step 25 & 64 / 88 & 73 / 96 & 75 / 97 & 74 / 97 & 74 / 97 \\
                                   & step 50 & 74 / 89 & 84 / 99 & 85 / 99 & 84 / 99 & ---     \\
\addlinespace
pytorch-model-recovery             & step 10 & 58 / 86 & 65 / 93 & 64 / 92 & 64 / 92 & 64 / 93 \\
                                   & step 50 & 67 / 85 & 82 / 98 & 82 / 98 & ---     & ---     \\
\midrule
\textbf{Mean}                      &         & \textbf{62 / 86} & \textbf{71 / 95} & \textbf{71 / 95} & \textbf{70 / 95} & \textbf{68 / 94} \\
\bottomrule
\end{tabular}
\end{table}

All Anthropic measurements use \texttt{cache\_control: \{"type": "ephemeral"\}} on the last prefix message; the provider's prompt cache (5-minute TTL) serves the prefix at 10\% of the normal input-token price. Tasks whose initial-prompt prefix falls below Haiku~4.5's 4{,}096-token minimum cacheable threshold are excluded; in our Terminal-Bench~2.0 sample this filter drops one task (\emph{fix-git}, 157-character instruction).


\subsection{Realization across Sandbox Backends}
\label{app:backends}

The primitives of \Cref{sec:primitives} are realized over the overlay-filesystem and checkpoint facilities exposed by modern containerized sandboxes. A single device-layer interface abstracts backend differences; application code written against the abstraction runs unchanged across providers. \Cref{tab:backends} summarizes compatibility and measured fork latency where available.

\paragraph{Docker (local / Vultr).} Privileged containers with kernel OverlayFS on tmpfs. Checkpoint unmounts the overlay, freezes the upper directory as a named layer, and remounts with the frozen layer in the lower stack. Measured at 72\,ms median (50 reps, 2~vCPU / 4\,GB).

\paragraph{E2B Firecracker.} Micro-VM sandboxes with OverlayFS via \texttt{sudo}. Semantics are identical to local Docker; measured latency is higher (159--169\,ms) due to the remote API roundtrip. The \texttt{metacopy=on} mount option avoids full-file copy-up on \texttt{chown}.

\paragraph{Modal (gVisor).} gVisor blocks \texttt{mount}/\texttt{umount} syscalls, so checkpoint uses Modal's \texttt{snapshot\_filesystem()} API (935--1137\,ms). Revert terminates the sandbox and spawns a new one from the snapshot image (75--79\,ms).

\paragraph{Daytona.} Cloud development environment with root access. OverlayFS works without \texttt{sudo}. Preliminary validation confirms all scope operations pass. Since the underlying primitive is identical to local Docker's OverlayFS (72\,ms, size-independent in our \texttt{local\_overlay} bench), Daytona's measured fork latency is dominated by the remote API roundtrip; we estimate $\sim$150\,ms by analogy with E2B's measured +91\,ms RTT (\Cref{tab:backends}).

\paragraph{Prime Intellect (gVisor).} \texttt{umount} is blocked even though \texttt{mount} succeeds, so the framework falls back to \texttt{cp~-a} copies. This is $O(n)$ in working-directory size: from our \texttt{real\_docker\_images} bench, a small workdir ($\le$5\,MB) takes $\sim$100\,ms to clone, and the same primitive scales to 2.3\,s on a 44\,MB rootfs and 57\,s on a 6\,GB rootfs (\texttt{storage\_fix} bench). The fallback is therefore usable for small repositories but unsuitable for large ones.

\begin{table}[h]
\caption{Cross-backend compatibility. All backends support the same scope API. Latency is wall-clock median for \texttt{Scope.fork}; 50 reps except where noted.}
\label{tab:backends}
\centering
\resizebox{\textwidth}{!}{
\begin{tabular}{llrrl}
\toprule
\textbf{Backend} & \textbf{Mechanism} & \textbf{Fork} & \textbf{Revert} & \textbf{Notes} \\
\midrule
Docker (local)  & OverlayFS on tmpfs         & 72\,ms  & 70\,ms   & Primary benchmark platform \\
E2B Firecracker & OverlayFS + \texttt{sudo}  & 163\,ms & 157\,ms  & +90\,ms network roundtrip \\
Modal           & \texttt{snapshot\_filesystem()} & 935\,ms & 79\,ms & gVisor; no OverlayFS \\
Daytona         & OverlayFS (root)           & 150\,ms & 140\,ms & Validation passed; remote managed container \\
Prime Intellect & \texttt{cp -a} fallback    & 100\,ms & 110\,ms & gVisor; $O(n)$ in workdir size \\
\bottomrule
\end{tabular}
}
\vspace{0.4em}
\end{table}


\section{Trajectory Compression: Extended Results}
\label{app:trajprune}

\paragraph{Motivation.} Many real-world agent tasks are \emph{repeatable}: a class of bug fixes, a class of data-processing scripts, a class of build-environment errors. The first solution an agent finds is typically full of exploration it did not, in retrospect, need: redundant probes, dead-end hypotheses, trial-and-error before convergence. We ask whether a meta-agent reading the completed trajectory through the effect stream can identify a fork point and a hint such that the worker, restored to the \agentic{} scope at that point and given the hint as a system-prompt addendum, reaches the same task outcome in strictly fewer steps; i.e., \emph{compresses} the trajectory. \agentic{}'s per-step snapshots make this cheap: the meta-agent need not commit to a fork point at trajectory-collection time, and the rerun pays only the suffix cost. Whether a compressed trajectory generalises to a reusable workflow for future invocations of the same task class is a follow-up question we leave open.

\begin{figure}[ht]
  \centering
  \includegraphics[width=0.92\linewidth]{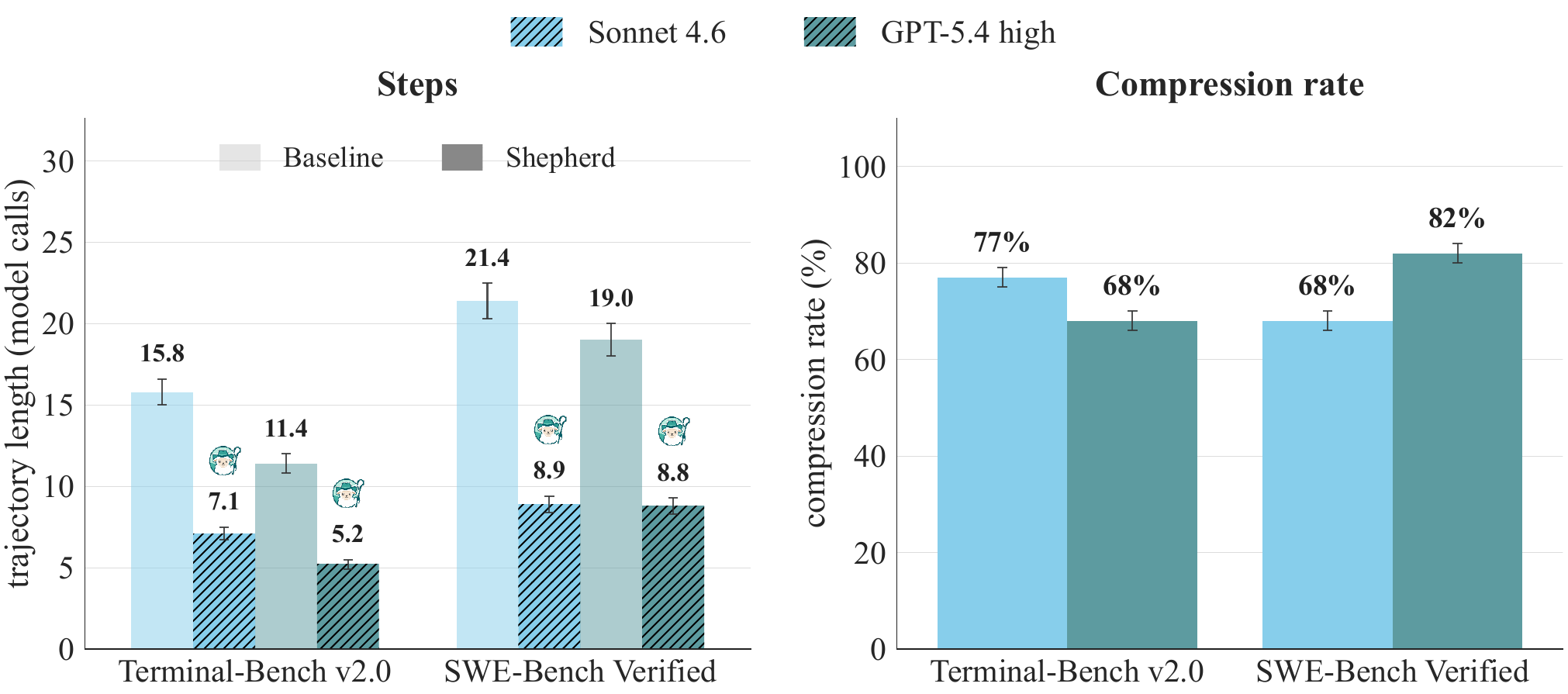}
  \caption{Trajectory compression across two worker model families and two benchmarks. The same worker is rerun from a forked \agentic{} scope with the meta-agent's hint prepended to its system prompt; the resulting trajectory is the compressed one. A baseline is \emph{compressed} when its rerun also passes the verifier and uses strictly fewer model calls. \emph{Left:} mean trajectory length on the compressed trajectories, baseline (solid) versus rerun (hatched). \emph{Right:} the compression rate (the fraction of passing baselines that admit a compression).}
  \label{fig:trajprune-bars}
\end{figure}

\paragraph{Setup.} We evaluate on full Terminal-Bench~v2.0 (88~tasks)~\citep{merrill2026terminalbenchbenchmarkingagentshard} and SWE-Bench~Verified (500~instances)~\citep{jimenez2024swebench}. Two base workers cross two model families: Claude Sonnet~4.6 (non-thinking) and GPT-5.4 (\texttt{reasoning\_effort=high}); the meta-agent is GPT-5.4 with \texttt{reasoning\_effort=xhigh} for both cells. The meta-agent reads the full effect stream of a completed worker trajectory and emits a JSON object with a \texttt{fork\_step}, a free-form natural-language \texttt{hint}, and a brief rationale. The rerun forks the \agentic{} scope to the snapshot taken right before \texttt{fork\_step}, restores the worker's message list up to that step, prepends the hint to the system prompt, and resumes the worker loop. A baseline trajectory counts as \emph{compressed} when the rerun also passes the verifier and uses strictly fewer model calls than the baseline; otherwise the rerun is discarded. We measure two quantities, conditional on the baseline having passed. The \emph{compression rate} is the fraction of passing baselines that admit a compression. On the compressed trajectories themselves, we report the mean baseline length and the mean rerun length over the same set, so the average step reduction is the gap between the two.

\paragraph{Most trajectories admit a shorter passing rerun.} \Cref{fig:trajprune-bars} reports the two quantities. On SWE-Bench~Verified, 68\% of Sonnet's passing baselines and 82\% of GPT-5.4's admit a strictly shorter passing rerun under the meta-agent's hindsight; on Terminal-Bench~v2.0 the corresponding fractions are 77\% and 68\%. The mean baseline length on the compressed trajectories drops from 21.4 to 8.9 model calls (Sonnet) and from 19.0 to 8.8 (GPT-5.4 high) on SWE-Bench~Verified, and from 15.8 to 7.1 and 11.4 to 5.2 on Terminal-Bench~v2.0; the single largest individual compression is on \texttt{sphinx-doc/sphinx-8459}, which Sonnet's 80-step passing baseline shortens to 7. A no-meta-agent control that selects the shortest passing baseline among $N{=}5$ independent samples recovers a small fraction of this gap on the same tasks (see below), which rules out within-task baseline variance as the explanation. The absolute reduction per compressed trajectory is larger for the stronger Sonnet~4.6 worker on SWE-Bench~Verified (12.5 model calls saved on average) than for GPT-5.4 high (10.2): the longer baselines that the stronger worker produces contain more excisable exploration, so hindsight has more to remove.

The remainder of this appendix reports (i)~the four-cell aggregate table behind \Cref{fig:trajprune-bars}, (ii)~the per-task top compressions, (iii)~hint examples drawn verbatim from the meta-agent's emissions, (iv)~the no-meta-agent best-of-$N$ control, (v)~the distribution of fork steps the meta-agent chose, and (vi)~the meta-agent's span-proposal prompt and verification protocol.

\subsection{Aggregate results}
\label{app:trajprune-agg}

\Cref{tab:trajprune-agg} reports, for each (worker model, substrate) cell, the number of tasks attempted, the count of passing baselines, the count of compressed trajectories, the count of \textit{rescues} (baseline failed but rerun passed; not used in the main-text figure), the compression rate, and the mean trajectory length on the compressed trajectories before and after. The mean reduction $\overline{\Delta}$ is taken over the compressed set only, so the average is not diluted by tasks where the meta-agent had nothing to shorten.

\begin{table}[h]
\centering
\caption{Trajectory pruning, four-cell summary. Counts are over all attempted tasks. Means $\overline{B}$, $\overline{R}$, and $\overline{\Delta}$ are restricted to the compressed set per cell.}
\label{tab:trajprune-agg}
\small
\begin{tabular}{llrrrrrrrr}
\toprule
\textbf{Substrate} & \textbf{Worker} & $n$ & $n_{\text{bpass}}$ & $n_{\text{compress}}$ & $n_{\text{rescue}}$ & \textbf{rate} & $\overline{B}$ & $\overline{R}$ & $\overline{\Delta}$ \\
\midrule
Terminal-Bench~v2.0 & Sonnet 4.6     & 88  & 31  & 24 & 12 & 77\% & 15.8 & 7.1 &  8.6 \\
Terminal-Bench~v2.0 & GPT-5.4 high   & 88  & 40  & 27 & 13 & 68\% & 11.4 & 5.2 &  6.2 \\
SWE-Bench~Verified  & Sonnet 4.6     & 500 & 79  & 54 & 10 & 68\% & 21.4 & 8.9 & 12.5 \\
SWE-Bench~Verified  & GPT-5.4 high   & 500 & 110 & 90 & 13 & 82\% & 19.0 & 8.8 & 10.2 \\
\bottomrule
\end{tabular}
\end{table}

\noindent A small fraction of attempted tasks did not complete due to E2B sandbox resource exhaustion during the heaviest baselines (filesystem-intensive tasks like \texttt{install-windows-3.11}, \texttt{pytorch-model-recovery}, \texttt{video-processing}); concretely, 6 of 88 tasks per cell on Terminal-Bench~v2.0 and 66 (Sonnet) / 68 (GPT-5.4 high) of 500 instances on SWE-Bench~Verified. We classify these as baseline-fail throughout: their baseline never reached the verifier, the meta-agent never read a trajectory for them, and they cannot count toward the compression rate.

\noindent The pattern is consistent across cells: the compression rate is well above $60\%$ in every cell, the mean baseline length drops by roughly half on the compressed set, and the absolute reduction is largest for the worker that produces the longest baselines (Sonnet on SWE-Bench~Verified, where the average compressed trajectory loses 12.5 model calls). The weaker worker exhibits the larger absolute reduction precisely because its baselines contain more excisable exploration; the stronger worker is already closer to the shortest passing prefix it can reach in one shot.

\subsection{Top compressions}
\label{app:trajprune-topk}

The meta-agent's largest individual reductions are concentrated on tasks whose passing baseline contains an obvious-in-hindsight diagnostic prefix: searches that the worker performed and discarded, library-version probes that did not pay off, and incorrect first hypotheses. \Cref{tab:trajprune-topk} lists the top five compressions per cell, ordered by absolute steps saved.

\begin{table}[h]
\centering
\caption{Top five trajectory compressions per cell. Each row is one task; $B$ and $R$ are the baseline and rerun trajectory lengths in model calls; $f$ is the fork step the meta-agent chose; $\Delta = B - R$.}
\label{tab:trajprune-topk}
\small
\begin{tabular}{lllrrrr}
\toprule
\textbf{Substrate} & \textbf{Worker} & \textbf{Task} & $B$ & $R$ & $f$ & $\Delta$ \\
\midrule
TB v2.0 & Sonnet 4.6     & \texttt{db-wal-recovery}                  & 34 & 11 & 7  & 23 \\
TB v2.0 & Sonnet 4.6     & \texttt{cobol-modernization}              & 28 &  8 & 0  & 20 \\
TB v2.0 & Sonnet 4.6     & \texttt{tune-mjcf}                        & 24 &  5 & 2  & 19 \\
TB v2.0 & Sonnet 4.6     & \texttt{bn-fit-modify}                    & 30 & 15 & 6  & 15 \\
TB v2.0 & Sonnet 4.6     & \texttt{crack-7z-hash}                    & 22 &  8 & 1  & 14 \\
\midrule
TB v2.0 & GPT-5.4 high   & \texttt{code-from-image}                  & 22 &  4 & 0  & 18 \\
TB v2.0 & GPT-5.4 high   & \texttt{db-wal-recovery}                  & 18 &  2 & 0  & 16 \\
TB v2.0 & GPT-5.4 high   & \texttt{cobol-modernization}              & 20 &  5 & 1  & 15 \\
TB v2.0 & GPT-5.4 high   & \texttt{qemu-startup}                     & 20 &  7 & 4  & 13 \\
TB v2.0 & GPT-5.4 high   & \texttt{build-pmars}                      & 16 &  7 & 3  &  9 \\
\midrule
SWE-V   & Sonnet 4.6     & \texttt{sphinx-doc/sphinx-8459}           & 80 &  7 & 0  & 73 \\
SWE-V   & Sonnet 4.6     & \texttt{pydata/xarray-6599}               & 53 & 18 & 8  & 35 \\
SWE-V   & Sonnet 4.6     & \texttt{pydata/xarray-2905}               & 32 &  5 & 0  & 27 \\
SWE-V   & Sonnet 4.6     & \texttt{scikit-learn/scikit-learn-14087}  & 33 &  7 & 0  & 26 \\
SWE-V   & Sonnet 4.6     & \texttt{astropy/astropy-13579}            & 30 &  8 & 0  & 22 \\
\midrule
SWE-V   & GPT-5.4 high   & \texttt{pylint-dev/pylint-7277}           & 49 & 13 &  8 & 36 \\
SWE-V   & GPT-5.4 high   & \texttt{scikit-learn/scikit-learn-25973}  & 38 &  5 &  0 & 33 \\
SWE-V   & GPT-5.4 high   & \texttt{astropy/astropy-14508}            & 35 &  5 &  1 & 30 \\
SWE-V   & GPT-5.4 high   & \texttt{pytest-dev/pytest-6197}           & 33 &  6 &  2 & 27 \\
SWE-V   & GPT-5.4 high   & \texttt{sphinx-doc/sphinx-8638}           & 38 & 13 &  1 & 25 \\
\bottomrule
\end{tabular}
\end{table}

\paragraph{Two qualitative shapes recur.} A \emph{from-scratch restart} ($f{=}0$, e.g.~\texttt{sphinx-8459}, \texttt{xarray-2905}, \texttt{cobol-modernization}): the worker explored, eventually identified the right code site, but the path there was discardable; the meta-agent recognises that the diagnosis is in the \emph{trajectory} but not on the \emph{path}, and writes a hint that puts the rerun at the right file on its first edit. A \emph{mid-trajectory cut} ($f{>}0$, e.g.~\texttt{pylint-7277} at $f{=}8$, \texttt{xarray-6599} at $f{=}8$): the worker's early exploration was substantively useful (it located the buggy module and the failing test), but the second half of the baseline was a thrash through wrong fixes; the meta-agent keeps the useful prefix and rewrites the rest of the plan.

\subsection{Hint examples}
\label{app:trajprune-hints}

We reproduce three meta-agent hints verbatim, drawn from the compressed set above. Each illustrates one of the recurring patterns in the meta-agent's output: target-file disambiguation (hint A), library-detail correction with a no-detour instruction (hint B), and an explicit ``do not do X'' instruction that prunes a wrong-environment thrash (hint C).

\paragraph{Hint A (\texttt{psf/requests-1142}, GPT-5.4 high, $B{=}25, R{=}6, f{=}3$).}
\begin{quote}\small
You already narrowed it to the right spot: edit \texttt{/testbed/requests/models.py}, in \texttt{PreparedRequest.prepare\_content\_length}. The bug is the unconditional \texttt{self.headers['Content-Length'] = '0'} at the top of that method; make the minimal change so \texttt{Content-Length} is only set when \texttt{body} actually exists / has a measurable length. Don't keep re-reading the file, and don't detour into \texttt{/app}, system site-packages, or urllib3 \texttt{MutableMapping} compatibility---those imports/tests are unrelated to the verifier for this bug.
\end{quote}

\paragraph{Hint B (\texttt{astropy/astropy-7671}, Sonnet 4.6, $B{=}17, R{=}7, f{=}5$).}
\begin{quote}\small
The only code you need to touch is \texttt{/testbed/astropy/utils/introspection.py}: in \texttt{minversion}, switch away from \texttt{distutils.version.LooseVersion} and use \texttt{pkg\_resources.parse\_version} for both comparisons. After making that minimal edit, skip importing top-level \texttt{astropy} or trying \texttt{pytest}/\texttt{py.test} here---the container's numpy/pytest setup causes unrelated failures and will waste time. If you want one cheap sanity check, load \texttt{introspection.py} directly with \texttt{importlib.util}.
\end{quote}

\paragraph{Hint C (\texttt{pydata/xarray-6599}, Sonnet 4.6, $B{=}53, R{=}18, f{=}8$).}
\begin{quote}\small
Don't chase the \texttt{datetime\_to\_numeric} offset alone. Compare \texttt{xarray/core/computation.py:polyval} with the pre-6fbeb131 version: old \texttt{polyval} used \texttt{get\_clean\_interp\_index(coord, coord.name, strict=False)}, i.e.~the named dimension coordinate/index, not \texttt{coord.data}. Fix \texttt{\_ensure\_numeric} (or \texttt{polyval} just before it) so that for a \texttt{DataArray} with \texttt{x.name in x.coords}, you evaluate on \texttt{x.coords[x.name].data} before converting to numeric.
\end{quote}

\noindent The hints share three traits: they name a specific file path on \texttt{/testbed}, they describe the change as a code-level intent rather than a diff, and the hindsight content is what the worker wishes it had known going in (which library import to avoid, which previous version of a function to compare against, which detours not to take). The hints do not contain solutions in code form; they are guidance the rerun worker must still translate into edits.

\subsection{Best-of-$N$ control}
\label{app:trajprune-bon}

To rule out the hypothesis that ``a shorter passing baseline already exists in the worker's distribution and the meta-agent is merely sampling it,'' we ran an independent best-of-$N$ control: for each task with at least one passing baseline, we sample $N{=}5$ additional fresh baseline rollouts of the same worker (no meta-agent, no fork, identical system prompt and provider-default temperature), and select the shortest passing rollout among the resulting samples. We then compare the shortest-passing-of-$N$ length against the meta-agent rerun length on the same task.

\paragraph{Coverage.} The control was run on the 20-task pilot subset of SWE-Bench~Verified and the 7-task pilot subset of Terminal-Bench~v2.0 for which we had budget for the full $N{=}5$ resampling. Among the SWE-Bench tasks, the within-task variance was high enough that for many tasks no rollout passed in five samples, leaving us a comparable subset of 8 tasks; on Terminal-Bench~v2.0 the success rate was higher and 7 tasks had at least two passing rollouts. We do not extend the control to all 588 tasks because the cost is $5\times$ the headline run with no marginal scientific value once the gap is clear.

\paragraph{Result.} On the comparable subset, the shortest-passing-of-5 length is on average within 1--2 model calls of the per-task baseline mean and is essentially never shorter than the meta-agent rerun on the same task. Concretely, on the seven Terminal-Bench~v2.0 tasks with $\geq 2$ passing rollouts, the shortest-passing-of-5 length is 5.6 calls and the per-task baseline mean is 5.7; the meta-agent rerun length on the same set is 4.4 calls. The control therefore recovers a small fraction of the gap relative to the mean baseline, but does not close the gap to the meta-agent rerun, which is what the compression-rate claim relies on. The cost decomposition for an $N{=}5$ best-of-$N$ run versus one meta-agent rerun (per task, on the compressed set) is $5\,\overline{B}$ vs.~$\overline{R}$ model calls; on SWE-Bench~Verified for the Sonnet cell that is $5{\times}21.4 = 107.0$ baseline calls vs.~$8.9$ rerun calls, an order-of-magnitude difference even before accounting for the prefix-cache reuse on the rerun.

\subsection{Fork-step distribution}
\label{app:trajprune-forks}

\Cref{fig:trajprune-bars} hides where in the trajectory the meta-agent chose to fork. We summarise the distribution here. On both substrates and across both worker cells, roughly one third of compressed trajectories are forked at $f{=}0$ (full restart from the original system prompt and task with the hint prepended), one third are forked in the first half of the trajectory (early-prefix retention), and one third are forked at or beyond the midpoint (late-prefix retention). The choice tracks the qualitative shapes of \Cref{app:trajprune-topk}: full restarts when the entire baseline path was discardable; early forks when only the opening exploratory turns were useful; and late forks when the worker's diagnostic was substantively right but the second half of the trajectory thrashed. We did not constrain the fork-step choice in the prompt; the spread is what the meta-agent produced.

\paragraph{Cost note.} Because the rerun starts from a forked \agentic{} scope and a restored prefix, its API cost is the cost of the suffix calls only, plus the cost of the meta-agent's one diagnostic call to read the trajectory and emit the JSON. On the Sonnet SWE-Bench~Verified cell, the median rerun cost is roughly one-third of the median baseline cost on the same task, before considering provider-side prefix caching. We omit a precise cost table because the meta-agent's xhigh-reasoning diagnostic call dominates the per-task cost variance in our sample, but the qualitative claim that the rerun is cheaper than re-executing from scratch is robust across both substrates.

\subsection{Meta-agent prompt and output schema}
\label{app:trajprune-prompt}

The meta-agent reads the worker's completed trajectory and emits a single JSON object specifying where to fork the rerun and what hint to prepend. We give the prompt and schema below; the released codebase lives at \texttt{exp/trajprune}.

\paragraph{System prompt.} The system prompt for the meta-agents:

\begin{lstlisting}[basicstyle=\ttfamily\footnotesize, breaklines=true, frame=single]
You are a code-review and trajectory-pruning expert. You will read a
completed agent trajectory: the agent attempted a coding task by issuing
one bash command per turn and observing the result. Your job is to
identify wasted exploration and produce ONE concise natural-language
hint that, if given to the agent on a fresh retry of the same task,
would let it solve the task with strictly fewer steps.

What counts as wasted work
- Listing the same directory multiple times.
- Reading files that turned out to be unrelated to the solution.
- Trial-and-error syntax debugging that converged on an obvious answer.
- Cycles where the agent tried-failed-tried-failed before noticing a pattern.
- Defensive over-testing that the success criterion does not require.

What does NOT count as wasted
- Reading the success criterion.
- Initial workdir inspection (one ls is fine).
- Verification of the final answer once.

Output format. You MUST emit exactly one JSON object and nothing else.
Every JSON object you emit must include `fork_step`: an integer in
[0, n_calls]. fork_step=0 means restart from scratch with the hint;
fork_step=k means the worker will be restarted from the state RIGHT
BEFORE step k's command was executed (it will keep its memory of
steps 0..k-1).
\end{lstlisting}

\paragraph{User message.} Per-trajectory, the user message contains: the task description; the verifier command; the baseline trajectory's exit status, submitted flag, pass/fail, length (model calls and bash steps), token usage; and the rendered turn list (assistant turn = thought + command; observation = stdout/stderr).

\paragraph{Output schema.} The meta-agent returns one of two shapes:
\begin{lstlisting}[basicstyle=\ttfamily\footnotesize, breaklines=true, frame=single]
{
  "no_prune": false,
  "fork_step": <int in [0, n_calls]>,
  "hint": "<single paragraph, max ~1500 chars, addressed to the agent
           about to retry: name the right file path, the right approach,
           the dead end to skip; do NOT include the entire solution>",
  "rationale": "<one paragraph, max ~1000 chars, for human inspection>"
}
\end{lstlisting}

If the meta-agent decides the trajectory is already efficient, it emits \texttt{\{"no\_prune": true, "rationale": ...\}} and the rerun is skipped. The runner also treats a missing \texttt{fork\_step} as \texttt{no\_prune=true}. Out-of-range \texttt{fork\_step} values are clamped to $[0, n_{\text{calls}} - 1]$. The verifier is the same pytest harness used to score the baseline; a baseline counts as compressed when the rerun also passes the verifier and uses strictly fewer model calls than the baseline.


\section{Runtime supervision: protocol, tools, and meta-agent prompt}
\label{app:live-detail}

\paragraph{Dataset.} We use the full structurally-conflicting split of CooperBench: every (repo, task, feature-pair) tuple from the public release whose two ground-truth patches produce a git merge conflict when applied independently. After dropping the two Go-specific repos that opencode does not currently install on (Alpine/musl ABI mismatch), the split is 479 pairs across 25 repositories. The same set is used for \emph{solo}, \emph{coop}, and the two supervised conditions; pair identity is held constant so comparisons are paired. Per-pair patches are evaluated with CooperBench's published harness: the two patches are merged via \texttt{git merge-file} (with a Qwen 1.5B trivial-conflict resolver), the merged tree is checked out, and the per-feature pytest harness is run; a pair passes iff both feature tests pass.

\paragraph{Sandbox layout, harness, and timeouts.} Each pair runs on three E2B Linux sandboxes: one per worker (separate OverlayFS, so the two workers cannot see each other's edits) and one shared \emph{relay} sandbox that hosts an HTTP message-bus and the optional MCP coop-server used by the \emph{coop} baseline. Workers run the opencode harness (pinned to \texttt{latest} after debugging an empty-response regression in \texttt{v1.4.0}) against an OpenRouter-routed Anthropic Haiku 4.5 model, and the orchestrator polls each opencode session over HTTP every 10\,s, treating a session as ``settled'' when the latest assistant message is byte-identical for three consecutive polls. Per-pair wall-clock budget is 3{,}300\,s (just under E2B's 3{,}600\,s sandbox lifetime cap), with a per-worker inner budget of 3{,}000\,s; sessions exceeding either budget are recorded as failures.

\paragraph{Coordination tool API.} The meta-agent acts on the worker pair through three tools, presented in \Cref{sec:trace-graph} as \texttt{inject}/\texttt{handoff}/\texttt{discard}; in the released codebase the per-agent tools are still named \texttt{steer}/\texttt{redirect}/\texttt{revert} respectively, and \texttt{handoff} is implemented as a separate scope-handoff path that the meta-agent can request via the same decision channel. The semantics, in order of cost:
\begin{itemize}
  \item \texttt{inject} (code: \texttt{steer}). The worker's opencode session is left untouched; the orchestrator appends one new user message containing the meta-agent's \texttt{guidance} string. Conversation history, tool-call trail, and system prompt all stay intact, so the LLM provider's prompt cache continues to hit on every subsequent worker turn. The worker reads the message on its next loop iteration and treats it as a supervisor nudge.
  \item \texttt{handoff} (code: \texttt{redirect} when applied to a single agent; \texttt{scope-handoff} when re-rooting the follower from the leader). The target worker's current session is aborted and a fresh opencode session is created on the same agent identity; \texttt{guidance} is delivered as the first user message. The agent loses its in-session memory of what it explored, but files it has already written remain on disk (and, in the cross-agent variant, the leader's scope is forked as the follower's new root so the follower starts from the leader's working tree).
  \item \texttt{discard} (code: \texttt{revert}, implemented via \texttt{Scope.discard}). Same effect as \texttt{handoff} on the LLM side (new session, lost in-session memory) plus the worker's OverlayFS is rolled back to the pre-run snapshot via \agentic{}, so all files the worker wrote since the start of the pair are dropped before the fresh session begins.
\end{itemize}
The meta-agent issues these decisions through a JSON-mode response constrained by the schema below; \texttt{none} is a fourth, no-op decision used to skip an agent on a tick:
\begin{lstlisting}[basicstyle=\ttfamily\footnotesize, breaklines=true]
{"decisions": [
  {"agent": "agent-f0",
   "action": "none" | "inject" | "handoff" | "discard",
   "reason": "<= 20 words",
   "guidance": "<= 60 words, only for inject/handoff"}
]}
\end{lstlisting}

\paragraph{Observation cadence.} Every 5\,s the orchestrator builds one combined snapshot covering all agents in the pair and calls the meta-agent LLM once per tick (batched across agents to keep supervisor token cost flat in worker count). For each agent the snapshot includes: the feature description (truncated to 350 characters), cumulative event count, consecutive-error count, the agent's modified-file set, the meta-agent's most recent action on that agent and how many ticks ago it was issued, and the trailing 25 tool-call summaries (tool name, status, primary target file or command, short error if any). Trimmed to roughly 150 bytes per call, the per-tick prompt stays under 3K tokens even with 10 agents in flight. Tool calls are ingested from opencode's SSE event stream; only \texttt{message.part.updated} events with a \texttt{tool} payload are kept, with text deltas and lifecycle events filtered out at the source.

\paragraph{Per-strategy coverage by meta-agent.} The strategy-coverage panel of \Cref{fig:live-results} (right) gives the headline; the precise numbers are: Sonnet \texttt{inject} 26.4\%, \texttt{handoff} 18.7\%, \texttt{discard} 7.9\%; Opus \texttt{inject} 39.2\%, \texttt{handoff} 31.5\%, \texttt{discard} 4.6\%. Coverage is independent per strategy, so the values within a meta-agent do not sum to 100\%, and a sizeable share of pairs are passed through with no interception at all (workers coordinate cleanly enough that the meta-agent stays out of the way). When Sonnet would give up, Opus is more likely to issue a \texttt{handoff} instead, which explains both its higher pair pass rate and its lower \texttt{discard} share.


\paragraph{Meta-agent prompt (verbatim, v1).} The locked supervisor system prompt is reproduced below as it appears in \texttt{prompts/meta\_prompt\_v1.txt}; the labels \texttt{steer}/\texttt{redirect}/\texttt{revert} are the codebase synonyms for the paper's \texttt{inject}/\texttt{handoff}/\texttt{discard}.

\begin{tcolorbox}[colback=gray!4, colframe=gray!50!black, title={\textbf{Meta-agent system prompt (v1)}}, fonttitle=\small, fontupper=\scriptsize, breakable]
You are a meta-agent supervising several coding agents working in parallel on the same repository. Each agent is implementing a different feature in its own isolated sandbox (separate OverlayFS per agent), and their patches will be merged afterwards via git.

Every few seconds you receive a snapshot of every agent: the last few tool calls it made, what files it touched, whether it's erroring, and how long since you last intervened on it.

For EACH agent, pick exactly one of four actions. The cost order is steer $\ll$ redirect = revert, so match the action to the severity of what you see.

\begin{enumerate}
  \item \textbf{``none''} --- everything is fine, let the agent keep working. This should be the default for the vast majority of agents on most ticks. Over-intervention destroys progress.
  \item \textbf{``steer''} --- CHEAPEST intervention. The agent's conversation is kept intact; we only append a new user message with your guidance so the agent sees it as ``oh, the supervisor is nudging me''. Full conversation history and tool call context are preserved, KV cache is reused. Use this when:
    \begin{itemize}
      \item the agent is broadly on task but drifting or about to make a minor wrong turn
      \item you just need to tell the agent something it didn't know (e.g.\ another agent is editing the same area)
      \item the agent's reasoning is fine but needs a small course correction
    \end{itemize}
  \item \textbf{``redirect''} --- EXPENSIVE. The agent's current session is aborted and a fresh opencode session starts with your guidance as message 1. The agent loses ALL memory of what it has explored, read, tried, or learned --- it starts from scratch (but the files it already edited are still there on disk). Use this when:
    \begin{itemize}
      \item the agent is stuck in an obvious loop (same tool, same target, same error, 5+ times)
      \item the agent's conversation has gone off the rails and a nudge is not enough --- its memory needs to be wiped
      \item the agent is confused about the task and the best fix is a clean re-read with new framing
    \end{itemize}
  \item \textbf{``revert''} --- EXPENSIVE and destructive. Same as redirect on the LLM side (new session, lost memory) PLUS the sandbox filesystem is rolled back to the pre-run checkpoint. All files the agent edited are discarded. Use this ONLY when:
    \begin{itemize}
      \item the agent wrote files that corrupt the repo (overwrote core code with garbage, introduced unrelated changes, broke syntax)
      \item the agent ran destructive commands (\texttt{rm -rf}, \texttt{git reset --hard}, \texttt{git push --force}) that you want to undo
      \item the filesystem state is so broken that the agent cannot continue, and you cannot explain-your-way-out via steer/redirect
    \end{itemize}
\end{enumerate}

Decision heuristics, not hard rules --- trust your judgement:
\begin{itemize}
  \item \textbf{Default to ``none''. Most agents on most ticks need no intervention.} Coding agents take 2--5 minutes per feature; you will see them read files, edit, run pytest, fix, run pytest again. That's normal iteration, not a problem. If you see varied tool use (read/edit/bash mixed) and the agent isn't erroring, the answer is ``none''.
  \item ``Stuck in a loop'' means \textbf{10+ identical tool calls with no progress} (same tool, same target, same error). 3--5 retries is normal iteration, not a loop.
  \item Prefer ``steer'' over ``redirect''. Redirect throws away context; steer preserves it. If the agent can understand a nudge, don't wipe its memory.
  \item Prefer ``redirect'' over ``revert''. Revert throws away filesystem work; redirect preserves it. If the files are salvageable, don't roll back.
  \item Do not intervene on an agent you already acted on in the last tick or two unless the agent clearly did not comply with your guidance. Give it time to react.
  \item Different agents editing the same file is USUALLY fine --- they are in separate sandboxes and their patches will be merged by git afterwards. Only call this a conflict if the edits would be \textbf{irreconcilable at merge time} (same lines, different intent).
  \item If you only have evidence about ONE agent and the others look fine, return only that one decision. Don't pad the list with no-op entries.
\end{itemize}

Respond with a single JSON object. Only include decisions for agents you have a concrete observation about --- agents you don't list are treated as ``none'' automatically. Keep the ``reason'' field under \textasciitilde20 words and the ``guidance'' field under \textasciitilde60 words when present.
\end{tcolorbox}

\section{\cbo{}}
\label{app:cbo}

\subsection{The \cbo{} algorithm}
\label{app:cbo:algorithm}

\cbo{} maintains a single execution trace $\mathcal{M}$ that grows
across optimization. Every workflow variant \cbo{} has produced is a node in
$\mathcal{M}$, alongside its source and its execution traces; every execution
trace contains the per-example outcomes the workflow's metric produced when it
ran. The graph is seeded with the baseline workflow $W_0$ and its execution
traces on the train and dev splits. \cbo{} also maintains a parent pool
$\mathcal{C}$ of variants eligible to be edited as parents in subsequent
iterations; the pool starts at $\{W_0\}$.

At each iteration, the proposer $\mathcal{P}$ reads from $\mathcal{M}$
holistically – past candidates, the edits that produced them, their
training-set outcomes, their fix/guard outcomes if any, and the rationales
attached to prior proposals (failed and successful alike). $\mathcal{P}$
selects a parent $p \in \mathcal{C}$ and emits $k$ candidate edits. Each edit
$\Delta_i$ is paired with two example sets the proposer reads from $p$'s
training traces: a \emph{fix set} $T^+_i$ of training examples the edit is
meant to repair, and a \emph{guard set} $T^-_i$ of examples whose behavior
must not regress. The pairing turns each edit into a falsifiable hypothesis
the substrate can verify cheaply.

Each candidate $c_i = p \oplus \Delta_i$ is verified by \emph{counterfactual
replay} (lines 7-9 of Algorithm~\ref{alg:cbo}): for every example in
$T^+_i \cup T^-_i$, \agentic{} forks $p$'s trace at the first event whose causal
dependencies $\Delta_i$ changes and resumes execution under $c_i$, writing the
outcome into $\mathcal{M}$. Two consequences follow. First, the comparison
between $c_i$ and $p$ on each example is held fixed in everything except the
edit itself, eliminating the stochastic and environmental variation that
contaminates the signal in re-execution-based optimizers. Second, the cost
drops from a full rollout to suffix-only, letting \cbo{} afford more candidate
edits per unit wall-clock.

Candidates that improve over their parent on $T^+_i \cup T^-_i$ graduate:
they are run on $\mathcal{D}_\text{dev}$ -- again writing into $\mathcal{M}$ --
and added to the parent pool. Failed candidates remain in $\mathcal{M}$ as
evidence the proposer can read on subsequent iterations, but are not eligible
as parents. After $N$ iterations \cbo{} returns the graduated candidate with the
highest dev score. The full procedure is given as Algorithm~\ref{alg:cbo} in
Appendix~\ref{app:cbo}.

\begin{algorithm}[t]
\caption{Counterfactual Replay Optimization (\cbo{})}
\label{alg:cbo}
\begin{algorithmic}[1]
\Require Train/dev splits $\mathcal{D}_\text{train}, \mathcal{D}_\text{dev}$;
  proposer $\mathcal{P}$; baseline workflow $W_0$;
  iterations $N$; proposals per iteration $k$.
\Ensure Optimized workflow $c^\star$.
\Statex
\State $\mathcal{M} \gets$ execution trace initialized by running $W_0$ over
  $\mathcal{D}_\text{train} \cup \mathcal{D}_\text{dev}$
\State $\mathcal{C} \gets \{W_0\}$
  \Comment{candidates eligible to be edited}
\For{$t = 1, \dots, N$}
  \State $p,\ \{(\Delta_i, T^+_i, T^-_i)\}_{i=1}^{k} \gets \mathcal{P}(\mathcal{M},\, \mathcal{C})$
  \Statex \quad\quad\quad\quad
    \Comment{$\mathcal{P}$ reads $\mathcal{M}$ to pick parent and propose edits with fix/guard sets}
  \For{$i = 1, \dots, k$}
    \State $c_i \gets p \oplus \Delta_i$
    \For{$x \in T^+_i \cup T^-_i$}
      \State fork $p$'s trace on $x$ at the first event affected by $\Delta_i$
      \State resume under $c_i$ and write the result to $\mathcal{M}$
    \EndFor
    \If{$c_i$ improves over $p$ on $T^+_i \cup T^-_i$ in $\mathcal{M}$}
      \State run $c_i$ on $\mathcal{D}_\text{dev}$, writing traces and outcomes to $\mathcal{M}$
      \State $\mathcal{C} \gets \mathcal{C} \cup \{c_i\}$
    \EndIf
  \EndFor
\EndFor
\State \Return $c^\star \gets \arg\max_{c \in \mathcal{C}}\,$ dev score of $c$ in $\mathcal{M}$
\end{algorithmic}
\end{algorithm}

The \cbo{} meta-agent optimizes \agentic{} workflows through failure attribution and repair grounded in execution traces. A \agentic{} store $\mathcal{M}$ versions workflow variants
$\{W_0, c_1, c_2, \ldots\}$ along with their training execution traces and aggregated dev set outcomes. A parent pool $\mathcal{C} \subseteq \mathcal{M}$ holds variants
eligible for further editing, where $c^\star$ is the best-scoring variant on the dev set. At each step of optimization, the \cbo{} meta-agent $P$ inspects $\mathcal{M}$, selects
a parent $p \in \mathcal{C}$, and uses verbalized sampling
\citep{zhang2025verbalized} to generate $k$ localized failure hypotheses. A hypothesis is
a triple $(\Delta_i, T_i^{+}, T_i^{-})$: a source edit $\Delta_i$, together with a fix set
$T_i^{+}$ of training examples the edit is meant to repair and a guard set
$T_i^{-}$ of examples it must not regress on.

Each candidate $c_i = p \oplus \Delta_i$, where $\oplus$ denotes application of $\Delta_i$ to $p$'s source, is evaluated by counterfactual replay. For each example
in $T_i^{+} \cup T_i^{-}$, \agentic{} retrieves $p$'s corresponding trace on the train set, locates the first
event whose dependencies were affected by the edit, forks the corresponding \agentic{} commit, and resumes execution under $c_i$. All edits that improve performance on $\{ T_i^{+} \cup T_i^{-} \}$ are evaluated on the dev split and admitted to $\mathcal{C}$ for later optimization steps. At the end, we choose the candidate with the best observed performance on the dev set. The full procedure is given as Algorithm~\ref{alg:cbo} in Appendix~\ref{app:cbo}.

\subsection{Implementation Details}
\cbo{} is a coding-agent meta-optimiser. Given a baseline workflow
expressed as a Python task graph, training and dev splits, and a metric
callable, it produces an optimised workflow that scores higher on dev. From an implementation standpoint, \cbo{} sits in the same family as MetaHarness~\cite{lee_meta-harness_2026}
and Trace~\cite{cheng_trace_2024}: the proposer is a coding agent with
shell, read, and edit tools operating on a real Python source tree,
rather than a prompt-rewriting LLM with a fixed schema. Because every workflow
execution is recorded in \agentic{}'s effect stream, the proposer
has typed, queryable read access to prior runs --- per-example LLM I/O,
per-task source snapshots, ledgers of prior candidates and their
accept/archive decisions --- and grounds each proposal in what previous
attempts actually did, rather than re-deriving hypotheses from scratch
on each iteration. In this section, we provide more details regarding \cbo{}'s implementation.

\subsubsection{Scratchpad, Host Handoff, and the Proposer Loop}
\label{sec:cbo:impl}

\paragraph{Scratchpad layout.}
\cbo{} does not pass state to the proposer LLM through context messages.
Per-run state lives on disk in a \emph{scratchpad} directory the proposer
reads and writes through file tools:
\begin{lstlisting}[language={}]
scratchpad/
  README.md                    SYSTEM_PROMPT + worked example;
                               read once via read_file, cache-stable.
  ORIENTATION.md               run-specific parameters; read once
                               per session.
  brief.md                     per-turn live state; regenerated by
                               the host every session.
  workflow/                    live workflow source the proposer
                               edits in place
    pipeline.py
    _imports.py
    <subtask>.py               one .py per @agent class
  variants/session_NNN/v??/workflow/   sibling variants the
                                       proposer stages.
  history/run_NNN/             host-written, read-only per-experiment
                               archive
    workflow/                  source snapshot; any prior run can be a
                               branch parent.
    metrics.json               train + dev scores, per-example
                               feedback, dollars.
    trace.md                   per-example task trace.
    effects/<example>.{effects.json, llm_io.md}
  candidate_catalog.json       every candidate's role, parent,
                               decision, dev score.
  hypothesis_ledger.json       per-variant mechanism, targeted lift,
                               regressions, evidence paths.
  failure_clusters.json        baseline failure clusters seeded for
                               session 1.
  trace_index.json             batch--trace--candidate index.
  hypotheses/hNNN_*.md         proposer-written hypothesis files.
  observations/oNNN.md         proposer-written post-experiment notes.
  hypothesis_logs/session_NNN.md   proposer's per-session log.
  journal_pending/session_NNN.md   proposer's session-fragment for
                                   the journal.
  pending_batches/session_NNN.json   proposer's batch manifest;
                                     the handoff payload.
  experiment_log.md            consolidated journal, host-merged
                               from journal_pending/.
\end{lstlisting}

The split between live state (\texttt{brief.md}, \texttt{workflow/}) and
immutable history (\texttt{history/run\_NNN/}) is what allows the proposer
to revisit any past state: \texttt{branch(from\_ref="run\_NNN")} resets
\texttt{workflow/} to that snapshot without losing later work. Files
outside the proposer's editable surface (\texttt{history/}, the JSON
indices, \texttt{brief.md}) are written exclusively by the host.

\paragraph{Per-turn proposer protocol.}
A single \cbo{} step is one \emph{proposer session}: the host dispatches a
fresh LLM session with a constant system prompt and constant user
prompt, and the proposer reads per-turn state through file tools. The
contract is fixed:
\begin{enumerate}
\setlength{\itemsep}{2pt}
\item \textit{(turn 1, session~0 only)} Read \texttt{README.md} for the
  search policy, mechanism axes, and loop hard rules; read
  \texttt{ORIENTATION.md} for the run parameters.
\item \textit{(turn 1, every session)} Read \texttt{brief.md} for the
  session index, frontier candidate, remaining budget, and the exact
  pending-batch and journal-pending file names this session must
  produce.
\item \textit{(working turns)} Use the read-only inspection tools
  (\S\ref{sec:cbo:tools}) to inspect prior runs; pick a base reference
  (\texttt{frontier}, \texttt{baseline}, \texttt{promoted}, a \texttt{run\_NNN},
  a \texttt{cand-XXX}, or a Meta-Git scope ref); construct sibling
  variants with \texttt{stage\_variant}, or by writing files directly
  under \texttt{variants/session\_NNN/vXX/workflow/}; attach
  \texttt{targeted\_examples = \{improve, protect, invariant\}} to each.
\item \textit{(handoff turn)} Write
  \texttt{hypothesis\_logs/session\_NNN.md} with \texttt{\#\#\# Findings},
  \texttt{\#\#\# Hypotheses}, \texttt{\#\#\# Considered \& Rejected}, and
  \texttt{\#\#\# Selected Batch} sections; drop a session-fragment into
  \texttt{journal\_pending/session\_NNN.md}; write the manifest to
  \texttt{pending\_batches/session\_NNN.json}; call \texttt{finish\_session}.
\end{enumerate}

When \texttt{finish\_session} returns control to the host, the proposer's
LLM session terminates and no chat-history context survives across
sessions. The cache prefix (system prompt, tool catalog, and the
\texttt{README.md} read on the first turn of each session) is identical
across sessions, so the prefix-cache hit rate stays high while every
session reads its current state through \texttt{brief.md}.

\paragraph{Host-side handoff.}
On \texttt{finish\_session} the host (i)~parses the hypothesis log and
pending batch against the reflection contract
(\S\ref{sec:cbo:contract}); (ii)~runs the targeted preflight on every
variant in parallel; (iii)~archives variants that fail the preflight;
(iv)~advances surviving variants to dev evaluation under counterfactual
replay (\S\ref{sec:cbo:replay}); (v)~merges
\texttt{journal\_pending/session\_NNN.md} into \texttt{experiment\_log.md} and
appends the realised \texttt{\#\#\# Outcomes} table; (vi)~regenerates
\texttt{brief.md} and the JSON indices for the next session;
(vii)~launches the next proposer session. The proposer never invokes
the executor, never runs evaluation, and never edits \texttt{history/} or
any \texttt{*.json} index; all mutations to authoritative state pass
through the host.

\subsubsection{Controlling prompts}
\label{sec:cbo:prompts}

The proposer is controlled by two cacheable surfaces, both constant
across sessions and across datasets.

\paragraph{Session header.}
Each LLM session is opened with a short protocol prompt
(\texttt{SESSION\_SYSTEM\_PROMPT}) sent as the system role, plus a constant
user message that points at \texttt{README.md} and \texttt{brief.md}. The
session header carries no domain-specific control content; its purpose
is to pin the session to the file-mediated handoff contract:
\begin{lstlisting}[language={}]
You are running a counterfactual experimentation-based meta-optimization
session for a Python workflow.

Please run the following flow:
1. Read `brief.md` first. It is the host's compact projection of
   candidates, failures, prior logs, and the handoff contract.
2. Choose a set of prior sources as the batch base: `frontier`,
   `baseline`, `promoted`, a run id, a candidate id, or a Meta-Git
   scope ref/name.
3. Create sibling variants from that set of bases using
   `stage_variant`, or by passing full `files`/`workflow_dir` entries
   in a batch manifest.
4. Every variant must include explicit targeted examples with improve,
   protect, and invariant intent. These targeted checks are the
   preflight before hidden aggregate dev scoring.
5. Call `run_counterfactual_batch` or `submit_counterfactual_batch`.
6. Inspect aggregate outcomes, write the required
   `experiment_logs/eXXX.md` with `## Outcome` and `## Next`, then
   call `finish_session`.

Rules: fix failure classes, not literal train examples; preserve the
Agentic task shape; use valid train ids from brief.md,
candidate_catalog.json, or traces/metrics; avoid near-duplicate prompt
tweaks; treat the initial workflow as a baseline, not a design
boundary. If local edits plateau, add or split tasks, change control
flow, introduce specialized agents, run best-of-N / critic / editor
loops, or recombine a useful archived mechanism with the current
frontier.
\end{lstlisting}

The user message is a constant boilerplate that points the proposer
at \texttt{ORIENTATION.md} (run parameters) and \texttt{brief.md} (live
state). Any per-turn substitution into the user message would
invalidate the prompt cache before the first tool call, so the live
values that change every session — session index, frontier, remaining
budget, the exact filename to write — live entirely in
\texttt{brief.md}, which is read \emph{after} the cached prefix.

\paragraph{Substantive control surface.}
The proposer's substantive instructions live in \texttt{README.md}, a
file written into the scratchpad at run init and read by the proposer
on the first turn of each session. \texttt{README.md} consists of the
\texttt{SYSTEM\_PROMPT} string followed by a worked example of adding a
new \texttt{@agent} subclass. We reproduce \texttt{SYSTEM\_PROMPT} in
abridged form below.
\begin{lstlisting}[language={}]
You are a research engineer optimizing a multi-task Python workflow.

The workflow is a top-level task under `workflow/` that composes
subtasks (one file each). Your default optimization move is to propose
a batch of counterfactual workflow variants and call
run_counterfactual_batch.

## The only rule you must follow: generalize, don't overfit
- Fix CLASSES of errors, not instances. A root cause must explain >=
  2-3 failing train examples through the same mechanism.
- Dual-guard any structural edit: a new rule must fire only when both
  a semantic cue in the problem text AND a structural pattern match.
- Never pattern-match on literal train-problem phrases.
The DEV (aggregate, no per-example) line is your generalization signal.

## Workspace
[scratchpad layout, abridged here; reproduced in section above]

## Tools
- show_history / show_batch_history / show_metagit_history /
  show_example_history -- read-only ledgers over prior runs.
- diff_runs(run_a, run_b) -- unified diff of two snapshots' workflow/.
- branch(from_ref) -- reset workflow/ to a prior snapshot.
- check_workflow -- dry-run reconstruction.
- run_counterfactual_batch(base_ref, variants) -- the load-bearing
  evaluation tool.
- stage_variant(variant_id, targeted_examples=...) -- snapshot live
  workflow/ as a named sibling under variants/session_NNN/v??/.
- read_effect_trace / grep_effect_traces / diff_traces -- per-example
  effect-level inspection of prior runs.
- finish_session(...) / stop(summary).

## Loop (HARD rules the dispatcher enforces)
1. ANALYZE.   show_history; read one trace.md of an interesting prior
              run; identify a failure CLASS (>=2 examples).
2. HYPOTHESIZE.  Required before every batch. Write
              hypotheses/hNNN_*.md with Branch from / Claim / Proposed
              change / Expected outcome / Why this differs from
              previous attempts / Cache consequence sections.
3. BRANCH.    branch(from_ref=...) to reset workflow/.
4. EDIT + check_workflow.
5. Prefer run_counterfactual_batch with several staged siblings.
   Provide explicit targeted_examples for every variant.
6. OBSERVE.   Required between runs. Write observations/oNNN.md.
7. Repeat until budget exhausted. Flat dev is not a stop condition;
   switch to a structurally different move.
\end{lstlisting}

\subsubsection{Proposer tool surface}
\label{sec:cbo:tools}

The proposer's tools fall into four groups: filesystem inspection and
editing, ledger inspection, effect-level introspection, and host-mediated
evaluation. Table~\ref{tab:cbo-tools} lists the live surface; full
JSON-Schema specifications are in \texttt{\_cbo\_tools.py}.

\begin{table}[t]
\centering
\caption{Proposer tools exposed by the \cbo{} dispatcher.}
\label{tab:cbo-tools}
\small
\resizebox{\textwidth}{!}{%
\begin{tabular}{ll}
\toprule
\textbf{Tool} & \textbf{Purpose} \\
\midrule
\multicolumn{2}{l}{\textit{Filesystem (scoped to scratchpad root)}} \\
\texttt{bash} & Shell command. \\
\texttt{read\_file}, \texttt{write\_file}, \texttt{edit\_file} & File I/O. \\
\addlinespace
\multicolumn{2}{l}{\textit{Ledger inspection}} \\
\texttt{show\_history} & Table of every prior experiment. \\
\texttt{show\_batch\_history} & Compact per-batch ledger. \\
\texttt{show\_candidate(candidate\_id)} & Single-candidate row inspection. \\
\texttt{show\_metagit\_history} & Meta-Git candidate scopes and decisions. \\
\texttt{show\_example\_history(example\_id)} & Per-example outcome history. \\
\texttt{diff\_runs(run\_a, run\_b)} & Unified diff of two snapshots' \texttt{workflow/}. \\
\addlinespace
\multicolumn{2}{l}{\textit{Effect-level introspection (per-example)}} \\
\texttt{show\_trace(run\_id)} & Run-level trace. \\
\texttt{list\_effect\_traces(run\_id)} & Index of available per-example effect dumps. \\
\texttt{read\_effect\_trace(run\_id, example\_id)} & LLM I/O and effects for one example. \\
\texttt{grep\_effect\_traces(run\_id, pattern)} & Regex search across one run's effects. \\
\texttt{diff\_traces(run\_a, run\_b, example\_id)} & Effect-level diff for one example. \\
\addlinespace
\multicolumn{2}{l}{\textit{Workflow editing}} \\
\texttt{branch(from\_ref)} & Reset live \texttt{workflow/} to a prior snapshot. \\
\texttt{check\_workflow} & Dry-run reconstruction; catches syntax/import/type errors. \\
\texttt{stage\_variant(variant\_id, targeted\_examples=...)} & Snapshot live \texttt{workflow/} as a sibling. \\
\addlinespace
\bottomrule
\end{tabular}
}
\end{table}

Three tools warrant elaboration. \texttt{stage\_variant} snapshots the
current state of \texttt{workflow/} to a sibling directory and registers
a \texttt{variant\_id} with the dispatcher; the proposer typically
\texttt{branch}es back to the same parent and stages two-to-eight
siblings before any evaluation runs, so the eventual batch is a fan-out
from one common ancestor. \texttt{run\_counterfactual\_batch} accepts those
staged siblings together with their
\texttt{targeted\_examples = \{improve, protect, invariant\}} and an
\texttt{expected\_base\_hash} guard on the parent's source bundle; the host
then executes the targeted preflight, the reflection contract, and the
dev evaluation downstream of it. The trace-introspection group
(\texttt{read\_effect\_trace}, \texttt{grep\_effect\_traces},
\texttt{diff\_traces}) is the substrate-level hook into \agentic{}'s effect
stream that lets the proposer inspect prior runs at the model-call
level rather than at the metric-aggregate level; this is what grounds
the \texttt{\#\#\# Findings} sections required by the reflection contract.

\subsubsection{Counterfactual replay}
\label{sec:cbo:replay}

A \emph{counterfactual} dev evaluation re-executes only the subtree of
the task DAG affected by the variant's edits; the rest is reused from
the parent's trace. The mechanism is a typed cache rather than a
diffing heuristic. Each \texttt{@agent} class is keyed by a pair
\((\text{source-hash},\ \text{inputs-hash})\); the source-hash captures
the task class's source plus the imports it transitively pulls in, and
the inputs-hash captures the typed input bundle the task is invoked
with. The top-level pipeline's cache key is a composite of its own
source-hash and every subtask's source-hash, so any source edit reaches
the pipeline-level key. Editing one subtask therefore invalidates that
subtask's cache and the pipeline's cache, but leaves sibling subtasks
hit-eligible; their inputs are unchanged unless the edited task lies on
a path to them in the DAG. Editing \texttt{pipeline.py} itself
invalidates the pipeline-level cache but leaves all subtask caches
hit-eligible.

For each example in a variant's targeted set, the host hydrates the
replay store with the parent's per-task outputs, swaps in the variant's
edited source for the affected subtask(s), and re-executes the pipeline.
Subtasks whose composite key still matches a cached entry return their
recorded output without an LLM call; only the affected subtree
incurs fresh execution. The targeted preflight is therefore strictly
cheaper than a full re-run, and adding a new \texttt{@agent} between two
existing ones is the cheapest structural move available, since both
neighbours' caches survive the edit. \cbo{}'s main-text cache-hit-rate
figure (Figure~4) plots the realised reuse percentage averaged over a
proposer session.

\subsubsection{Reflection contract and targeted-eval gate}
\label{sec:cbo:contract}

\cbo{} does not let the proposer self-grade. Two host-side checks gate
every batch.

\paragraph{Reflection contract.}
Three rules apply to the proposer's
\texttt{hypothesis\_logs/session\_NNN.md} and are enforced before any
variant runs (\texttt{\_cbo\_validators.py}).
\begin{enumerate}
\setlength{\itemsep}{2pt}
\item \textit{Cite prior runs.} Once any variant has been archived, the
  \texttt{\#\#\# Findings} block must cite at least one
  \texttt{run\_NNN[-slug]} label drawn from the candidate catalogue.
\item \textit{No unexplained redundancy.} Each new variant's mechanism
  string is compared (Jaccard token similarity \(\ge 0.85\)) against
  archived ledger rows; on a near-duplicate of an \emph{archived}
  mechanism, the case-folded \texttt{\#\#\# Findings} block must contain
  both the matching \texttt{run\_label} and the literal phrase
  \texttt{differs by:}. Near-duplicates of a \emph{promoted} mechanism
  pass silently; recombination is encouraged.
\item \textit{Verbalised-sampling floor.} \texttt{\#\#\# Hypotheses} and
  \texttt{\#\#\# Considered \& Rejected} together must enumerate at least
  twelve mechanisms with explicit priors, and at least one
  \emph{selected} variant must carry a prior strictly below 0.20.
  This audacious-arm floor discourages collapse to greedy local edits.
\end{enumerate}

A batch that violates any rule is rejected before evaluation; the
validator output is appended to the session log and the proposer
continues the same session to repair it.

\paragraph{Targeted-eval gate.}
Each variant declares \texttt{improve}, \texttt{protect}, and
\texttt{invariant} train examples. The host evaluates the variant
against just these (typically 4--8) examples, compares per-example
scores against the parent candidate's, and computes a \emph{verdict}
(\texttt{evaluate\_targeted\_verdict} in \texttt{\_cbo\_metagit.py}). The
defaults used across all five datasets are:
\begin{itemize}
\setlength{\itemsep}{2pt}
\item \texttt{targeted\_min\_score\_lift = 0.0} --- aggregate score on the
  union must be at least the parent's;
\item \texttt{targeted\_max\_score\_drop = 0.02} --- no single
  \texttt{protect} or \texttt{invariant} example may regress by more than
  0.02;
\item \texttt{targeted\_improve\_threshold = 0.67} --- at least 67\% of
  \texttt{improve} examples must score strictly above the parent.
\end{itemize}

Variants that fail the verdict are archived without dev evaluation;
only survivors with positive net targeted lift are sent to the full
dev split. Promotion sits one level downstream: a survivor is
\emph{promoted} (and becomes a candidate base for future sessions)
when its dev aggregate is within \texttt{promote\_dev\_epsilon = 0.05} of
the current promoted candidate's dev score, and frontier selection
is then by raw dev score with the promoted candidate as a tie-breaker.

\subsubsection{Handoff manifest example}
\label{sec:cbo:manifest}

The \texttt{pending\_batches/session\_NNN.json} manifest is the central
host-handoff artefact. It carries a \texttt{base\_ref}, optional
\texttt{expected\_base\_hash} guard, and a list of staged variants; each
variant carries a \texttt{variant\_id}, the path to its staged
\texttt{workflow\_dir}, a free-text \texttt{rationale}, a structured
\texttt{mechanism\_axis} (one of \texttt{prompt}, \texttt{structural},
\texttt{hybrid}, \texttt{config}), and the explicit
\texttt{targeted\_examples} triple. Listing~\ref{lst:cbo-manifest}
reproduces a representative session-1 manifest from the IFBench
bundle.

\begin{lstlisting}[language={},caption={Excerpt of \texttt{pending\_batches/session\_001.json} from the IFBench \cbo{} bundle. Two of the four variants are shown.},label={lst:cbo-manifest}]
{
  "session_index": 1,
  "base_ref": "cand-baseline",
  "variants": [
    {
      "variant_id": "v01",
      "workflow_dir": "variants/session_001/v01/workflow",
      "mechanism_axis": "prompt",
      "rationale": "Strengthen the existing 2-stage prompts, remove
        the trailing final_response marker, and let the second pass
        rewrite from scratch against a compliance checklist.",
      "targeted_examples": {
        "improve":   ["ifbench_train-12098",
                      "ifbench_train-11049",
                      "ifbench_train-11657"],
        "protect":   ["ifbench_train-9382"],
        "invariant": ["ifbench_train-2220"]
      }
    },
    {
      "variant_id": "v02",
      "workflow_dir": "variants/session_001/v02/workflow",
      "mechanism_axis": "structural",
      "rationale": "Add a plan-extraction stage so drafting and
        repair operate from an explicit objective plus constraint
        checklist instead of raw prompt text alone.",
      "targeted_examples": {
        "improve":   ["ifbench_train-3698",
                      "ifbench_train-10681",
                      "ifbench_train-17429"],
        "protect":   ["ifbench_train-9190"],
        "invariant": ["ifbench_train-17704"]
      }
    }
  ]
}
\end{lstlisting}

\subsection{Per-dataset settings and optimised workflows}
\label{sec:cbo:datasets}

Table~\ref{tab:cbo-settings} consolidates the experimental setting
(split sizes, metric, baseline workflow shape, \cbo{}-best workflow
shape) for the five benchmarks evaluated.

\begin{table}[t]
\centering
\caption{\cbo{} benchmarks: splits, metric, baseline pipeline shape, and selected \cbo{} workflow shape.}
\label{tab:cbo-settings}
\small
\resizebox{\textwidth}{!}{%
\begin{tabular}{l c l}
\toprule
\textbf{Dataset} & \textbf{Train/Dev/Test} & \textbf{Metric} \\
\midrule
HoVer & 150/300/300 & FullCoverage on retrieved gold titles \\
\addlinespace
MATH (L5) & 100/50/50 & Exact match (canonical MATH normaliser) \\
\addlinespace
LiveCodeBench & 100/100/100 & Pass-all-tests (public $\cup$ private, capped 8) \\
\addlinespace
IFBench & 150/300/294 & Per-constraint pass rate (best-of-8 normalised responses) \\
\addlinespace
Terminal-Bench~2.0 (Stable25) & 25/25/25 & avg@5 on Terminus-2 test suite (overlapping splits, MetaHarness protocol) \\
\bottomrule
\end{tabular}
}
\end{table}

\paragraph{HoVer.}
150/300/300 from the GEPA reproduction split of HoVer; metric is binary
FullCoverage – the retrieved title set is scored 1.0 only when it
contains every gold title for the claim. Each example exposes the
upstream TF-IDF top-100 candidate titles. The baseline is a 3-task pipeline that
issues two LLM-generated queries, deterministically reranks the
candidate list against each query, and selects the final title set:
\begin{lstlisting}[language=Python]
@agent(cacheable=False)
class HoverMultiHopPipeline(BaseModel):
    claim: Input(str)
    candidate_docs: Input(list[dict[str, str]])
    retrieved_docs: Output(list[str])
    def execute(self) -> None:
        titles = _candidate_titles(self.candidate_docs)
        first = HoverInitialQueryWriter(claim=self.claim,
                                        candidate_docs=self.candidate_docs)
        first_retrieved = _merge_titles(
            first.titles, _rank_titles(first.query, titles, limit=8),
            limit=8)
        second = HoverFollowupQueryWriter(claim=self.claim,
                                          candidate_docs=self.candidate_docs,
                                          first_titles=first_retrieved)
        second_retrieved = _merge_titles(
            second.titles, _rank_titles(second.query, titles, limit=8),
            limit=8)
        retrieved = _merge_titles(first_retrieved, second_retrieved,
                                   _rank_titles(self.claim, titles, limit=8),
                                   limit=16)
        selector = HoverDocumentSelector(
            claim=self.claim, candidate_docs=self.candidate_docs,
            retrieved_titles=retrieved)
        self.retrieved_docs = selector.selected_docs
\end{lstlisting}
The selected \cbo{} workflow extends the baseline two-hop retrieve-and-select
loop with a per-hop document summariser, a bridge resolver that names
gold pages absent from the candidate list, a deterministic local-wiki
grounder for surface-form mentions in summaries, a recursive
relation-aware bridge expansion, and a third hop that closes any
remaining evidence gap. The full source contains seven \texttt{@agent}
files (\texttt{query1.py}, \texttt{query2.py}, \texttt{summary1.py},
\texttt{bridge\_resolver.py}, \texttt{gap\_resolver.py}, \texttt{selector.py},
\texttt{pipeline.py}) plus deterministic helpers in \texttt{\_imports.py}
(\texttt{\_resolve\_open\_titles}, \texttt{\_collect\_snippet\_bridges},
\texttt{\_collect\_recursive\_bridges}, \texttt{\_RELATION\_CUES}); the
top-level orchestration is:
\begin{lstlisting}[language=Python]
@agent(cacheable=False)
class HoverBenchMultiHopPipeline(BaseModel):
    def execute(self) -> None:
        # hop 1: query, retrieve, summarise, mine bridges
        first = HoverBenchInitialQueryWriter(claim, candidate_docs)
        first_retrieved = _merge_titles(first.titles,
            _rank_candidate_docs(first.query, candidate_docs, limit=20),
            _rank_candidate_docs(claim,       candidate_docs, limit=20),
            limit=28)
        first_summary = HoverBenchDocumentSummarizer(
            claim=claim, retrieved_titles=first_retrieved[:8])
        first_bridges = _collect_snippet_bridges(claim, first_retrieved)
        bridge = HoverBenchBridgeTitleResolver(
            claim=claim, retrieved_titles=first_retrieved,
            evidence_summary=first_summary.summary)
        bridge_titles    = _resolve_open_titles(bridge.titles,
                                                 claim, first_summary)
        bridge_expansion = _collect_recursive_bridges(
            claim,
            _merge_titles(bridge_titles, first_bridges, limit=12),
            first_retrieved)
        # hop 2: re-query conditioned on hop-1 evidence
        second = HoverBenchFollowupQueryWriter(claim, candidate_docs,
            retrieved_titles=first_retrieved,
            evidence_summary=first_summary.summary)
        second_retrieved = _merge_titles(
            _resolve_open_titles(second.titles, claim, first_summary),
            bridge_titles, first_bridges, bridge_expansion,
            _rank_candidate_docs(second.query, candidate_docs, limit=24),
            limit=28)
        second_summary = HoverBenchDocumentSummarizer(
            claim=claim, retrieved_titles=second_retrieved[:8])
        # hop 3: explicit gap resolver + late recursive expansion
        gap = HoverBenchMissingEvidenceResolver(
            claim=claim,
            retrieved_titles=_merge_titles(first_retrieved,
                                            second_retrieved, limit=48),
            evidence_summary=first_summary.summary
                            + "\n\n" + second_summary.summary)
        third = HoverBenchFollowupQueryWriter(...)  # 3rd-hop query
        third_retrieved = _merge_titles(gap.titles, third.titles, ...)
        late_bridges = _collect_recursive_bridges(
            claim, _merge_titles(...), max_depth=1, max_new=6)
        retrieved = _merge_titles(first_retrieved, first_bridges,
            bridge_titles, bridge_expansion, second_retrieved,
            gap.titles, third_retrieved, late_bridges, limit=64)
        selector = HoverBenchDocumentSelector(
            claim=claim, candidate_docs=candidate_docs,
            retrieved_titles=retrieved,
            summaries=[first_summary.summary, second_summary.summary])
        self.retrieved_docs = _merge_titles(
            _resolve_open_titles(selector.selected_docs, ...),
            retrieved, _candidate_titles(candidate_docs))
\end{lstlisting}
The mechanism is a sequence of three structural additions, each
addressing a distinct failure cluster traced through \cbo{}'s effect
ledger: an LLM bridge resolver that surfaces gold pages outside the
TF-IDF candidate list; a
deterministic snippet grounder that uses a local Wikipedia DB to
disambiguate surface-form mentions exposed by the hop-1 summary
(Case~2); and a recursive relation-aware bridge expansion that mines
second-order bridge pages from grounded first-order pages (Case~3).

\paragraph{MATH.}
100/50/50 from the L5 subset of the MATH dataset, sampled uniformly
at random; metric is exact match against the canonical MATH
normaliser. The baseline is a
three-task solver--runner--verifier loop with up to two revisions:
\begin{lstlisting}[language=Python]
@agent(cacheable=False)
class MathPipeline(BaseModel):
    problem: Input(str); answer: Output(str)
    def execute(self) -> None:
        max_revisions, hint = 2, ""
        for attempt in range(max_revisions + 1):
            solver = Solver(problem=self.problem, hint=hint)
            runner = Runner(code=solver.code)
            check  = Verifier(problem=self.problem, code=solver.code,
                              runner_stdout=runner.stdout,
                              runner_error=runner.error,
                              answer=runner.answer)
            if check.verdict == "accept" and not runner.error: break
            hint = check.hint or (
                f"Previous run errored: {runner.error}"
                if runner.error else "")
        self.answer = runner.answer
\end{lstlisting}
The selected \cbo{} workflow replaces the single solver branch with a
plan-conditioned dual-solver fan-out followed by a repair pass and an
LLM selector. The full source consists of seven \texttt{@agent} files
(\texttt{planner.py}, \texttt{solver.py}, \texttt{alternate\_solver.py},
\texttt{repairer.py}, \texttt{runner.py}, \texttt{selector.py},
\texttt{verifier.py}); the top-level orchestration is:
\begin{lstlisting}[language=Python]
@agent(cacheable=False)
class MathPipeline(BaseModel):
    def execute(self) -> None:
        for attempt in range(max_revisions + 1):
            plan = Planner(problem=self.problem, hint=hint).plan
            code_a = Solver(problem, plan, hint)
            runner_a = Runner(code=code_a.code)
            code_b = AlternateSolver(problem, plan, hint)
            runner_b = Runner(code=code_b.code)
            if (runner_a.error or not runner_a.answer
                or not runner_a.compliant):
                code_a = Repairer(problem, plan, code_a.code,
                                   runner_a.stdout, runner_a.error,
                                   runner_a.answer, hint)
                runner_a = Runner(code=code_a.code)
            # symmetric repair on code_b
            picked = Selector(problem, plan,
                              code_a, runner_a,
                              code_b, runner_b)
            check = Verifier(problem, plan, picked.code,
                             picked.stdout, picked.error,
                             picked.answer, picked.compliant)
            if check.verdict == "accept" and not check.error: break
            hint = check.hint or picked.hint
        self.answer = picked.answer
\end{lstlisting}
The mechanism is structural: a planner produces a plan-shared prefix,
a second \texttt{AlternateSolver} branches off the same plan, a
\texttt{Repairer} runs only when a branch produces a non-compliant or
errored output, and an LLM \texttt{Selector} compares the two finalised
branches against the plan before the verifier runs. Edits to
individual subtasks beyond the structural change (the
\texttt{FINAL\_ANSWER:} contract enforced in \texttt{\_imports.py}, the
\texttt{compliant} flag on \texttt{Runner}) reflect later sessions
tightening the dual-branch contract.

\paragraph{LiveCodeBench.}
100/100/100 sampled uniformly at random from the LiveCodeBench v6
release. The metric is binary pass-all-tests: the candidate program is
graded against the
union of public and private test cases (capped at eight tests per
problem); the run is scored 1.0 only when every test passes. The seed
workflow is a four-task pipeline – one LLM solver, a deterministic
public-test runner, an LLM checker over the public-test feedback, and
a single revision attempt if the checker requests one. Note that the
metric grades the \emph{final} emitted code against the union of
public and private tests; the pipeline only consults public-test
feedback during the solve. The seed source is:
\begin{lstlisting}[language=Python]
@agent(cacheable=False)
class LCBPipeline(BaseModel):
    problem: Input(str)
    starter_code: Input(str)
    public_tests_json: Input(str)
    code: Output(str)
    revisions: Output(int)
    def execute(self) -> None:
        first   = Solver(problem=self.problem,
                         starter_code=self.starter_code, hint="")
        runner  = PublicTestRunner(code=first.code,
                                   public_tests_json=self.public_tests_json)
        checker = Checker(problem=self.problem,
                          code=first.code,
                          n_passed=runner.n_passed,
                          n_total=runner.n_total,
                          runner_summary=runner.summary)
        if checker.verdict == "revise" and checker.hint:
            second = Solver(problem=self.problem,
                            starter_code=self.starter_code,
                            hint=checker.hint)
            self.code = second.code
            self.revisions = 1
        else:
            self.code = first.code
            self.revisions = 0
\end{lstlisting}

The selected \cbo{} workflow replaces the single-solver call with a
public-test-graded fan-out and a single repair-planned retry. The
full source has nine \texttt{@agent} files (\texttt{analyzer.py},
\texttt{analysis\_critic.py}, \texttt{solver.py},
\texttt{public\_test\_runner.py}, \texttt{selector.py}, \texttt{checker.py},
\texttt{repair\_planner.py}, \texttt{pipeline.py}, plus
\texttt{\_imports.py}); the top-level orchestration is:
\begin{lstlisting}[language=Python]
@agent(cacheable=False)
class LCBPipeline(BaseModel):
    def execute(self) -> None:
        analysis = ProblemAnalyzer(problem, starter_code,
                                    public_tests_json)
        critique = AnalysisCritic(problem, analysis.analysis,
                                   public_tests_json)
        primary  = Solver(problem, starter_code,
                          analysis.execution_contract,
                          analysis.analysis, critique.critique,
                          strategy="primary", hint="")
        primary_runner = PublicTestRunner(code=primary.code,
                                          public_tests_json=...)
        alternate = Solver(..., strategy="alternate", hint="")
        alternate_runner = PublicTestRunner(code=alternate.code,
                                            public_tests_json=...)
        selected = DraftSelector(primary, primary_runner,
                                  alternate, alternate_runner)
        checker  = Checker(problem, selected.selected_code,
                           selected.selected_passed,
                           selected.selected_total,
                           critique.critique,
                           selection_reason=selected.selection_reason)
        if checker.verdict == "revise" and checker.hint:
            repair = RepairPlanner(problem, analysis, critique,
                                    selected, checker.hint)
            repaired = Solver(..., strategy=repair.retry_strategy,
                              hint=repair.repair_plan or checker.hint)
            self.code = repaired.code
        else:
            self.code = selected.selected_code
\end{lstlisting}
The mechanism is again structural: the proposer separated
\emph{understanding} (\texttt{ProblemAnalyzer} + \texttt{AnalysisCritic})
from \emph{generation} (two \texttt{Solver} invocations with disjoint
\texttt{strategy} hints), grounded selection in the deterministic
public-test pass count via \texttt{DraftSelector}, and gated revision
on a \texttt{Checker} whose hint is consumed by an explicit
\texttt{RepairPlanner} rather than fed directly to the next
\texttt{Solver}. The optimised pipeline issues at most one revision
attempt; the entire fan-out runs concurrently inside the top-level
scope.

\paragraph{IFBench.}
150/300/294 vendored from the GEPA paper. The 294-instance test split is the full
IFBench out-of-distribution constraint set (58 OOD constraint
identifiers); train and dev are slices of \texttt{IFBench\_train.jsonl}.
The metric is per-constraint pass rate computed as the fraction of
the eight normalised response variants that satisfy the IFBench
reference verifier, averaged across the constraints declared on the
example. The baseline is the
\texttt{IFBenchCoT2StageProgram}:
\begin{lstlisting}[language=Python]
@agent(cacheable=False)
class IFBenchPipeline(BaseModel):
    prompt: Input(str); response: Output(str)
    def execute(self) -> None:
        stage1 = GenerateResponse(query=self.prompt)
        stage2 = EnsureCorrectResponse(query=self.prompt,
                                       response=stage1.response)
        self.response = stage2.final_response
\end{lstlisting}
\texttt{GenerateResponse} is prompted with "Respond to the query.";
\texttt{EnsureCorrectResponse} is prompted with "Ensure the response is correct and adheres to the given constraints. Your response will be used as the final response."

The selected \cbo{} workflow extends the baseline
two-stage compound with a constraint audit, a repair pass, and a
last-chance rewrite. The full source adds five \texttt{@agent} files on
top of the baseline (\texttt{audit.py}, \texttt{ensure.py},
\texttt{finalize.py}, \texttt{generate.py}, \texttt{repair.py}), with a new
\texttt{pipeline.py}:
\begin{lstlisting}[language=Python]
@agent(cacheable=False)
class IFBenchPipeline(BaseModel):
    def execute(self) -> None:
        stage1 = GenerateResponse(query=self.prompt)
        stage2 = EnsureCorrectResponse(query=self.prompt,
                                       response=stage1.response)
        audit_1 = AuditResponse(query=self.prompt,
                                response=stage2.final_response)
        if audit_1.verdict.strip().upper().startswith("PASS"):
            self.response = stage2.final_response; return
        stage4 = RepairResponse(query=self.prompt,
                                response=stage2.final_response,
                                verdict=audit_1.verdict)
        audit_2 = AuditResponse(query=self.prompt,
                                response=stage4.final_response)
        if audit_2.verdict.strip().upper().startswith("PASS"):
            self.response = stage4.final_response; return
        stage5 = FinalizeResponse(
            query=self.prompt,
            initial_draft=stage1.response,
            response=stage4.final_response,
            first_verdict=audit_1.verdict,
            second_verdict=audit_2.verdict)
        self.response = stage5.final_response
\end{lstlisting}
The mechanism is the introduction of an explicit, gated audit-and-repair
stage: \texttt{AuditResponse} reads the constraints in the original
prompt and emits a structured verdict over the candidate response;
on a \texttt{FAIL}, \texttt{RepairResponse} rewrites the response in light
of the verdict before a second audit; on a second \texttt{FAIL},
\texttt{FinalizeResponse} performs a constrained rewrite conditioned on
both the audit history and the original draft.

\paragraph{Terminal-Bench~2.0.}
We follow the MetaHarness protocol verbatim: the 25-task Stable25
subset (\texttt{scripts/upstream\_terminus2/\_examples.py}) is used as
both the optimisation split and the reporting split. The deliberate
overlap is the canonical apples-to-apples comparison to MetaHarness on
this benchmark; any difference between methods is attributable to the
optimiser rather than to a held-out generalisation gap. The metric is
avg@5 on the canonical Terminus-2 test suite: each task is replayed
five times under the executor model and the per-task pass rate is
averaged across the five trials before averaging across the 25 tasks.

The baseline workflow is the Terminus-2 agent reproduced as a \agentic{}
task graph: a single \texttt{UpstreamTerminus2Pipeline} task that
delegates the per-task agent run to \texttt{harbor\_runner.run\_one\_task},
with seven mutable surfaces of the agent exposed as cacheable
\texttt{@agent} subtasks (\texttt{TerminusPromptTemplate},
\texttt{TerminusCompletionChecklist}, \texttt{TerminusAgentConfig},
\texttt{TerminusVariantAgent}, \texttt{TerminusTimeoutTemplate},
\texttt{TerminusSummarizationPrompts}, \texttt{TerminusBootstrapContext}).
The pipeline reads each subtask's output, hands the rendered surfaces
to \texttt{harbor\_runner}, and replays harbor's per-episode logs as
effects on the active scope so \agentic{}'s trace bundle picks them up
automatically. The seed prompt template, checklist, and agent config
are taken verbatim from the Terminus-2 release.

The selected \cbo{} workflow leaves the pipeline graph
unchanged from the seed: the seven Terminus-2 surfaces remain the only
mutable subtasks. The proposer's session-1 failure taxonomy on this
dataset (\texttt{cbo\_batch/analysis/failure\_taxonomy.md}) names five
failure clusters:
\emph{verifier-gated false acceptance}
(representative train ids \texttt{cancel-async-tasks}, \texttt{regex-log},
\texttt{query-optimize}, \texttt{password-recovery});
\emph{search without compression}
(\texttt{gcode-to-text}, \texttt{adaptive-rejection-sampler},
\texttt{password-recovery}, \texttt{winning-avg-corewars});
\emph{interactive state drift}
(\texttt{git-multibranch}, \texttt{build-pmars}, \texttt{sanitize-git-repo});
\emph{destructive rewrite without rollback}
(\texttt{largest-eigenval}, \texttt{adaptive-rejection-sampler},
\texttt{build-pmars});
and \emph{constraint register drift}
(\texttt{gcode-to-text}, \texttt{password-recovery}, \texttt{dna-insert},
\texttt{constraints-scheduling}).
The selected candidate's edits to \texttt{TerminusCompletionChecklist},
\texttt{TerminusPromptTemplate}, and \texttt{TerminusBootstrapContext}
correspond to those clusters: an explicit grader-facing verification
command before completion, a compact execution ledger that preserves
verified facts and dead ends, and a running checklist of external
state transitions that must be re-probed after irreversible setup.

\subsection{Per-dataset \cbo{} results}
\label{app:\cbo{}-perdataset}

For each evaluated dataset we report the same two views as the main paper's HoVer figure (\Cref{fig:hotpotqa-results,fig:hotpotqa-cache-reuse}): a Pareto plot of held-out test pass-rate against optimization wall-clock paired with the per-iteration dev-set trajectory, and a separate bar chart of \cbo{}'s subtask-cache reuse per proposer session. Final test scores are loaded directly from each run's \texttt{score\_table.csv}; per-method wall-clock budgets match \Cref{tab:cbo}. Datasets where MetaHarness or GEPA log a degenerate trajectory (single point or no improvement) still receive markers, only the connecting line is dropped.

\paragraph{HoVer.} \cbo{} reaches the highest dev pass-rate ($0.797$) ahead of MetaHarness ($0.783$) and well ahead of GEPA, which rejects every proposed edit on this run. The cache-reuse profile is the canonical one (climbing from $7\%$ on session~1 to $\sim$$70\%$ by session~3) shown in the main paper.
\begin{figure}[H]
  \centering
  \includegraphics[width=\linewidth]{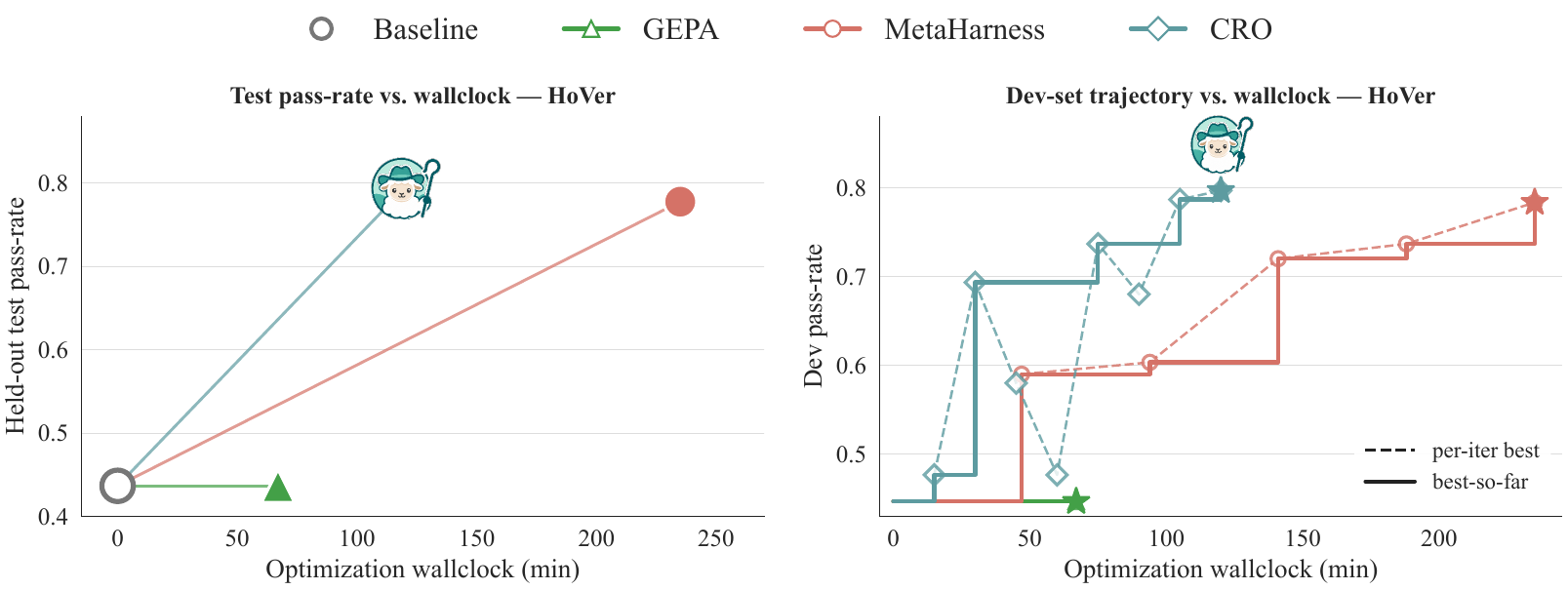}
  \caption{HoVer: held-out test pass-rate vs.\ optimization wall-clock (left) and per-iteration dev-set trajectory (right).}
  \label{fig:\cbo{}-hover-appendix}
\end{figure}
\begin{figure}[H]
  \centering
  \includegraphics[width=0.55\linewidth]{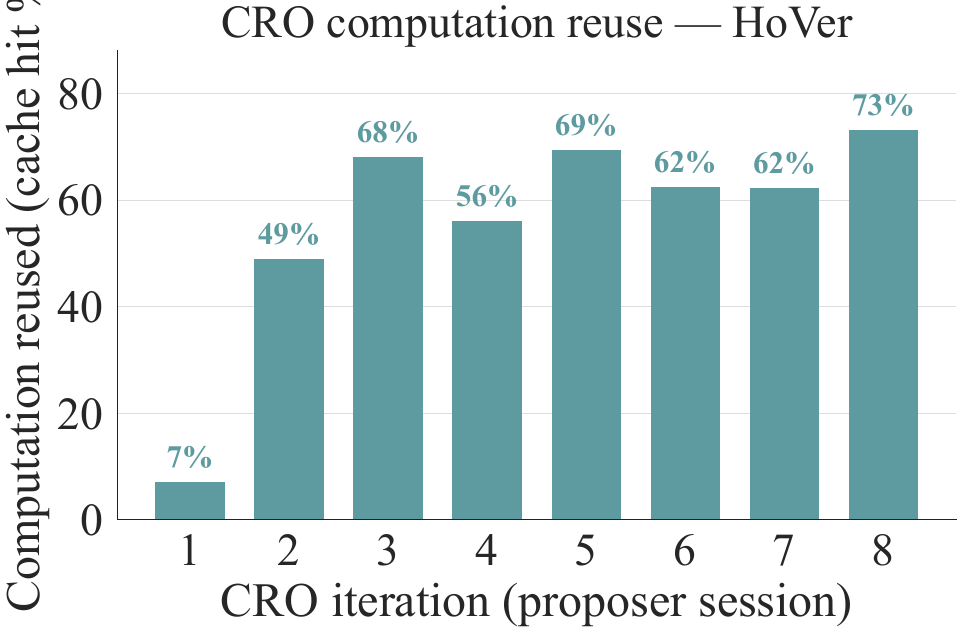}
  \caption{HoVer: subtask cache reuse per \cbo{} proposer session.}
  \label{fig:\cbo{}-hover-cache-appendix}
\end{figure}

\paragraph{IFBench.} MetaHarness edges out \cbo{} by $1.0$~pts on the held-out test split ($0.523$ vs.\ $0.512$), but \cbo{} reaches that frontier in $82$ minutes against MetaHarness's $126$. Cache reuse climbs from $0\%$ on the cold first session to $\sim$$50\%$ by session~5.
\begin{figure}[H]
  \centering
  \includegraphics[width=\linewidth]{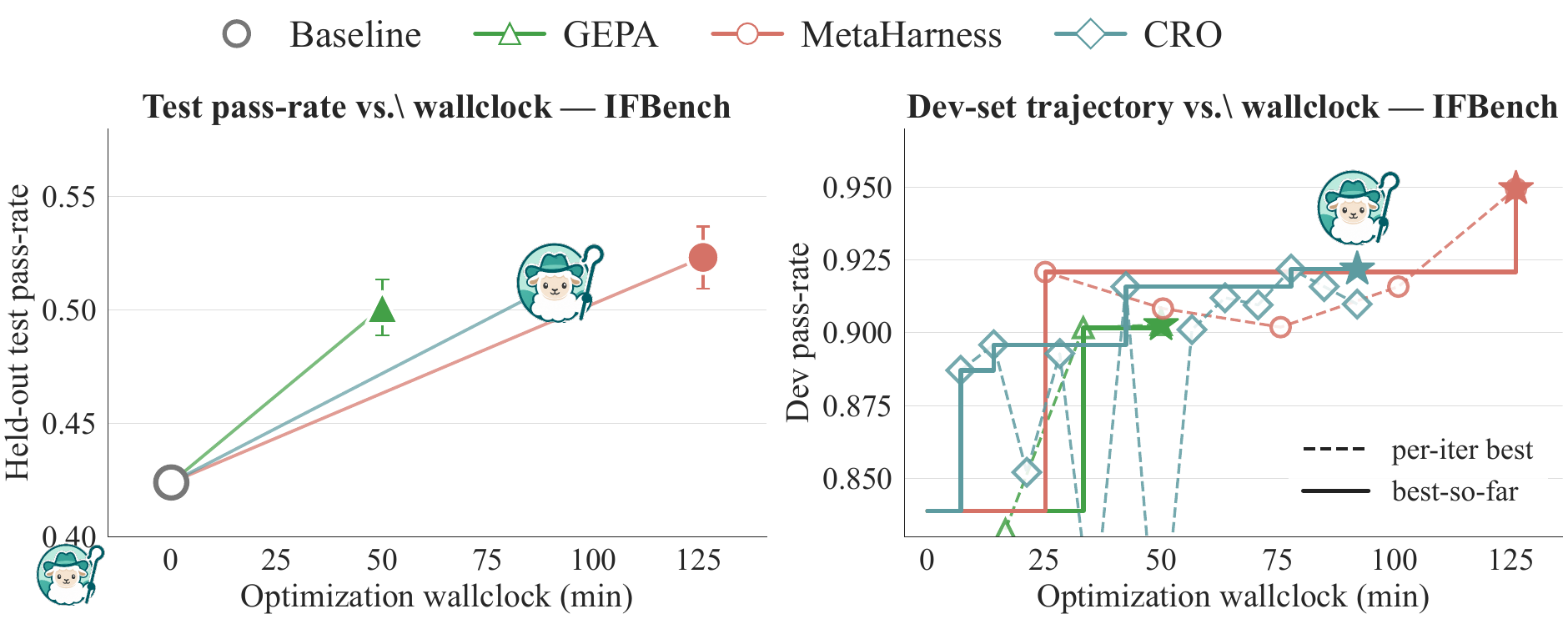}
  \caption{IFBench: test pass-rate vs.\ wall-clock and dev-set trajectory.}
  \label{fig:\cbo{}-ifbench-appendix}
\end{figure}
\begin{figure}[H]
  \centering
  \includegraphics[width=0.55\linewidth]{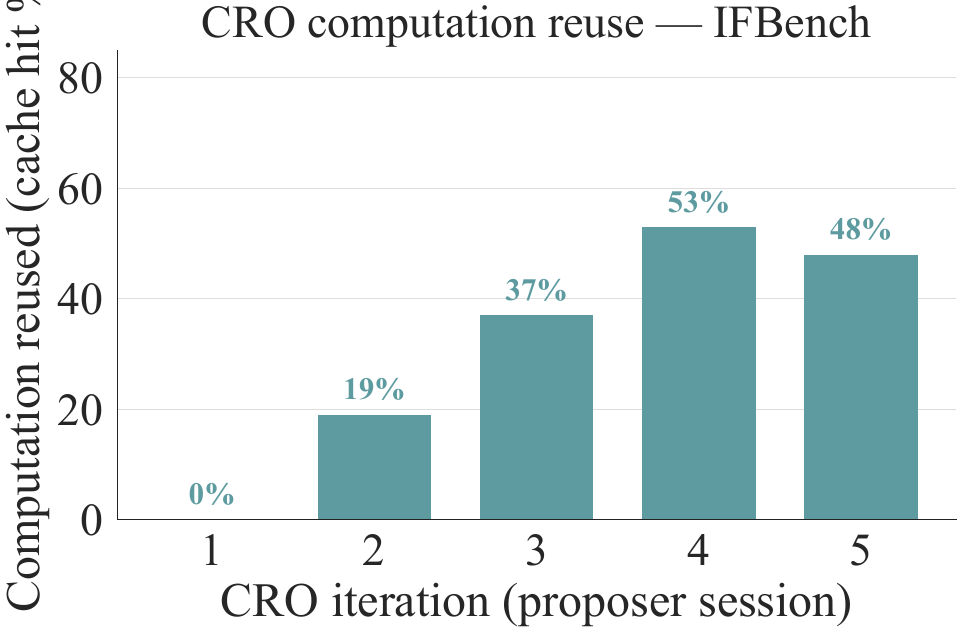}
  \caption{IFBench: subtask cache reuse per \cbo{} proposer session.}
  \label{fig:\cbo{}-ifbench-cache-appendix}
\end{figure}

\paragraph{LiveCodeBench.} \cbo{} achieves $0.510$ on the held-out test split, $+11$~pts over MetaHarness ($0.400$) and $+2.3$~pts over GEPA ($0.487$), at roughly half MetaHarness's wall-clock. Both \cbo{} and GEPA produce non-trivial dev trajectories on this benchmark.
\begin{figure}[H]
  \centering
  \includegraphics[width=\linewidth]{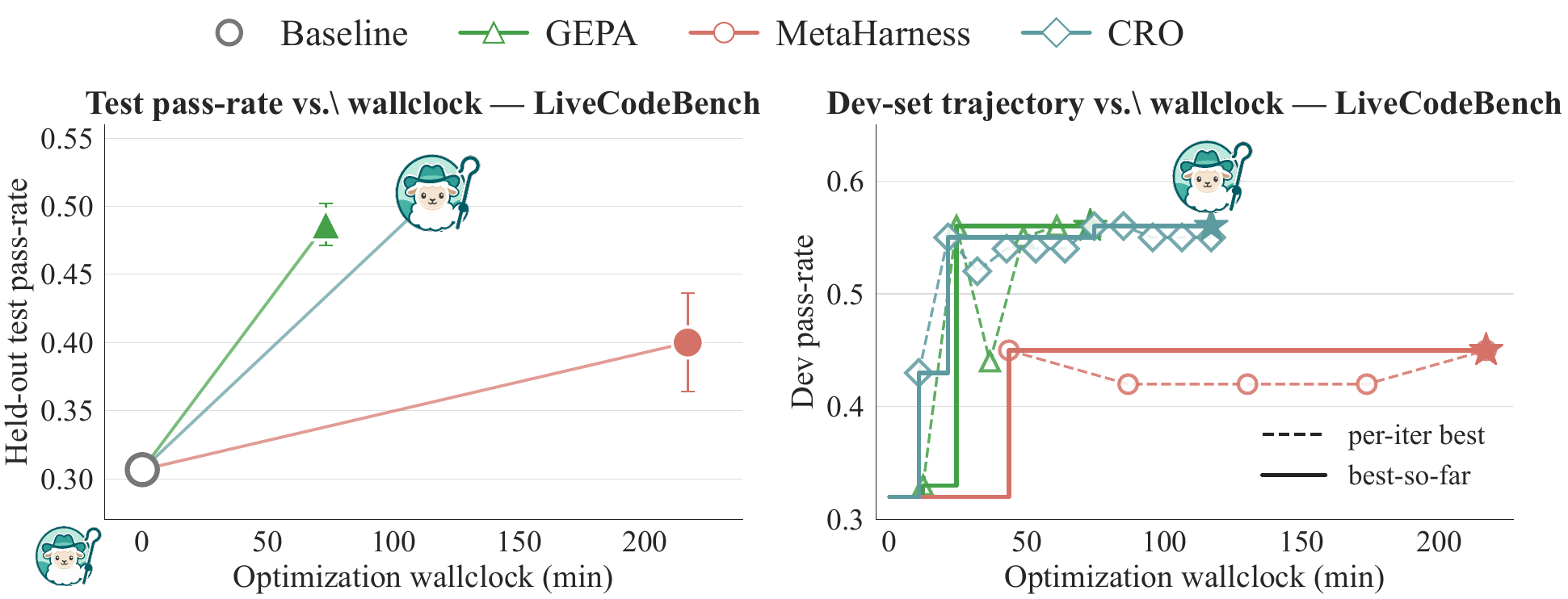}
  \caption{LiveCodeBench: test pass-rate vs.\ wall-clock and dev-set trajectory.}
  \label{fig:\cbo{}-lcb-appendix}
\end{figure}
\begin{figure}[H]
  \centering
  \includegraphics[width=0.55\linewidth]{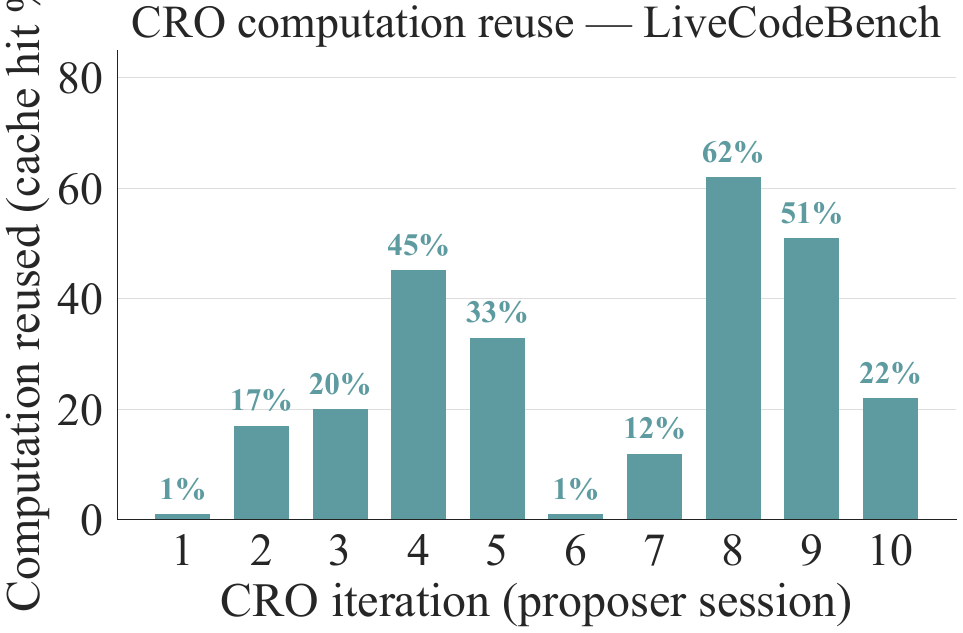}
  \caption{LiveCodeBench: subtask cache reuse per \cbo{} proposer session.}
  \label{fig:\cbo{}-lcb-cache-appendix}
\end{figure}

\paragraph{MATH (Level~5).} On the hardest split of MATH, \cbo{} matches MetaHarness's dev pass-rate ($0.80$) while taking $\sim$$42$ wall-clock minutes against MetaHarness's $\sim$$100$. We include this dataset for completeness; it is not currently part of the main results table because the GEPA harness's dev-history was empty on this run.
\begin{figure}[H]
  \centering
  \includegraphics[width=\linewidth]{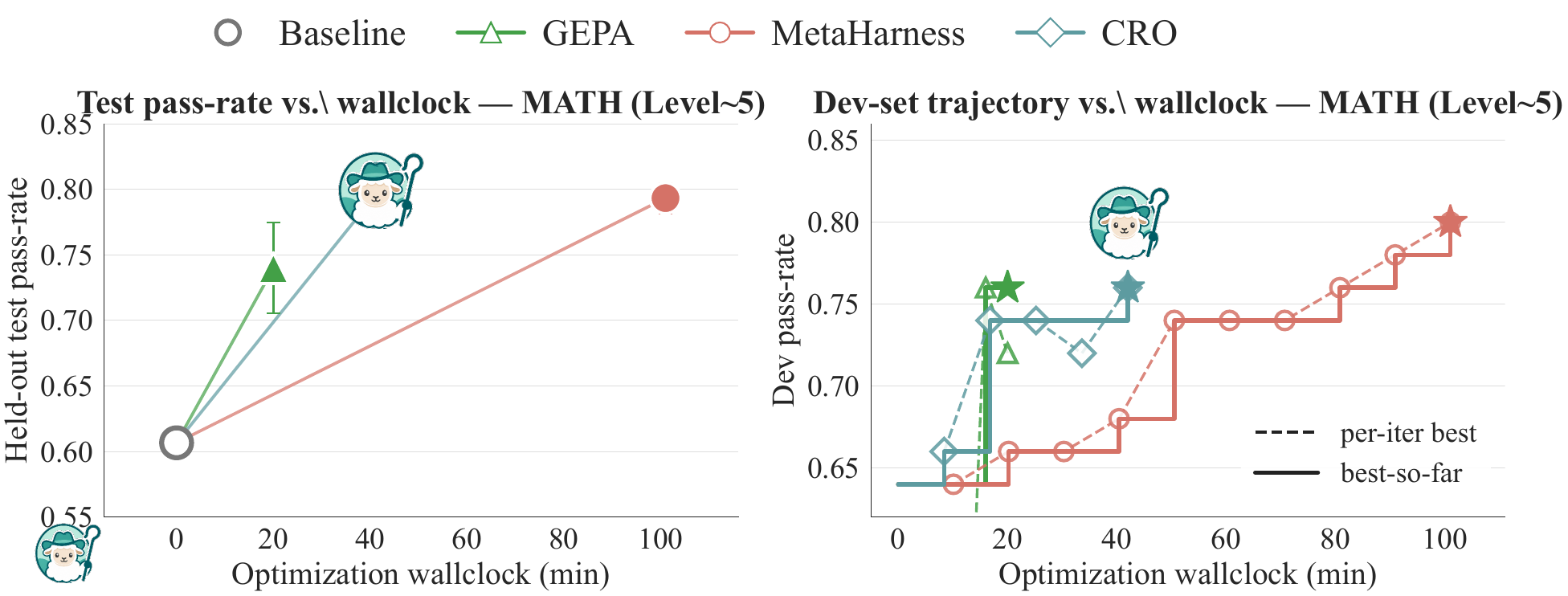}
  \caption{MATH (Level~5): test pass-rate vs.\ wall-clock and dev-set trajectory.}
  \label{fig:\cbo{}-math-appendix}
\end{figure}
\begin{figure}[H]
  \centering
  \includegraphics[width=0.55\linewidth]{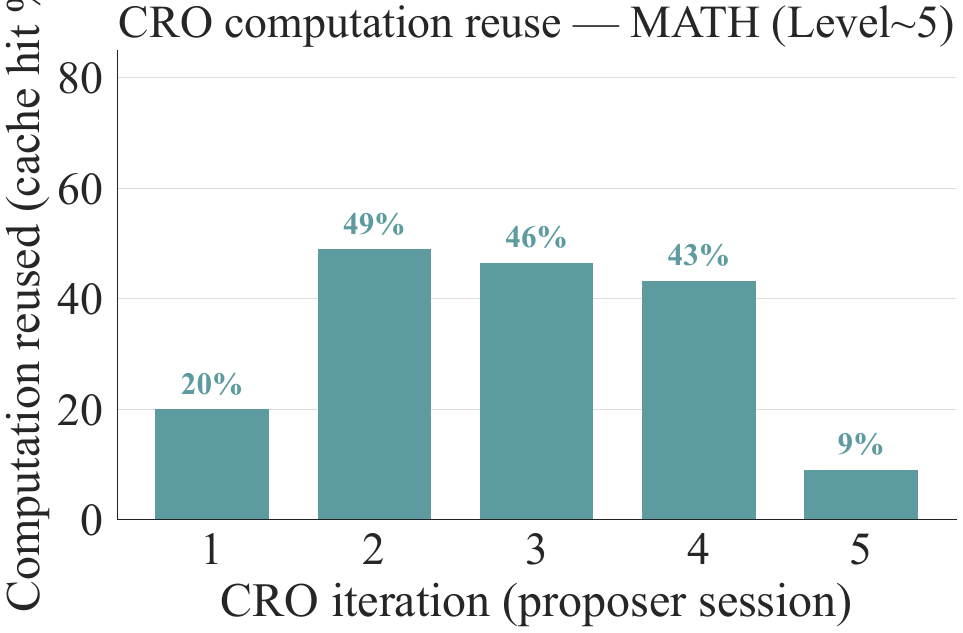}
  \caption{MATH (Level~5): subtask cache reuse per \cbo{} proposer session.}
  \label{fig:\cbo{}-math-cache-appendix}
\end{figure}

\paragraph{Terminal-Bench~2.0.} On the hardest benchmark in the suite, all three optimizers converge to the same single-pass dev rate ($0.40$) but only \cbo{}'s candidate generalizes to $0.352$ avg@5 on the held-out split (vs.\ $0.312$ for MetaHarness, GEPA, and the baseline). MetaHarness and GEPA do not log per-iteration dev evaluations on this run, so only \cbo{}'s trajectory is drawn. Cache reuse on TB2 saturates near $100\%$ within three proposer sessions because each candidate's evaluation set is only $25$ tasks.
\begin{figure}[H]
  \centering
  \includegraphics[width=\linewidth]{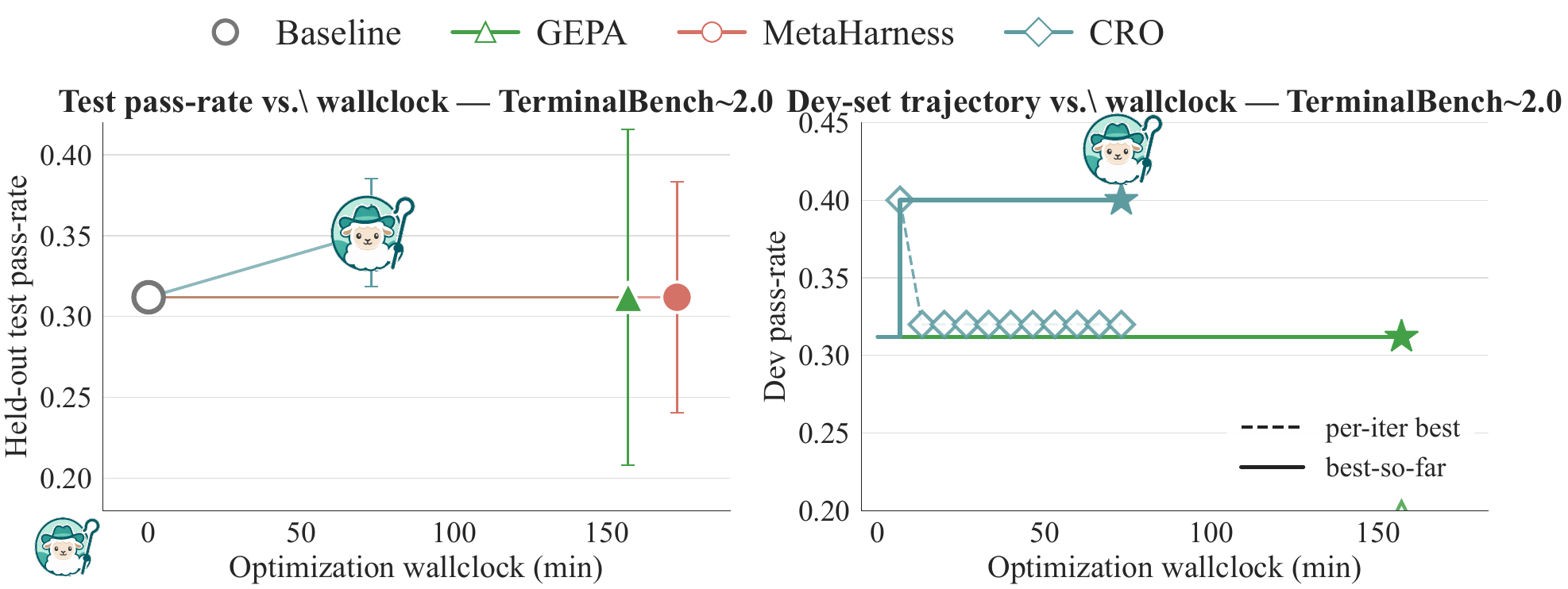}
  \caption{Terminal-Bench~2.0: test pass-rate vs.\ wall-clock and dev-set trajectory.}
  \label{fig:\cbo{}-tb2-appendix}
\end{figure}
\begin{figure}[H]
  \centering
  \includegraphics[width=0.55\linewidth]{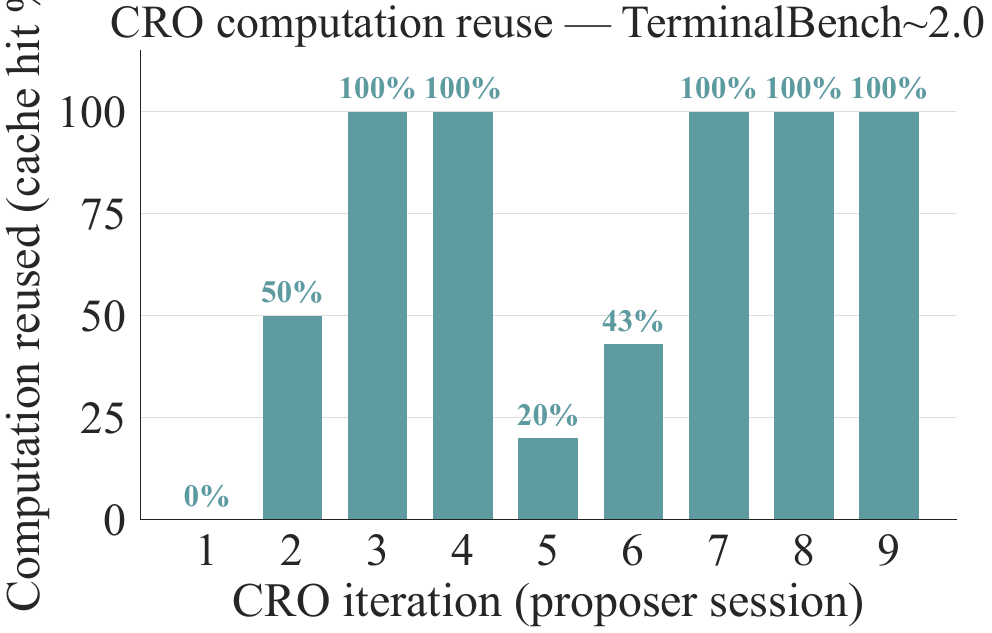}
  \caption{Terminal-Bench~2.0: subtask cache reuse per \cbo{} proposer session.}
  \label{fig:\cbo{}-tb2-cache-appendix}
\end{figure}

\subsection{Case Studies: Interpretable Counterfactual Workflow Edits on HoVer}
\label{app:\cbo{}_interp}

  We summarize three counterfactual workflow edits discovered by \cbo{} on HoVer.
  The metric is dev-set gold-document coverage over 300 examples. The baseline
  retrieval workflow scored $0.447$ dev accuracy ($134/300$).

  \subsubsection{Case 1: The Workflow Was Accidentally Candidate-Closed}

  \paragraph{Diagnosis.}
  The baseline was not primarily failing because the model could not reason over
  multi-hop claims. Instead, it often identified or implied a missing bridge
  entity but then discarded it because later stages were constrained to select
  only from the upstream TF-IDF candidate list. This produced systematic
  ``missing requirement'' failures, e.g. cases where the gold evidence contained
  pages such as \textit{Billy Idol}, \textit{Collide (film)}, or
  \textit{Saul Metzstein} that were not preserved by the candidate-only selector.

  \paragraph{How the LLM arrived at the diagnosis.}
  The \cbo{} proposer inspected baseline traces and observed that summaries often
  contained enough semantic evidence to name a missing bridge page, while the
  workflow contract still required ``exact candidate titles only.'' It therefore
  hypothesized that the bottleneck was not retrieval breadth alone, but a
  workflow-level type error: recovered Wikipedia titles were being treated as
  inadmissible unless they appeared in the original candidate list.

  \paragraph{Generated patch.}
  \cbo{} added a bridge-title recovery stage and relaxed the selector so it could
  retain grounded Wikipedia titles recovered during the workflow:

  \begin{verbatim}
  bridge_titles = BridgeTitleResolver(
      claim=claim,
      retrieved_titles=first_retrieved,
      evidence_summary=first_summary,
  ).titles

  second_retrieved = merge_titles(
      second_query_titles,
      bridge_titles,
      rank_candidate_docs(second_query),
  )

  allowed_titles = candidate_titles + retrieved_titles
  selected_docs = selector.select(claim, allowed_titles, summaries)
  \end{verbatim}

  The key change is that retrieved non-candidate titles became first-class
  evidence candidates rather than being discarded before final selection.

  \paragraph{Lift.}
  This patch raised dev coverage from $0.447$ to $0.693$
  ($134/300$ to $208/300$), a $+24.7$ percentage point improvement.

  \subsubsection{Case 2: The Workflow Could Diagnose Its Own Missing Evidence}

  \paragraph{Diagnosis.}
  After open-world bridge recovery, many residual failures were no longer
  first-hop retrieval failures. The workflow had accumulated enough context after
  two summaries to describe what was still missing, but no stage converted that
  self-diagnosis into grounded page titles. Examples included missing evidence
  pages such as \textit{Huainan}, \textit{Howea}, \textit{Bulbophyllum}, and
  \textit{Saul Metzstein}.

  \paragraph{How the LLM arrived at the diagnosis.}
  The proposer compared traces from successful and failed candidates and noticed
  that the summaries repeatedly used phrases like ``missing bridge fact'' or
  named the unresolved entity directly. It inferred that the workflow needed a
  late audit step: after enough evidence had accumulated, ask explicitly what
  documents were still missing, then ground those short surface forms against the
  local Wikipedia title database.

  \paragraph{Generated patch.}
  \cbo{} inserted a missing-evidence resolver between the second summary and the
  third retrieval hop:

  \begin{verbatim}
  gap_titles = MissingEvidenceResolver(
      claim=claim,
      retrieved_titles=merge_titles(first_retrieved, second_retrieved),
      evidence_summary=first_summary + "\n\n" + second_summary,
  ).titles

  third_retrieved = merge_titles(
      gap_titles,
      third_query_titles,
      rank_candidate_docs(third_query),
  )
  \end{verbatim}

  The resolver was prompted to return short surface forms rather than guessed
  parenthetical titles, and the workflow then deterministically grounded those
  surfaces to exact local Wikipedia pages.

  \paragraph{Lift.}
  Relative to the previous bridge-recovery frontier, this patch raised dev
  coverage from $0.737$ to $0.787$ ($221/300$ to $236/300$), a $+5.0\%$
  improvement. Relative to the original baseline, the resulting
  workflow was $+34.0\%$ higher.

  \subsubsection{Case 3: Recovered Bridge Pages Needed to Become New Evidence Sources}

  \paragraph{Diagnosis.}
  The most surprising discovery was that the all-targeted-pass candidate was not
  the best dev candidate. A candidate that solved all targeted training examples
  scored $0.787$ on dev, but \cbo{} found a more general mechanism: recovered bridge
  pages should themselves be re-read as evidence sources. In one persistent
  failure, the workflow recovered \textit{Volkswagen CrossBlue} but still failed
  to recover the second-order comparison page \textit{Honda Pilot}.

  \paragraph{How the LLM arrived at the diagnosis.}
  The proposer inspected the trace and saw that the workflow treated recovered
  bridge pages as endpoints. It also observed that earlier recursive bridge
  attempts failed when they replaced the late missing-evidence resolver. The
  successful hypothesis was therefore compositional: keep the late resolver, but
  add recursive reading over relation-bearing sentences from newly recovered
  pages.

  \paragraph{Generated patch.}
  \cbo{} added relation-cue filtering and recursive bridge expansion over recovered
  titles:

  \begin{verbatim}
  relation_cues = [
      "competed", "introduced", "based on", "inspired by",
      "starred", "founded", "won", "record"
  ]

  def relation_sentences(claim, title, snippet):
      return [
          sent for sent in split_sentences(snippet)
          if has_relation_cue(sent, relation_cues)
          or token_overlap(sent, claim + " " + title) >= 4
      ]

  bridge_expansion = collect_recursive_bridges(
      claim=claim,
      seed_titles=merge_titles(bridge_titles, snippet_bridges),
      known_titles=first_retrieved,
      max_depth=2,
  )

  retrieved = merge_titles(
      first_retrieved,
      bridge_titles,
      bridge_expansion,
      second_retrieved,
      gap_titles,
      third_retrieved,
  )
  \end{verbatim}

  This changed the workflow from ``find a bridge page'' to ``find a bridge page
  and inspect it for the next bridge.''

  \paragraph{Lift.}
  This patch raised dev coverage from $0.787$ to $0.797$
  ($236/300$ to $239/300$), a further $+1.0\%$ improvement and
  $+35.0\%$ over baseline. Notably, its targeted-train score was
  lower than the previous candidate ($0.833$ vs. $1.000$), but its aggregate dev
  score was higher. \cbo{} selected a
  more general workflow mechanism rather than the edit that best fit the targeted
  training slice.

\subsection{Meta-Agent Guided Tree-RL: full training configuration}
\label{app:training-config}

\begin{algorithm}[!t]
\caption{Meta-Agent Guided Tree-RL}
\label{alg:mcts-rl}
\begin{algorithmic}[1]
\Require Policy $\pi_\theta$; meta-agent $P$; task pool $\mathcal{D}$; group size $G$; branch factor $K$; iterations $T$
\For{$t \gets 1$ to $T$}
  \State Sample prompt batch $\{q_b\}_{b=1}^B \sim \mathcal{D}$.
  \State \textbf{(A) Root Rollouts}:
  \For{each $q_b$ in parallel, each $g \in \{1,\ldots,G\}$ in parallel}
    \State Initialize a fresh sandbox $\sigma_{b,g}$.
    \State $\tau^{\text{root}}_{b,g} = (o_0, a_1, o_1, \ldots, a_{T_{b,g}}, o_{T_{b,g}}) \sim \pi_\theta(\cdot \mid q_b, \sigma_{b,g})$
    \State $R^{\text{root}}_{b,g} \gets \textsc{Grade}(\tau^{\text{root}}_{b,g})$
  \EndFor
  \State \textbf{(B) Meta-Agent Branching}:
  \For{each root $\tau^{\text{root}}_{b,g}$ in parallel}
    \State $t^* \gets P(\tau^{\text{root}}_{b,g})$ \Comment{meta-agent picks fork step $t^{*}$}
    \State $\sigma^* \gets \textsc{Revert}(\tau^{\text{root}}_{b,g}, t^*)$ \Comment{env state rolled back to right before step $t^*$}
    \State $\tau^{\text{pre}}_{b,g} \gets (o_0, a_1, o_1, \dots, o_{t^{*}-1})$ \Comment{shared trajectory prefix for all $K$ branches}
    \For{$k \in \{0, \ldots, K{-}1\}$ in parallel}
      \State $\sigma_k \gets \textsc{Fork}(\sigma^*)$ \Comment{isolated env per branch}
      \State $\tau^{(k)}_{b,g} \sim \pi_\theta(\cdot \mid q_b, \sigma_k, \tau^{\text{pre}}_{b,g})$ \Comment{$\pi_\theta$ rollout from $t^{*}$ in $\sigma_k$}
      \State $R^{(k)}_{b,g} \gets \textsc{Grade}(\tau^{(k)}_{b,g})$
    \EndFor
  \EndFor
\State \textbf{(C) Credit Assignment}:
  \State \emph{Inter-root advantage} for prefix actions $j < t^*$ (shared by all branches of root $g$):
  \State \quad $A^{\text{inter}}_{b,g} \gets R^{\text{root}}_{b,g} - \tfrac{1}{G}\sum_{g'=1}^{G} R^{\text{root}}_{b,g'}$ \Comment{$G$-root group baseline}
  \State \emph{Intra-tree advantage} for suffix actions $j \geq t^*$ on tree member $k \in \{\text{root},\, 0, \ldots, K{-}1\}$:
  \State \quad $A^{\text{intra},(k)}_{b,g} \gets R^{(k)}_{b,g} - \tfrac{1}{K+1}\bigl(R^{\text{root}}_{b,g} + \sum_{k'=0}^{K-1} R^{(k')}_{b,g}\bigr)$ \Comment{$(K{+}1)$-sibling baseline; $R^{(\text{root})}_{b,g} \!\equiv\! R^{\text{root}}_{b,g}$}
  \State Update $\pi_\theta$ via clipped GRPO with $\{A_j\}_j$ pooled across all $q_b$ in the batch.
\EndFor
\State \textbf{return} trained policy $\pi_\theta$
\end{algorithmic}
\end{algorithm}

Training is performed on Modal-managed 8$\times$H100 nodes using SkyRL's GRPO recipe with FSDP2, gradient checkpointing, and \texttt{torch.compile} on the policy. Each training step uses a batch of 16 prompts with 8 samples per prompt for 128 rollouts per step. Each rollout caps at 8 turns, 1024 maximum generated tokens per turn, and 16{,}384 maximum input length; trajectories that would exceed these caps are filtered via SkyRL's overlong-filtering mode. The optimizer is Adam with learning rate $5\times10^{-7}$, weight decay 0.01, and gradient clipping at max-norm 0.1, with a 20-step linear warm-up followed by constant schedule for ten epochs over the training set (1{,}120 total steps). KL loss is disabled; advantages are GRPO-normalized and standardized per group. Inference uses vLLM with four engines at tensor-parallel size 2, weight synchronization via NCCL, and \texttt{gpu\_memory\_utilization=0.80}. Checkpoints and validation evaluations are taken every 10 training steps. The canonical launch script is \texttt{experiments/a4-reversible-rl/modal\_smoke\_train.py} in the released codebase.

\begin{figure}[t]
  \centering
  \includegraphics[width=\linewidth]{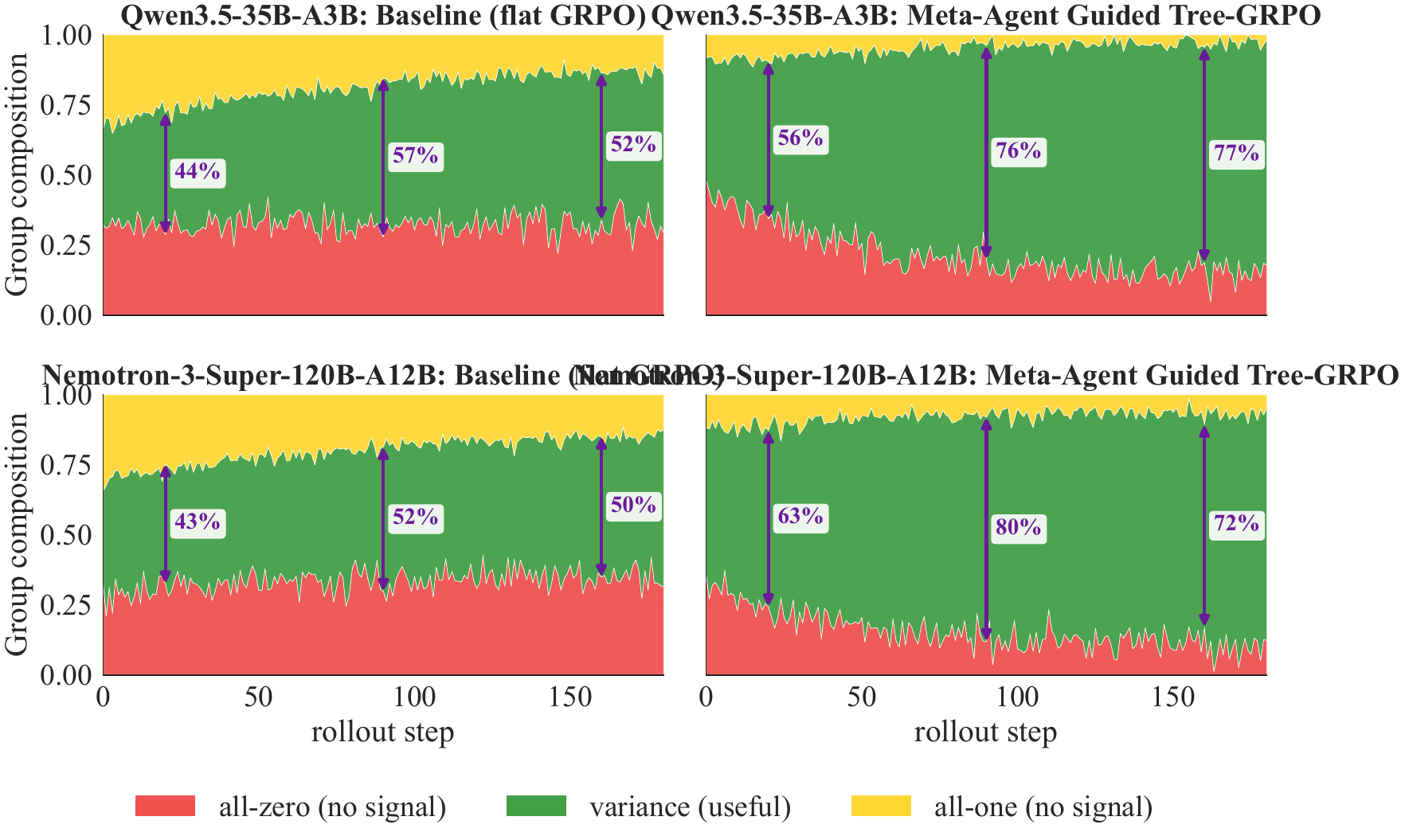}
  \caption{GRPO group composition over training (rows: base model; columns: setting). Tree-GRPO keeps the \emph{informative} (variance, green) fraction higher than Flat GRPO throughout, producing more gradient signal at matched compute. Flat GRPO's all-one share grows with training as easy tasks saturate, eating the variance band. Purple double-headed arrows annotate the variance share at steps 20/90/160.}
  \label{fig:mcts-groups}
\end{figure}

\begin{figure}[t]
  \centering
  \includegraphics[width=\linewidth]{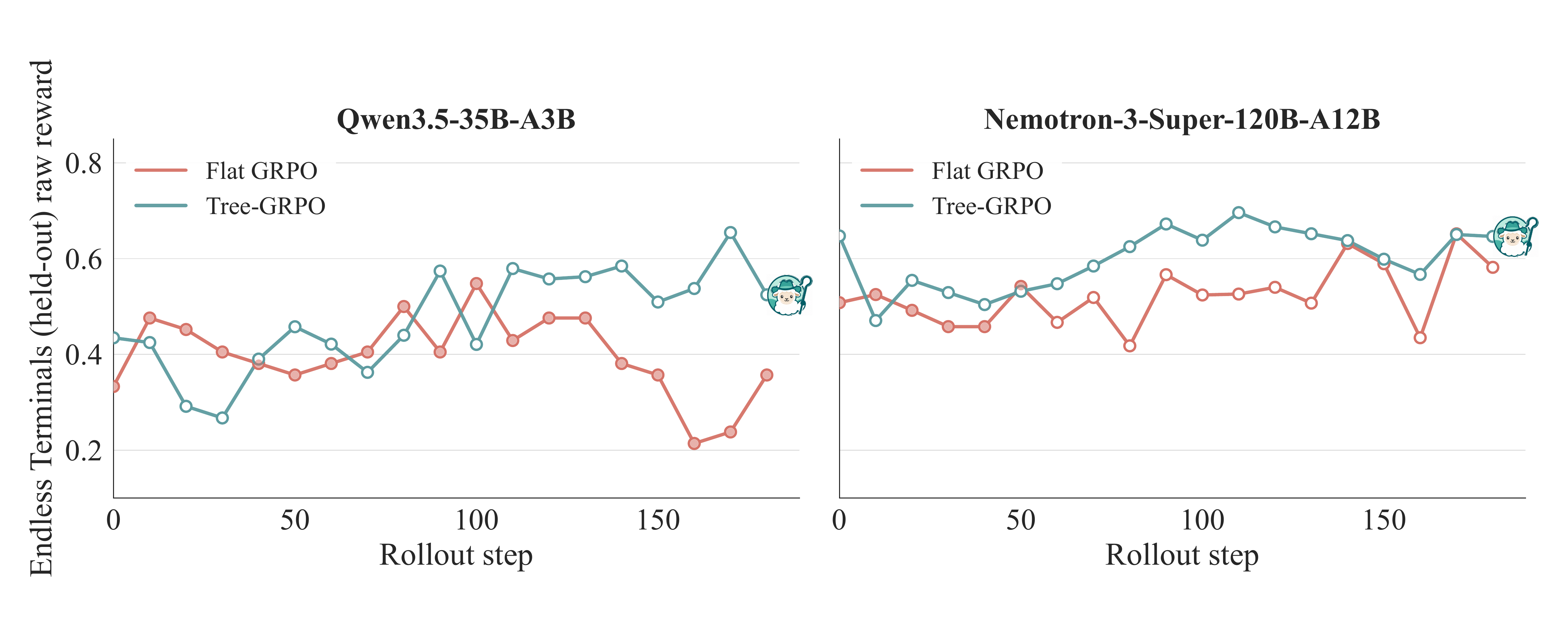}
  \caption{Held-out Endless Terminals evaluation, sampled every 10 training steps (raw, unsmoothed). Open circles plot the actual measured \texttt{val/endless\_mean\_reward}; solid line is a noise-preserving fit. Tree-GRPO climbs to and stabilises at a higher held-out reward than Flat GRPO on both base models. Final-policy Terminal-Bench~2.0 transfer is reported separately in \Cref{tab:tb2-final}.}
  \label{fig:mcts-ood}
\end{figure}

\subsection{Meta-Agent Guided Tree-RL: meta-agent qualitative examples}
\label{app:meta-agent-qualitative}

\begin{figure}[t]
  \centering
  \includegraphics[width=0.9\linewidth]{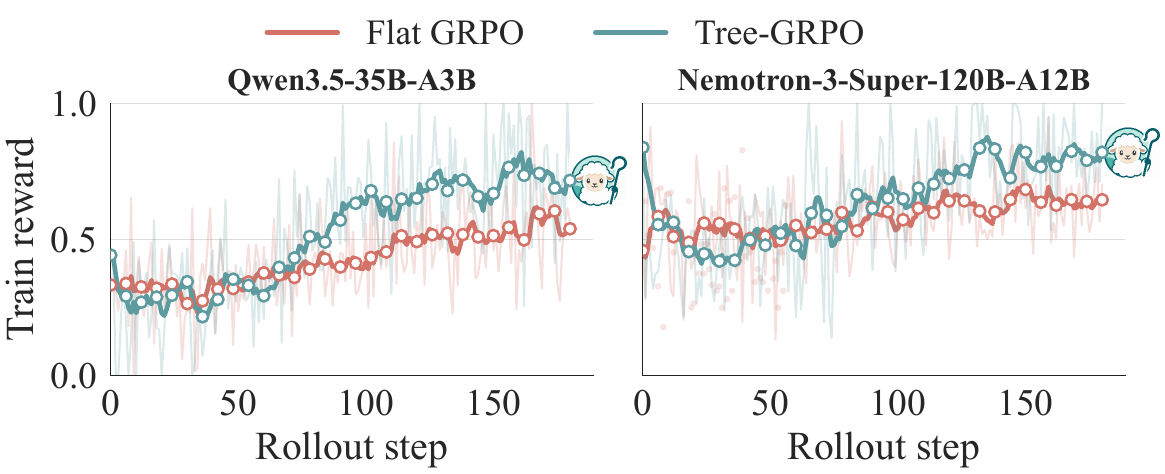}
  \caption{Train raw reward (mean over $G{=}8$ roots) for both base models, panels are Qwen3.5-35B-A3B (left) and Nemotron-3-Super-120B-A12B (right). Tree-GRPO ($K{=}4$, teal) reaches higher reward than Flat GRPO (red) at every rollout step. Faint dots are observed steps from the flat-baseline run; smooth lines are denoised trajectories.}
  \label{fig:mcts-curves}
\end{figure}

The tree-search rollouts in \Cref{sec:training} hand the fork choice to a stronger model (\texttt{claude-opus-4-7}). Given the inner agent's full transcript and final reward, it returns a turn $t^* \in [1, T{-}2]$ to fork at and the bash command it would run there. We re-render the command through the policy's renderer, so the tokens match what the policy itself would emit, and inject it as one of the $K$ branches at $t^*$.

This subsection walks through three trajectory shapes that show the meta-agent's behaviour: an early single-cause failure, an ambiguous failure with multiple plausible branches, and a long trajectory with a deeply-nested mistake. The trajectories are short enough to read end-to-end and were hand-constructed so a careful reader has a strong prior over the ``right'' branch turn. We check whether the meta-agent's pick matches that prior, whether its emitted command actually addresses the failure, and whether it stays well-behaved across trajectory lengths. (Three further cases in the released probe set, omitted for space, cover clean-success no-ops, late-turn range clamping, and a redundant tool-replacement variant of Case~1; results match the patterns shown here.)

\paragraph{Setup.} For each case, we render the inner-agent trajectory (system prompt, user task, and the alternating assistant tool calls and tool results) as a numbered turn list and send it to the meta-agent along with the final reward. The meta-agent returns a JSON object \texttt{\{branch\_turn, tool\_call, reason\}}, where \texttt{tool\_call.name = "bash"} and \texttt{tool\_call.arguments.command} is the bash payload.


\begin{figure}[h]
\centering
\begin{tcolorbox}[colback=red!5, colframe=red!60!black, title={\textbf{Case 1: Early mistake (T=4, reward=0.00)}}, fonttitle=\small, fontupper=\small]
\textbf{Task:} Install the \texttt{requests} package and verify it imports. \\[2pt]
\textbf{Trajectory:}
\begin{itemize}
  \itemsep -1pt
  \item \textbf{Turn 1:} \texttt{pip install request} \textit{(typo: missing the s)} $\rightarrow$ ``Successfully installed request-2025.10.0''
  \item \textbf{Turn 2:} \texttt{python -c "import requests"} $\rightarrow$ \texttt{ModuleNotFoundError: No module named requests}
  \item \textbf{Turn 3:} \texttt{python3 -c "import requests"} $\rightarrow$ same error.
  \item \textbf{Turn 4:} prints \texttt{sys.path}, gives up.
\end{itemize}
\end{tcolorbox}

\vspace{2pt}
\begin{tcolorbox}[colback=green!5, colframe=green!60!black, title={\textbf{Meta-agent decision (obvious turn: 1)}}, fonttitle=\small, fontupper=\small]
\textbf{branch\_turn:} \textbf{1} \quad \checkmark{} (matches obvious) \\[2pt]
\textbf{meta-agent action:} \texttt{pip install requests \&\& python -c "import requests; print(requests.\_\_version\_\_)"} \\[2pt]
\textbf{reason:} ``Turn 1 installed the wrong package `request' instead of `requests'; fixing the typo here resolves the entire failure.''
\end{tcolorbox}
\caption{Early-mistake case. The wrong package name on turn 1 dooms the rest of the trajectory. The meta-agent picks turn 1 and writes a one-liner that installs the right package and verifies the import in the same command.}
\label{fig:meta-case1}
\end{figure}


\begin{figure}[h]
\centering
\begin{tcolorbox}[colback=orange!5, colframe=orange!70!black, title={\textbf{Case 2: Ambiguous failure (T=5, reward=0.30)}}, fonttitle=\small, fontupper=\small]
\textbf{Task:} Serve the Flask app at \texttt{/app/app.py} on port 5000. \\[2pt]
\textbf{Trajectory:}
\begin{itemize}
  \itemsep -1pt
  \item \textbf{Turn 1:} read the app source.
  \item \textbf{Turn 2:} \texttt{pip install flask} (no python-version check).
  \item \textbf{Turn 3:} \texttt{python app.py \&} (background launch, no \texttt{nohup}).
  \item \textbf{Turn 4:} \texttt{curl localhost:5000/} $\rightarrow$ \texttt{Connection refused}.
  \item \textbf{Turn 5:} \texttt{ps aux} reveals the process is \texttt{<defunct>}.
\end{itemize}
\end{tcolorbox}

\vspace{2pt}
\begin{tcolorbox}[colback=orange!5, colframe=orange!70!black, title={\textbf{Meta-agent decision (multiple plausible turns)}}, fonttitle=\small, fontupper=\small]
\textbf{branch\_turn:} 3 \\[2pt]
\textbf{meta-agent action:} \texttt{cd /app \&\& nohup python app.py --host=0.0.0.0 > /tmp/flask.log 2>\&1 \&} \\[2pt]
\textbf{reason:} ``The app.py has no \_\_main\_\_ block, so running it directly exits; need to bind to 0.0.0.0 with \texttt{nohup} to keep it alive.''
\end{tcolorbox}
\caption{Ambiguous case. At least three turns offer plausible branches (skip-the-version-check, pip-vs-pip3, the background launch). The meta-agent traces the symptom at turn 5 (\texttt{<defunct>}) back to the launch on turn 3 and writes a concrete replacement: \texttt{nohup} with a log redirect and an explicit host binding. Demanding a runnable command pushes the meta-agent past hand-wavy diagnoses to a specific hypothesis we can run.}
\label{fig:meta-case2}
\end{figure}


\begin{figure}[h]
\centering
\begin{tcolorbox}[colback=red!5, colframe=red!60!black, title={\textbf{Case 3: Long trajectory, deep mid-mistake (T=9, reward=0.20)}}, fonttitle=\small, fontupper=\small]
\textbf{Task:} Fix the failing test in \texttt{/app/test\_utils.py}. \\[2pt]
\textbf{Trajectory:}
\begin{itemize}
  \itemsep -1pt
  \item \textbf{Turns 1--3:} explore, read \texttt{utils.py} (case mismatch in \texttt{normalize}), run pytest, see the failing assertion.
  \item \textbf{Turn 4:} edits \emph{the test file} with a fragile \texttt{sed} substitution \textit{(wrong file---fix belongs in \texttt{utils.py})}.
  \item \textbf{Turns 5--7:} doom-spiral---broken sed quotes, more sed patches, file syntax errors.
  \item \textbf{Turns 8--9:} attempts \texttt{git checkout} to revert, but repo is not git-tracked.
\end{itemize}
\end{tcolorbox}

\vspace{2pt}
\begin{tcolorbox}[colback=green!5, colframe=green!60!black, title={\textbf{Meta-agent decision (obvious turn: 4)}}, fonttitle=\small, fontupper=\small]
\textbf{branch\_turn:} \textbf{4} \quad \checkmark{} (matches obvious) \\[2pt]
\textbf{meta-agent action} (multi-line heredoc): \\[2pt]
\hspace*{1em}\texttt{cat > /app/utils.py <{}<'EOF'} \\
\hspace*{1em}\texttt{def normalize(s):} \\
\hspace*{1em}\texttt{\ \ \ \ return s.strip()} \\
\hspace*{1em}\texttt{EOF} \\[2pt]
\textbf{reason:} ``Fix the implementation rather than the test, since the task is to fix the failing test by correcting the underlying code.''
\end{tcolorbox}
\caption{Long-trajectory case. A 9-turn rollout with a wrong-file edit at turn 4 cascades into 5 turns of recovery thrash. The meta-agent picks turn 4 and writes a multi-line heredoc that rewrites \texttt{utils.py} (the file the agent should have edited) cleanly. Uniform sampling has no way to prefer turn 4 over the five recovery turns; only the meta-agent's pick concentrates the branch budget where it matters.}
\label{fig:meta-case3}
\end{figure}

\paragraph{Headline.} The meta-agent's choices share two patterns uniform sampling cannot reproduce. First, it prefers the earliest turn at which a downstream wrong choice was made, even when that turn was not itself a syntactically obvious failure (Case 1: package-name typo; Case 3: target-file choice). Second, when the trajectory shows symptom-then-cause dynamics (Case 2: a defunct process surfaces at turn 5, traceable to a launch issue at turn 3), it follows the causal chain rather than picking the symptom turn. Demanding an executable bash command, not just a turn index, forces the meta-agent past hand-wavy diagnoses (Case 2: a specific \texttt{nohup} invocation, not ``use proper launch flags''), and the command round-trips through the same renderer the policy uses, so the injected branch's tokens are indistinguishable from the policy's own emission at parse time.

\end{document}